\documentclass{article}

   \usepackage[final, nonatbib]{neurips_2023}

\usepackage[utf8]{inputenc} % allow utf-8 input
\usepackage[T1]{fontenc}    % use 8-bit T1 fonts
\usepackage{hyperref}       % hyperlinks
\usepackage{url}            % simple URL typesetting
\usepackage{booktabs}       % professional-quality tables
\usepackage{amsfonts}       % blackboard math symbols
\usepackage{nicefrac}       % compact symbols for 1/2, etc.
\usepackage{microtype}      % microtypography
\usepackage{xcolor}         % colors
\usepackage{graphicx}
\usepackage{tcolorbox}
\usepackage{multirow}    
\usepackage{amsmath,amssymb,amsfonts}
\usepackage{subfig}
\usepackage{pifont}

\usepackage{tikz}
\usetikzlibrary{trees}
\usetikzlibrary{shapes,arrows}
\usepackage{adjustbox}
\usepackage{cleveref}
\tikzstyle{block} = [draw, fill=white, rectangle, minimum height=2em, minimum width=2em]
\tikzstyle{sum} = [draw, fill=white, circle, scale=0.005, node distance=0.5cm]
\tikzstyle{input} = [coordinate]
\tikzstyle{output} = [coordinate]  
\tikzstyle{pinstyle} = [pin edge={to-,thin,black}]

\usepackage{ltablex}
\usepackage{xcolor}
\newcolumntype{b}{X}
\newcolumntype{s}{>{\hsize=2.2\hsize}X}

\title{Hierarchical Network Fusion for Multi-Modal Electron Micrograph Representation Learning with Foundational Large Language Models}
\author{
  Sakhinana Sagar Srinivas$^{1}$\thanks{Designed, programmed the software, and drafted manuscript}, \hspace{1mm}Geethan Sannidhi$^{2}$\thanks{Conducted experiments and analyzed visual results}, \textbf{Venkataramana Runkana$^{1}$}\\
  TCS Research$^{1}$, International Institute of Information Technology(IIIT)$^{2}$ \\
  \texttt{sagar.sakhinana@tcs.com}, \texttt{geethan.iiitp.ac.in}, \texttt{venkat.runkana@tcs.com}
}

\begin{document}% PatchFusion-CL-LLM
\maketitle

\vspace{-8mm}
\begin{abstract}
\vspace{-4mm}
Characterizing materials with electron micrographs is a crucial task in fields such as semiconductors and quantum materials. The complex hierarchical structure of micrographs often poses challenges for traditional classification methods. In this study, we propose an innovative backbone architecture for analyzing electron micrographs. We create multi-modal representations of the micrographs by tokenizing them into patch sequences and, additionally, representing them as vision graphs, commonly referred to as patch attributed graphs. We introduce the Hierarchical Network Fusion (HNF), a multi-layered network structure architecture that facilitates information exchange between the multi-modal representations and knowledge integration across different patch resolutions. Furthermore, we leverage large language models (LLMs) to generate detailed technical descriptions of nanomaterials as auxiliary information to assist in the downstream task. We utilize a cross-modal attention mechanism for knowledge fusion across cross-domain representations(both image-based and
linguistic insights) to predict the nanomaterial category. This multi-faceted approach promises a more comprehensive and accurate representation and classification of micrographs for nanomaterial identification. Our framework outperforms traditional methods, overcoming challenges posed by distributional shifts, and facilitating high-throughput screening.
\end{abstract}

\vspace{-8mm}
\section{Introduction}
\begin{figure}[htbp]
     \centering
     \subfloat[High intra-class dissimilarity: The electron micrographs of the same nanomaterial (\textit{MEMS} device) can exhibit a high degree of heterogeneity.]{\includegraphics[width=0.18\textwidth]{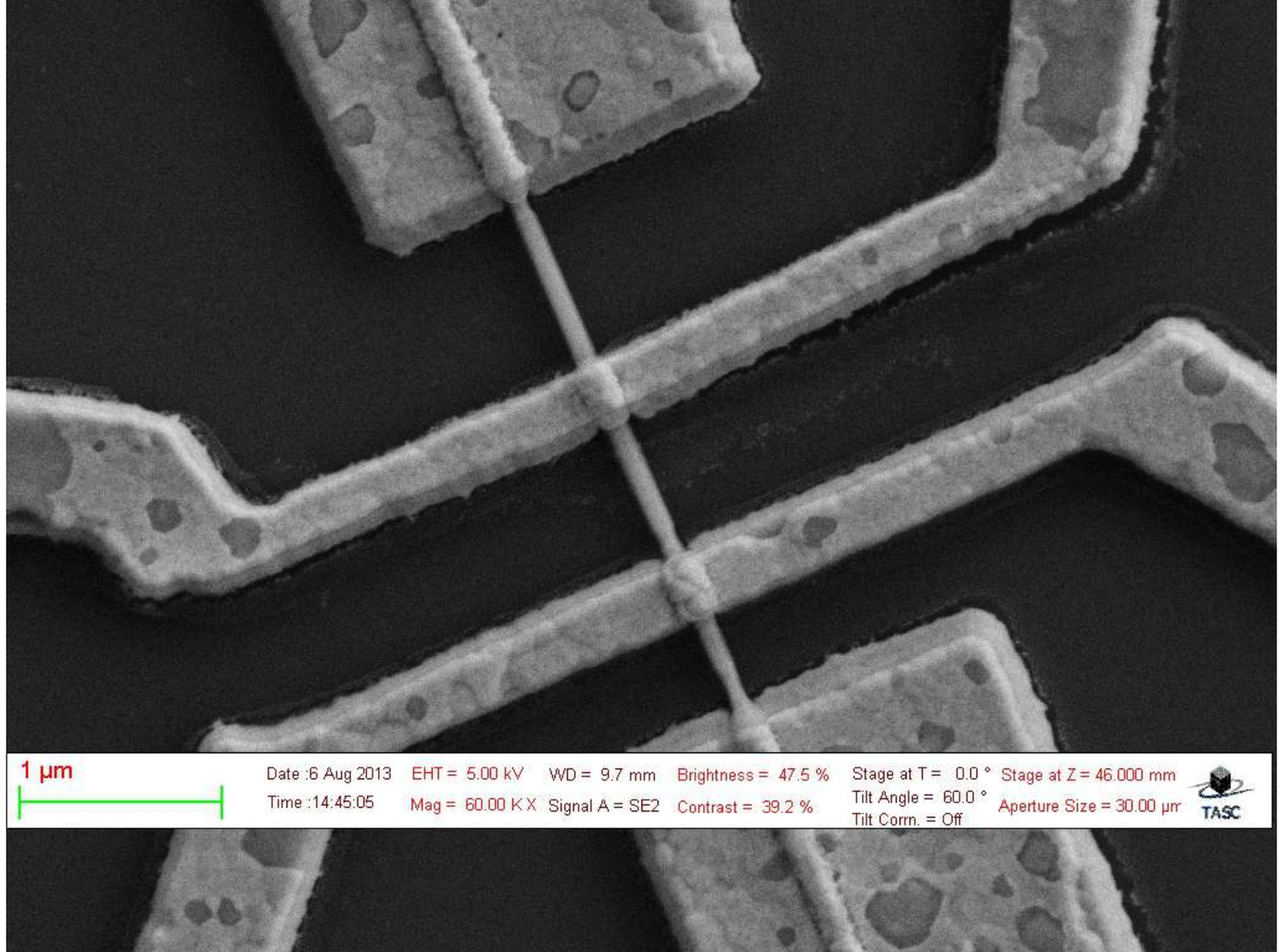}
     \includegraphics[width=0.18\textwidth]{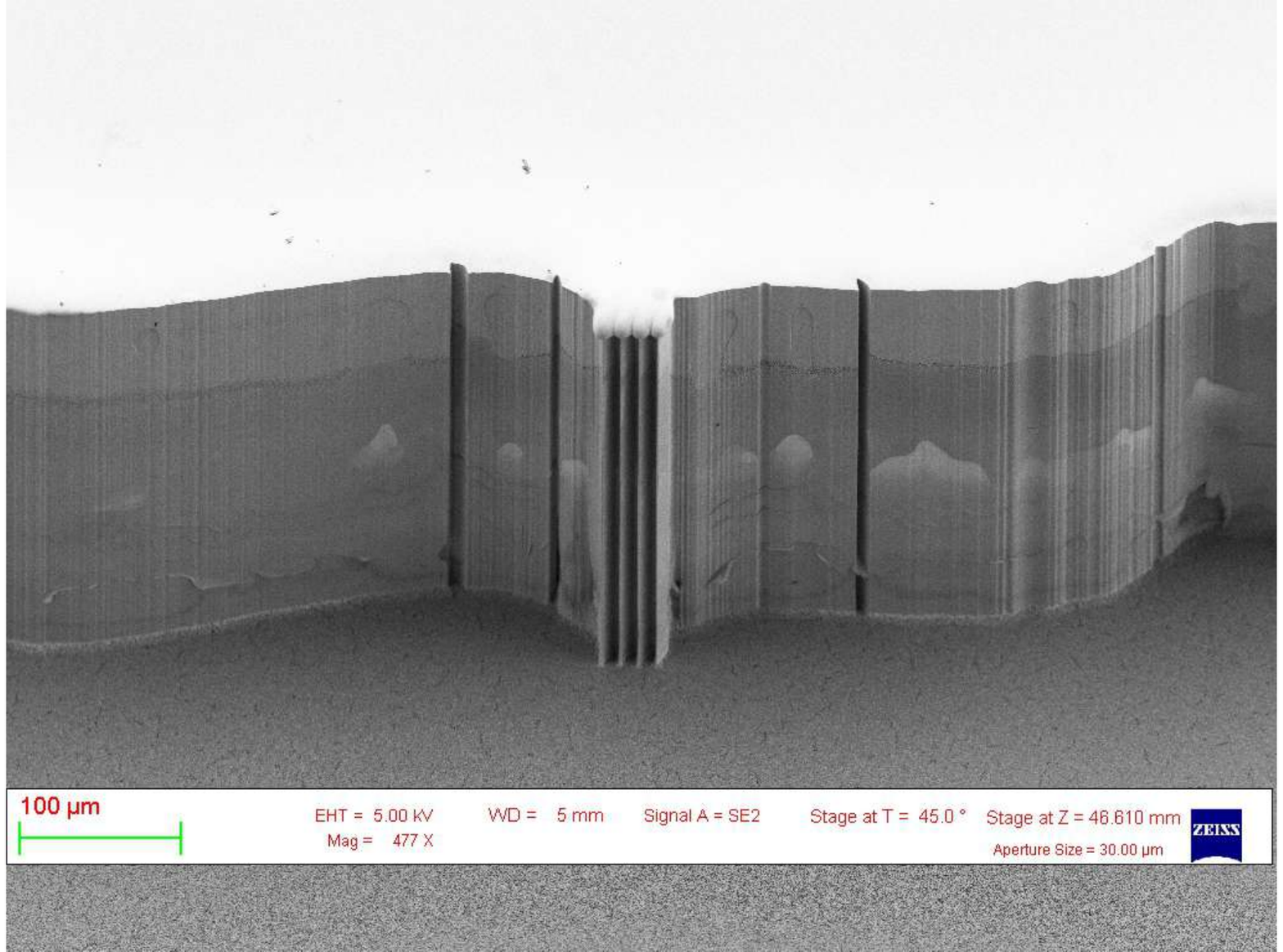}
     \includegraphics[width=0.18\textwidth]{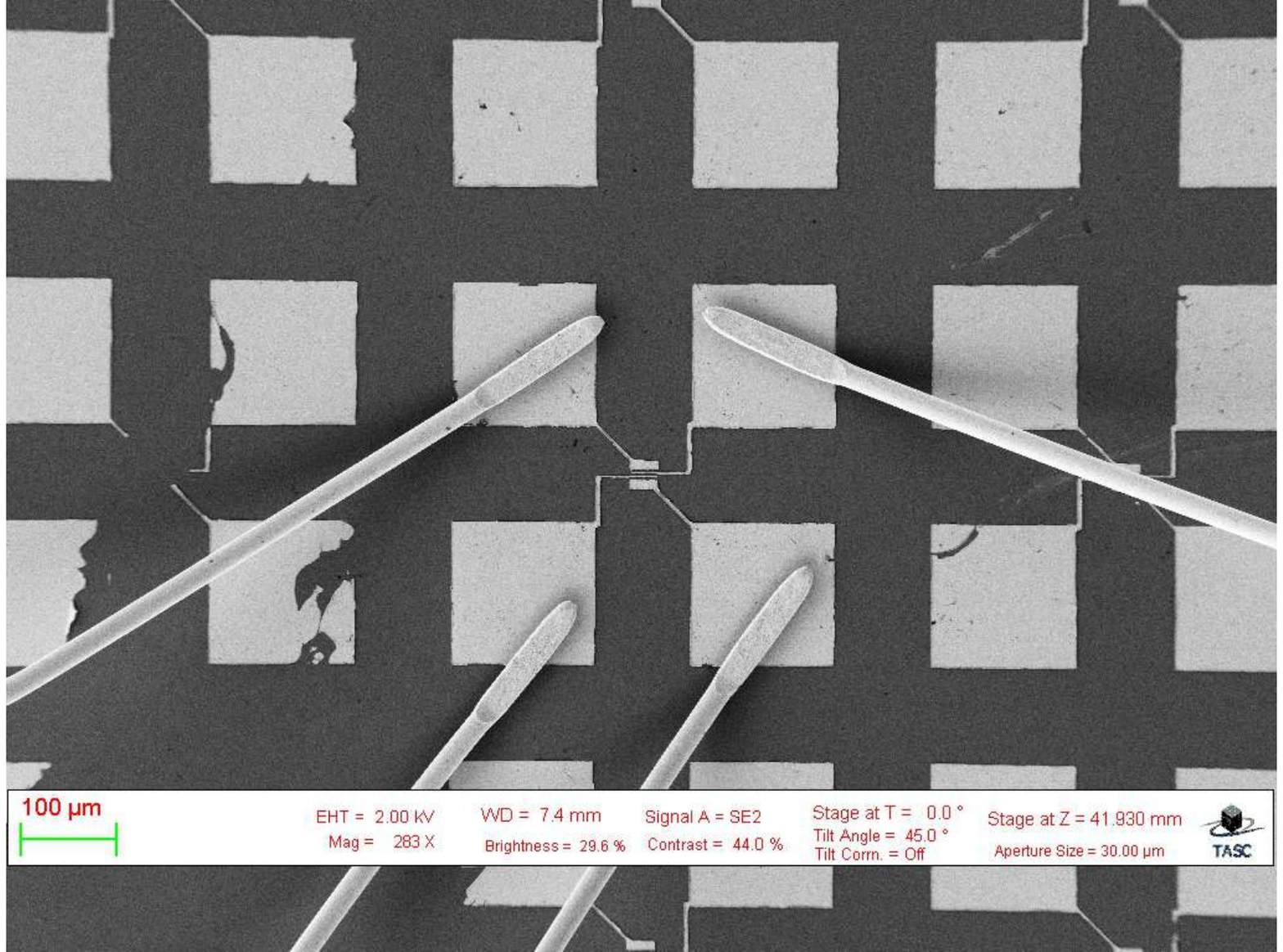}
     \includegraphics[width=0.18\textwidth]{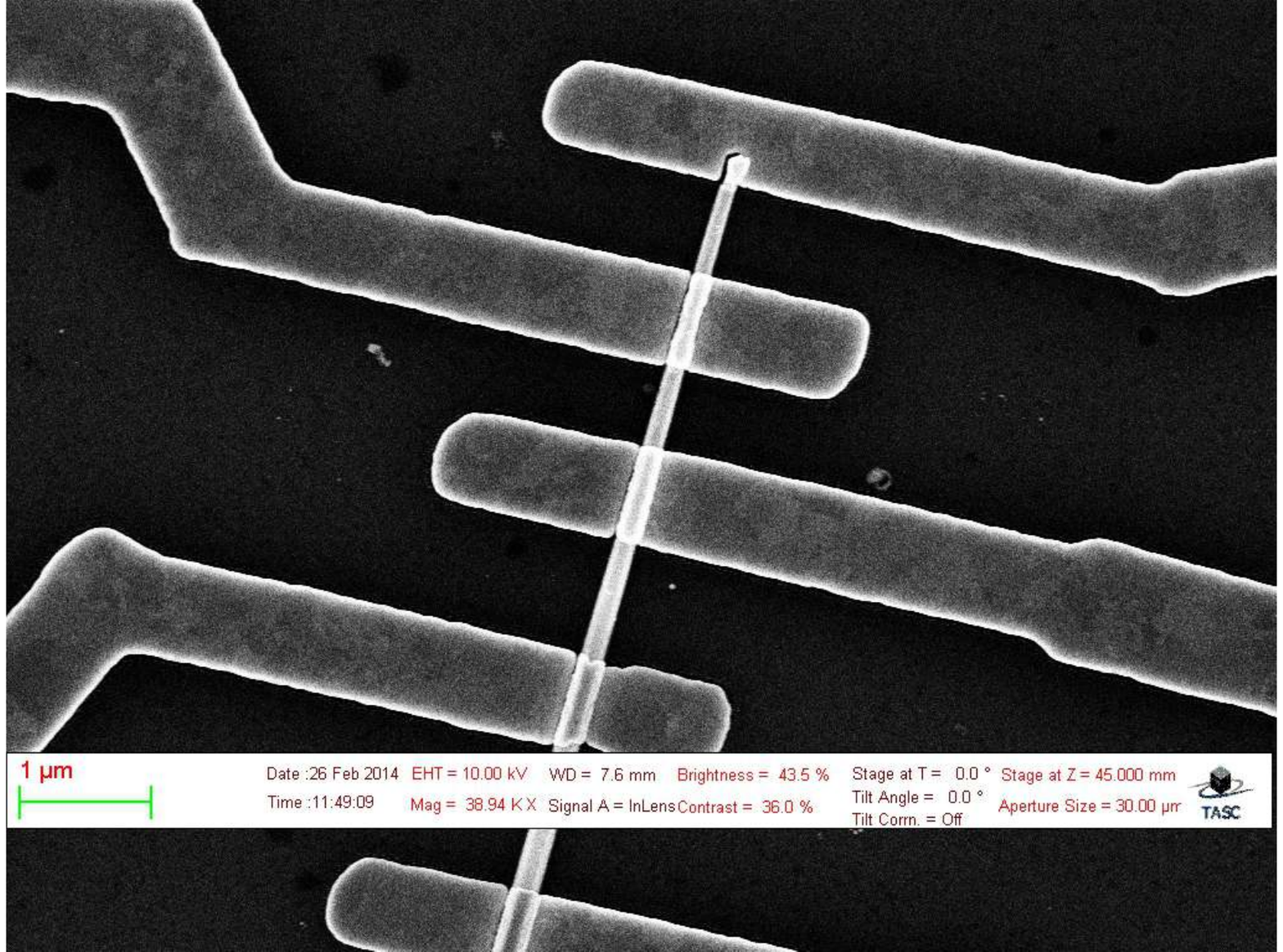}
     }
     \vspace{-3mm}
     \qquad
     \subfloat[High inter-class similarity: Electron micrographs across different nanomaterial categories (\textit{listed from left to right as porous sponges, particles, powders, and films}) exhibit a noteworthy degree of similarity.]{\includegraphics[width=0.18\textwidth]{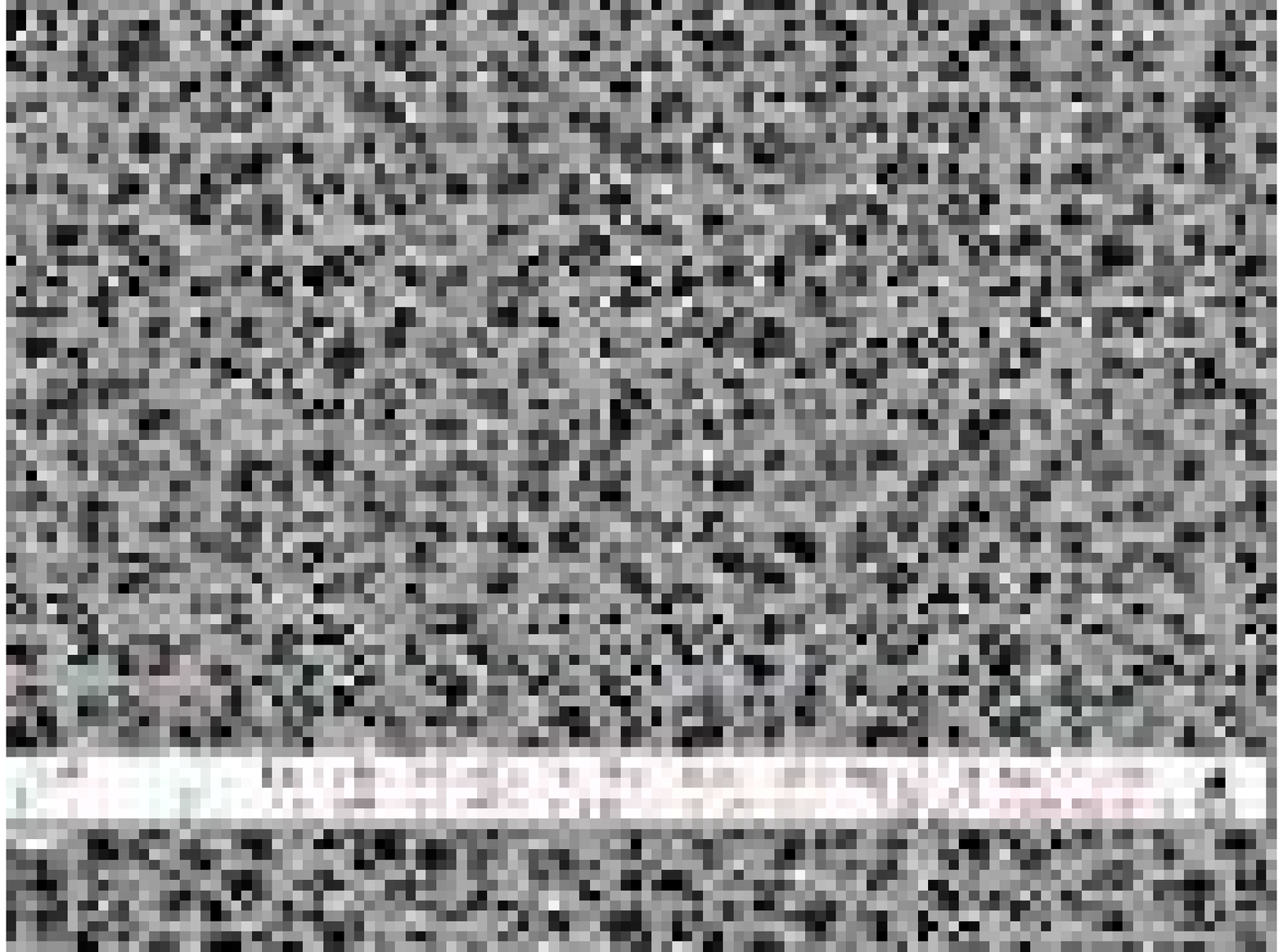}
     \includegraphics[width=0.18\textwidth]{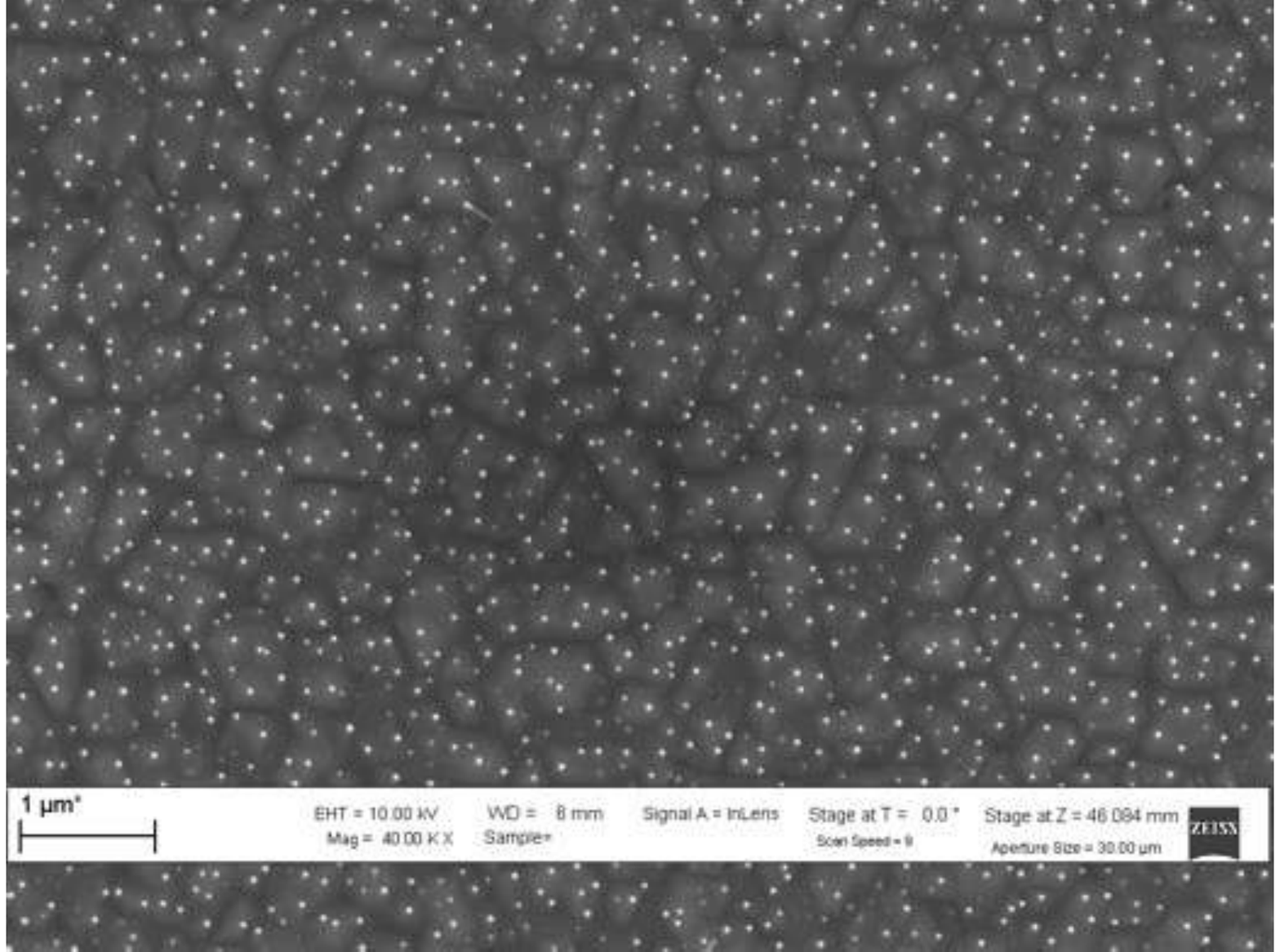}
     \includegraphics[width=0.18\textwidth]{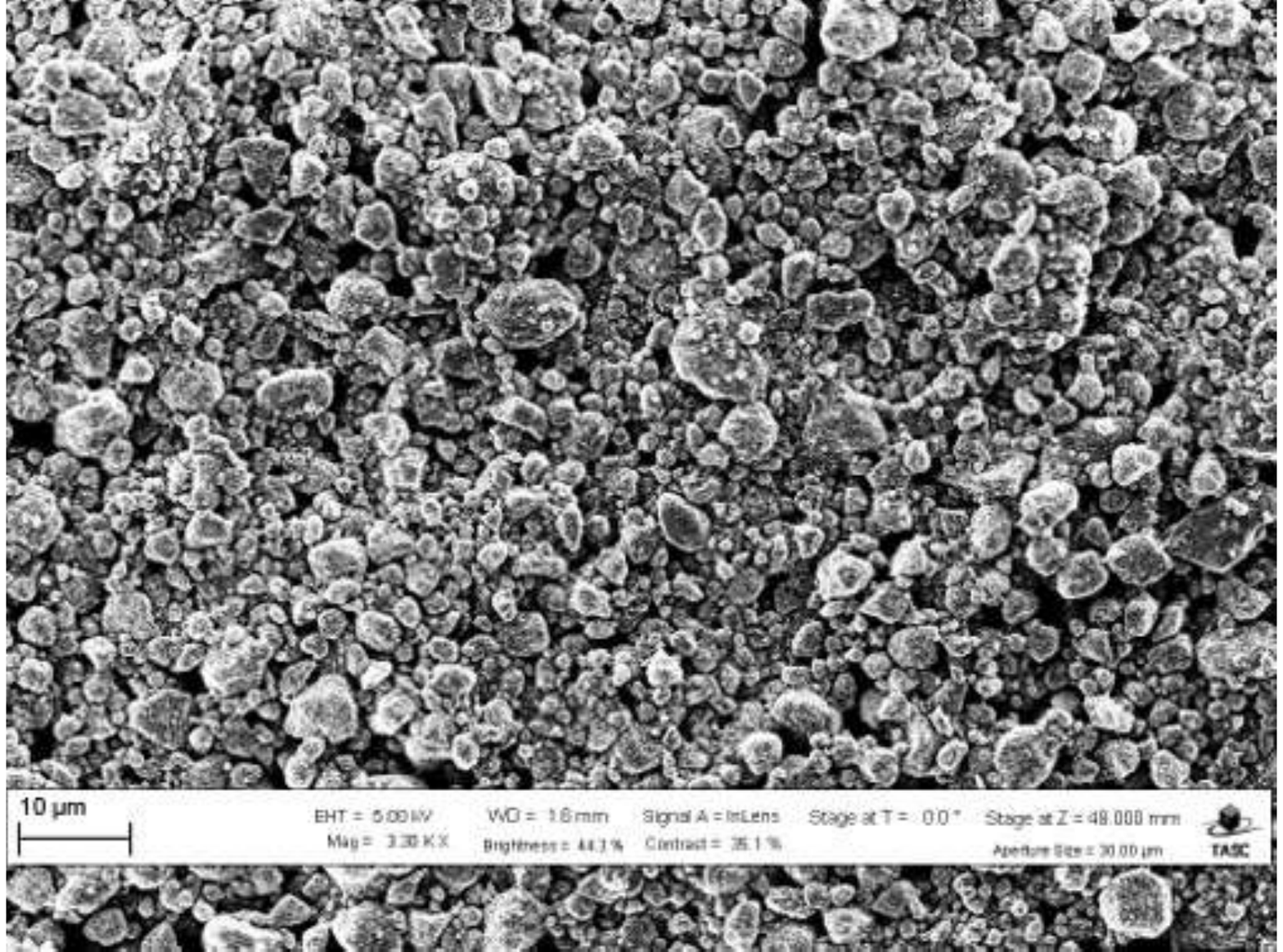}
     \includegraphics[width=0.18\textwidth]{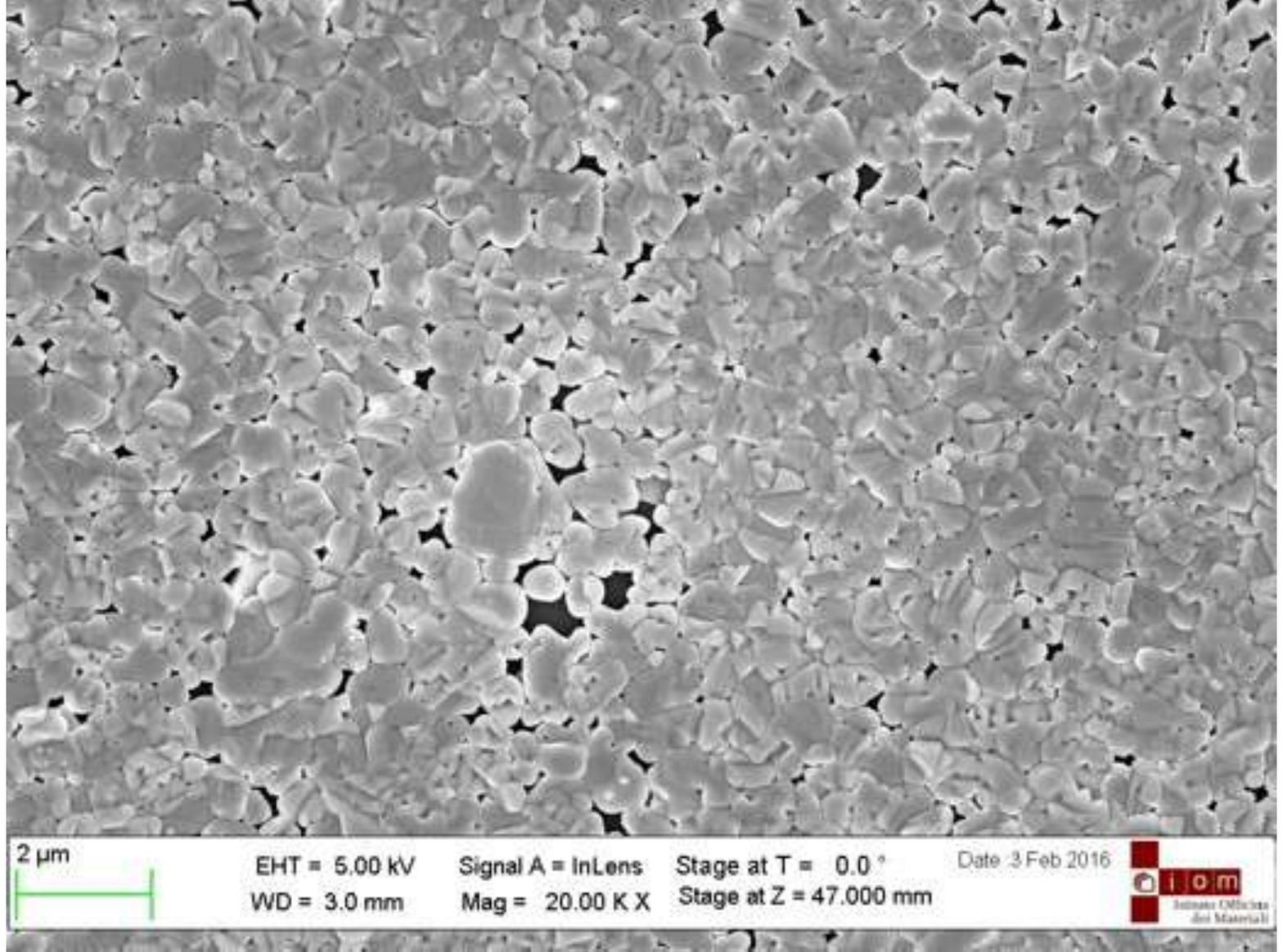}
     }
     \vspace{-3mm}
     \qquad
     \subfloat[Multi-spatial scales of patterns: The spatial heterogeneity of visual patterns in electron micrographs of \textit{nanoparticles} is evident.]{\includegraphics[width=0.18\textwidth]{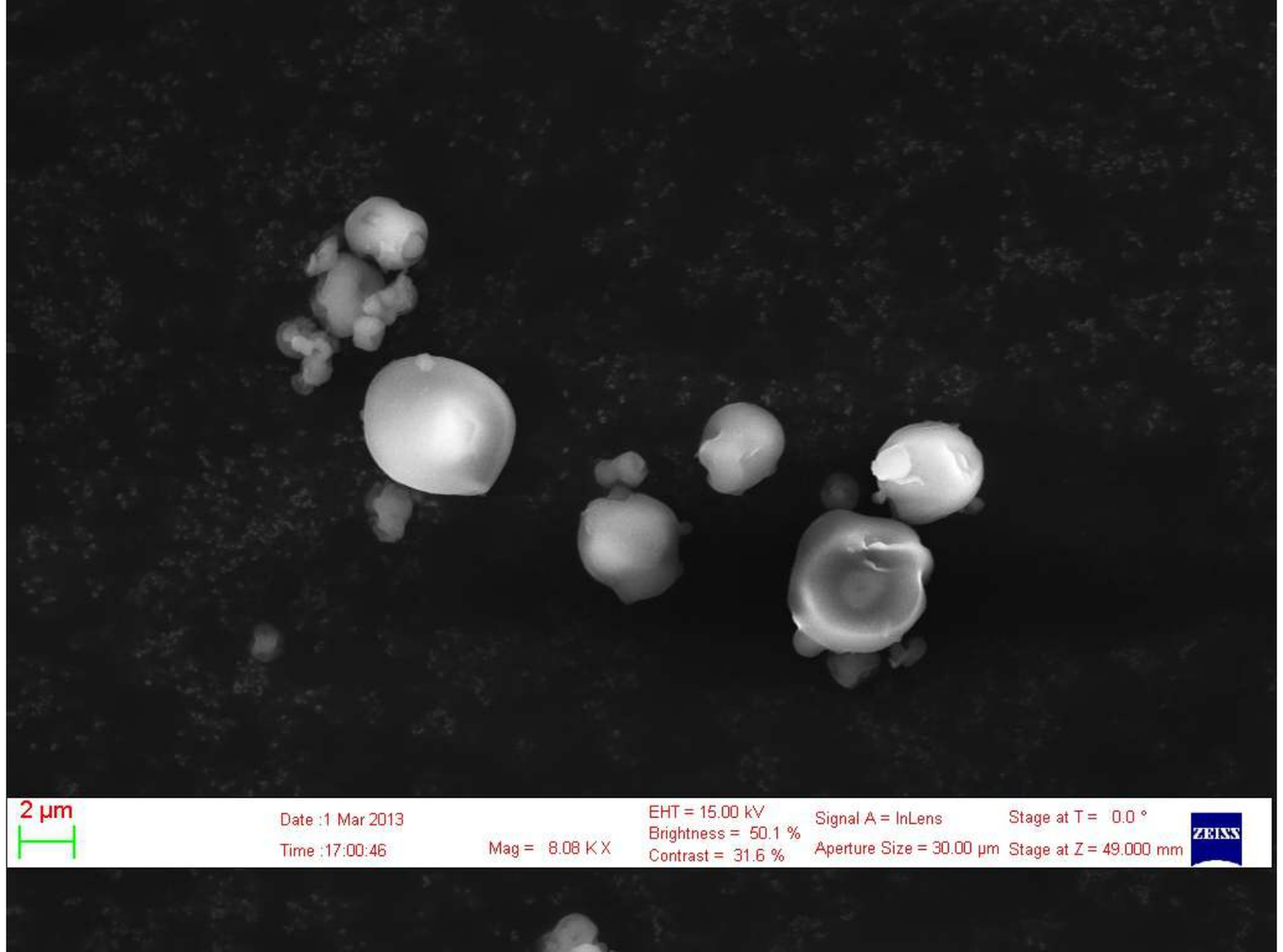}
     \includegraphics[width=0.18\textwidth]{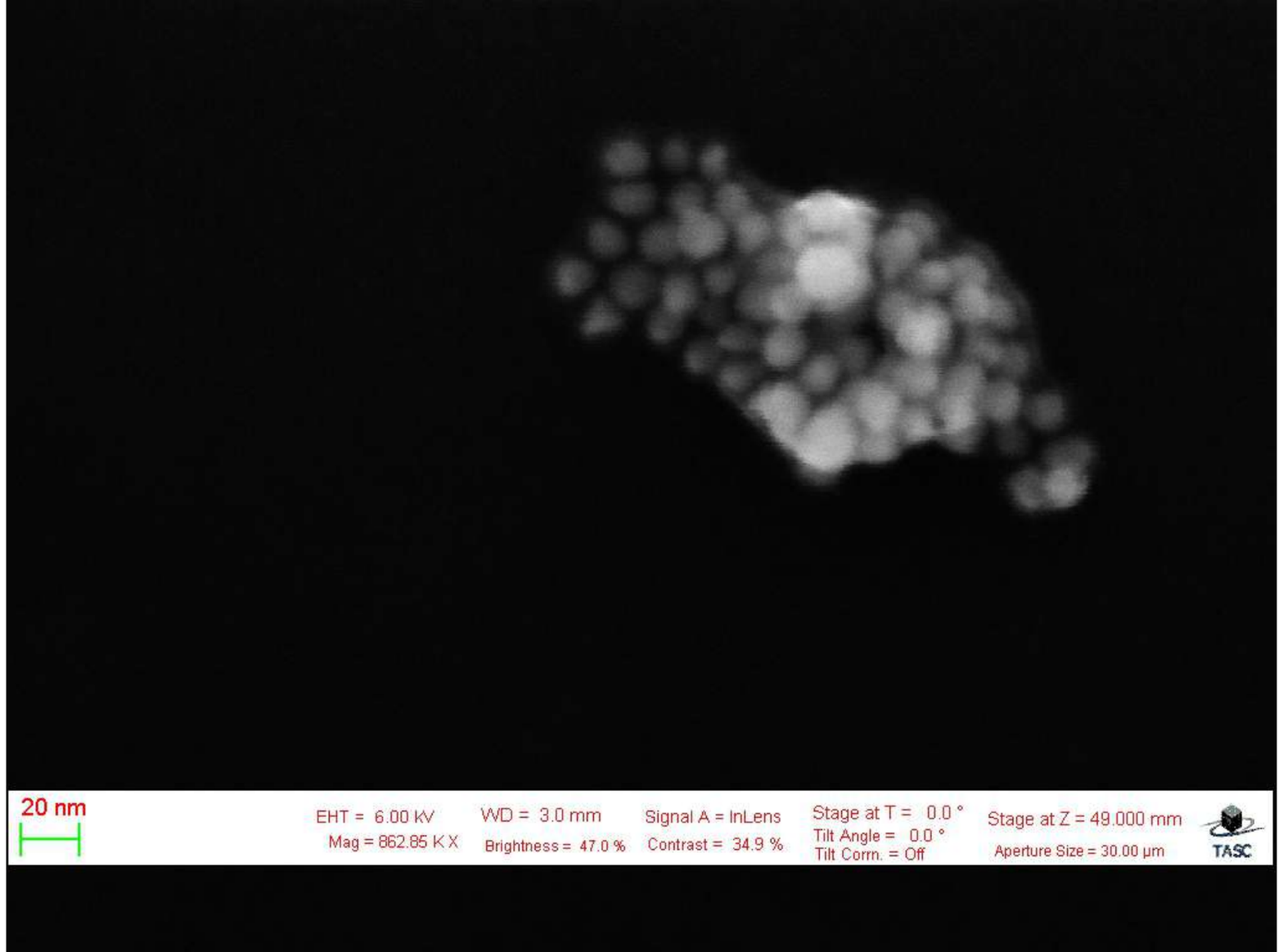}
     \includegraphics[width=0.18\textwidth]{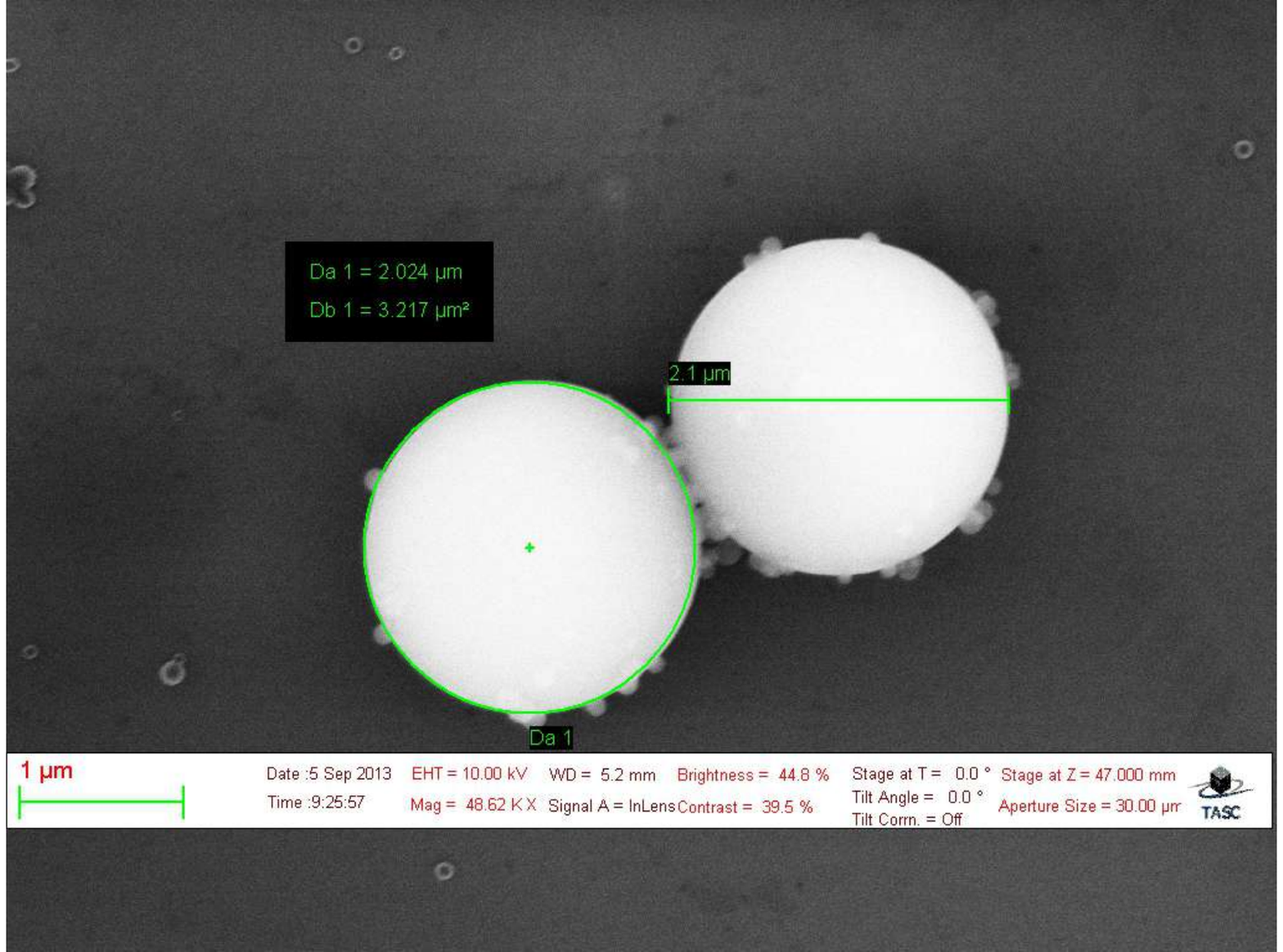}
     \includegraphics[width=0.18\textwidth]{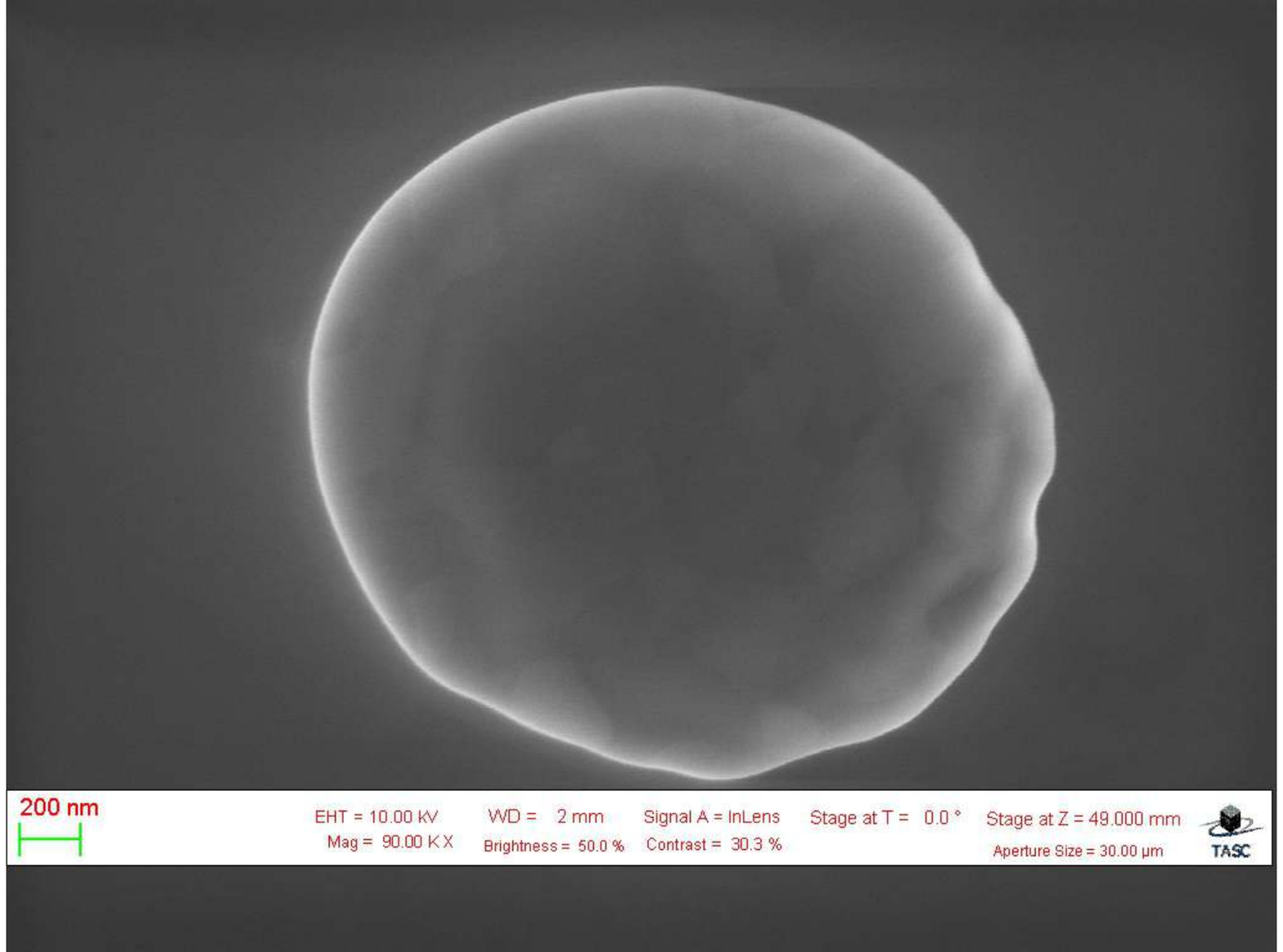}}
     \vspace{-2mm}
     \caption{The figure provides a visual representation of the challenges of classifying electron micrographs in the SEM dataset(\cite{aversa2018first}).}
     \vspace{-5mm}
     \label{fig:figure1}
\end{figure}

\vspace{-4mm}
Semiconductors are the foundation of modern electronics, driving advancements in computing, communication systems, transportation systems, and space exploration. The precise design, development, and testing of semiconductor devices is essential for ensuring the reliability, durability, and performance of high-tech chips. Advanced imaging and analysis techniques\cite{holt2013sem} are key to fabricating and integrating nanoscale components and enabling advanced inspection, which is essential for the development of next-generation miniaturized semiconductor devices\cite{campseuropean}, with sizes now reaching as small as 7 nm or even smaller. However, the increased complexity of producing chips under 7 nanometers introduces greater potential for error, jeopardizing the consistency of high-quality chip production and magnifying variability in chip performance.  The semiconductor industry utilizes various electron beam tools, including scanning and transmission electron microscopy, to create high-resolution images of these devices. These images, known as electron micrographs, reveal the complex microstructures of materials, which are crucial for the accurate design and evaluation of semiconductor devices. The fabrication of nanoscale components is a challenging task that requires precise control over the manufacturing process. Furthermore, these images facilitate monitoring of the process and defect detection, enabling subsequent process optimization or design adjustments to mitigate defects. The autolabeling of electron micrographs for nanomaterial identification, while advantageous, remains a significant challenge. Figure \ref{fig:figure1} shows the challenges in nanomaterial identification tasks. This is largely attributed to distributional shifts such as manufacturing variations or material property changes, exacerbated by high intra-class dissimilarity within nanomaterials, high inter-class similarity between different nanomaterials, and the existence of visual patterns at multiple scales or spatial heterogeneity. To overcome the challenges in this work, we propose an end-to-end framework for automatic nanomaterial identification based on hierarchical network fusion for multi-modal electron micrograph representation learning with large language models (referred to as ``MultiFusion-LLM" for shorthand notation). We hypothesize that electron micrographs exhibit hierarchical dependencies among patches (segmented portions of an electron micrograph). These dependencies can be captured using multiple patch sequences and vision graph structures at different spatial resolutions of the patches. To explore this, we tokenize the electron micrographs into grid-like patches to obtain a patch sequence. Additionally, we represent the micrograph as a vision graph, where patches are connected by undirected edges that represent pairwise visual similarity. Figure \ref{fig:figure0} shows the modalities (patch sequence, graph) that offer unique insights and assist in capturing complex patterns. We introduce a $<\hspace{-1mm}\textit{cls}\hspace{-1mm}>$ token to the patch sequence and a virtual node to the vision graph. This special token/virtual node encapsulates the entire patch sequence and captures global graph information in their respective contexts. We aim to capture fine- and coarse-grained hierarchical dependencies by treating the micrographs as sequence structures and vision graphs at multiple scales of patch size. The main contributions of this work can be summarized:

\vspace{-3mm}  
\begin{itemize} 
\item[\ding{51}] We have developed the Hierarchical Network Fusion (HNF), a cascading network architecture that enhances the classification accuracy by analyzing and integrating two complementary representations of electron micrographs: patch sequences and vision graphs, which are created at various patch sizes. Vision graphs, constructed using a nearest-neighbor graph technique, identify local patch relationships and capture graph-structured priors. Meanwhile, patch sequences help in capturing spatial dependencies between various patches in a micrograph, going beyond the limitations of sparse graph structure priors. The HNF is a multi-layered network featuring an inverted pyramid architecture that generates a multi-scale representation of an electron micrograph by creating a series of patch sequences and vision graphs at different scales of patch size. This inverted pyramid is constructed by progressively increasing the patch size at each layer. Each layer of the pyramid represents the original micrograph-based patch sequence and vision graph at a distinct scale, offering increasingly higher resolutions. By considering information at multiple scales, the HNF facilitates a more comprehensive representation of the electron micrograph, capturing both fine- and coarse-grained details. At each layer, the patch embeddings are iteratively refined using bidirectional Neural Ordinary Differential Equations (Neural ODEs) \cite{chen2018neural}, while the Graph Chebyshev Convolution (GCC) Networks \cite{he2022convolutional, defferrard2016convolutional} encode the vision graphs in a layer-wise manner to compute graph-level embeddings. A mixture-of-experts (MOE) technique with a gating mechanism optimally combines predictions from both modalities at each layer by calculating a weighted sum of classification token and virtual node embedding to improve classification accuracy. This facilitates an intermodal mutual information exchange, fostering interaction and knowledge integration between the two modalities. This innovative approach enables the seamless integration of causal information from patch sequences to refine the vision graph embeddings, and structural and semantic information from visual graphs to ground the patch embeddings, fostering enhanced interaction and knowledge fusion within the architecture. Our framework constructs a multi-scale representation of a micrograph with the aim of optimally preserving both the high-level features and structural information embedded in the graphs, as well as the causal relations embedded in the patch sequences, thereby enabling a more comprehensive representation of the micrograph.
\vspace{-1.5mm}
\item[\ding{51}] Our approach utilizes Zero-shot Chain-of-Thought (Zero-Shot CoT) prompting with large language models (LLMs)\cite{brown2020language, chowdhery2022palm, touvron2023llama} to generate technical descriptions of nanomaterials, including synthesis methods, properties, and applications. We pre-train smaller language models (LMs) \cite{devlin2018bert, he2020deberta} through self-supervised masked language modeling (MLM)\cite{barba2023dmlm, devlin2018bert} on these generated textual descriptions, enabling domain-specific customization for improved language understanding. Subsequently, we fine-tune the pre-trained LMs for task-adaptation to compute contextualized token embeddings for nanomaterial identification tasks. We employ a weighted sum-pooling attention mechanism to compute text-level embeddings from token embeddings, encapsulating the vast domain-specific knowledge present in the text data. Our approach leverages LLM-based technical descriptions on nanomaterials to identify characteristic features that distinguish them from other nanomaterial categories, incorporating domain-specific knowledge as auxiliary information for downstream training.
\end{itemize}

\vspace{-6mm}
\section{Problem Statment}
\label{pm}
\vspace{-4mm}
In this study, the focus is on the electron micrograph classification task, a type of inductive learning task where the objective is to assign labels to new, unseen micrographs utilizing a labeled dataset denoted as $\mathcal{D}_L = (\mathcal{I}_L, \mathcal{Y}_L)$. A multi-modal encoder, formulated as the non-linear function $f_{\gamma} : \mathcal{I} \rightarrow \mathcal{Y}$ is trained on labeled dataset to predict labels ($\mathcal{Y}_U$) of unlabeled micrographs ($\mathcal{I}_U$). Here, $\gamma$ denotes the trainable parameters. The objective is to minimize the loss function $\mathcal{L}_{\mathcal{I}}$, which is articulated as

\vspace{-2mm}
\resizebox{0.90\linewidth}{!}{
\begin{minipage}{\linewidth}
\begin{equation}
 \min _{\gamma} \mathcal{L}_{\mathcal{I}}\left(\mathcal{I}_{i}, \gamma\right)=\sum_{\left(\mathcal{I}_{i}, y_{i}\right) \in \mathcal{D}_{L}} \ell\big(f_{\gamma}(\mathcal{I}_{i}), y_{i}\big)
 \end{equation}
 \end{minipage}
}

\vspace{-3mm}
where $y^{\text{pred}}_{i} = f_{\gamma}(\mathcal{I}_i)$ denote the multi-modal encoder predictions and $\ell(\cdot, \cdot)$ denotes the cross-entropy loss. 

\vspace{-6mm}

\vspace{-1mm}
\begin{figure}[!ht]
\centering
\resizebox{0.95\linewidth}{!}{ 
\hspace*{0mm}\includegraphics[keepaspectratio,height=4.5cm,trim=0.0cm 0.0cm 0cm 2.25cm,clip]{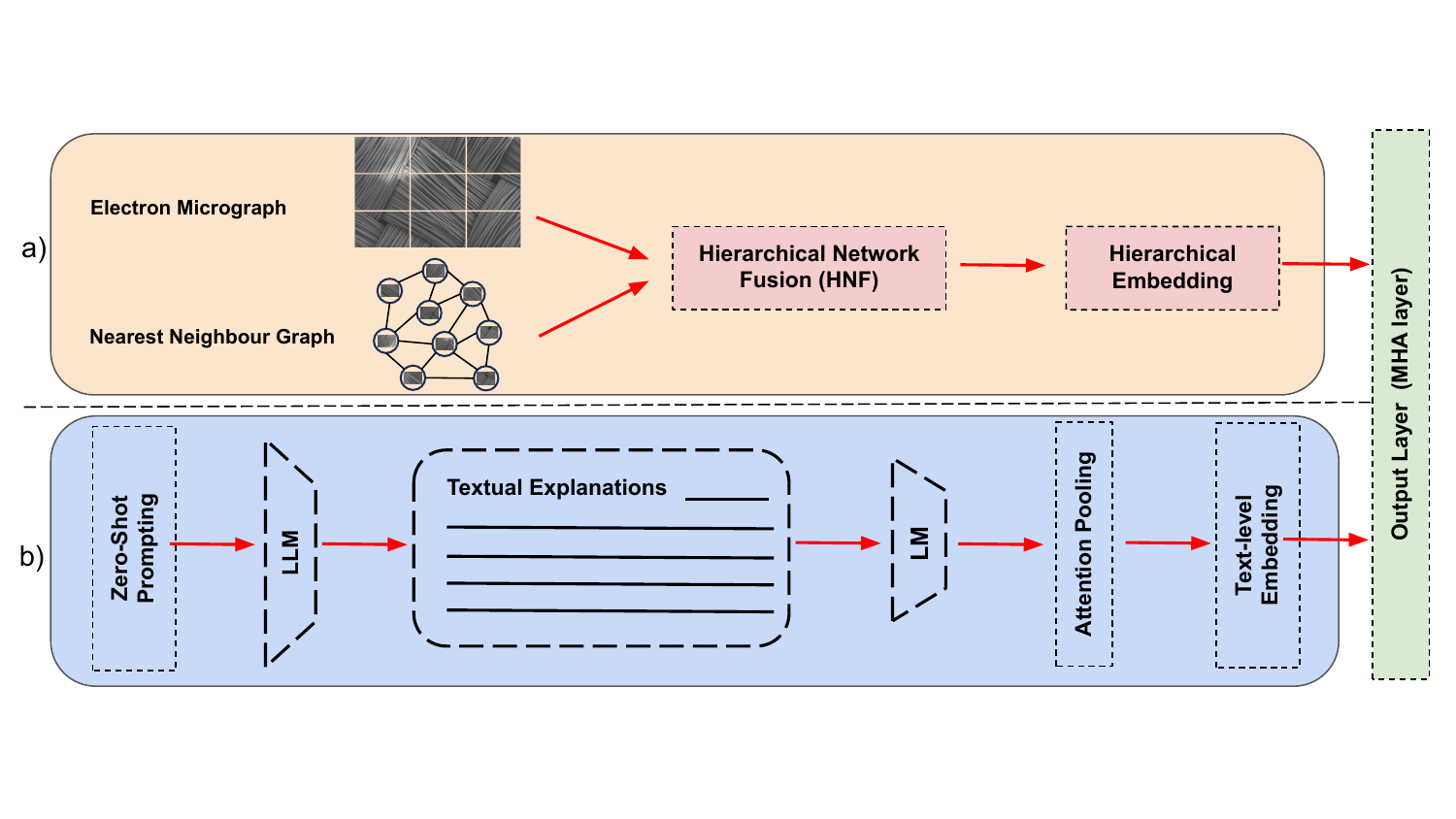} % left, bottom, right, top
}
\vspace{-13mm}
\caption{Our framework includes three methods: (a) Hierarchical Network Fusion (HNF), (b) Zero-shot Chain-of-Thought (Zero-Shot CoT) prompting with large language models (LLMs), and (c) an output layer modeled with the multi-head attention (MHA) mechanism \cite{vaswani2017attention} for integrating cross-domain embeddings and facilitating label prediction. LLMs take a prompt, not an electron micrograph, as input.}
\label{fig:figure3}
\vspace{-5mm}
\end{figure}

\vspace{-3mm}
\section{Proposed Method}
\label{pm}

\vspace{-3mm}
\subsection{Formalism}
\vspace{-3mm}
Let's consider an input electron micrograph denoted by $\mathbf{I'''}$, which has dimensions of $h \times w \times c$, where $h$, $w$, and $c$ represent the height, width, and number of channels of the micrograph, respectively. We divide the micrograph into a grid of patches, each having dimensions of $p \times p \times c$, with $p$ representing the patch size. The number of patches along each spatial dimension is given by $n = hw/p^2$. Subsequently, we reshape the 3D micrograph into a 2D patch tensor, denoted as $\mathbf{I''} \in \mathbb{R}^{\hspace{0.5mm}n \times (p^2c)}$. These patches are linearly transformed to create a new tensor, $\mathbf{I'} \in \mathbb{R}^{\hspace{0.25mm}n \times d}$, where $d$ is the patch embedding dimension. To account for the position of each patch within the micrograph, we introduce position embeddings represented by a matrix $\mathbf{E}_{pos} \in \mathbb{R}^{\hspace{0.5mm}n \times d_{pos}}$, where $d_{pos}$ denotes the position embedding dimension. We then add the position embedding matrix to the transformed patch tensor $\mathbf{I'}$, resulting in the final tensor $\mathbf{I} \in \mathbb{R}^{\hspace{0.25mm}n \times d}$. In general, $d_{pos} = d$. Finally, we construct a k-nearest neighbors graph to analyze the pairwise relationships between micrograph patches. This vision graph, denoted as $\mathcal{G}$, is undirected and represents the connectivity of patches based on their pairwise proximity. The graph structure is described by a binary adjacency matrix, $A \in \mathbb{R}^{n \times n}$. If patch $j$ is one of the k-nearest neighbors of patch $i$, then $A_{ij} = 1$; otherwise, $A_{ij} = 0$.

\vspace{-4mm}
\subsection{Hierarchical Network Fusion(HNF)}
\label{HNF}
\vspace{-3mm}
We tokenize electron micrographs by dividing them into grid-like patches. This approach yields two complementary representations of micrographs:  (a) We represent an electron micrograph as a vision graph, where patches are connected by edges that represent pairwise visual similarity constructed using a nearest-neighbor graph technique. The vision graph captures local patch relationships and utilizes graph-structural priors to analyze pairwise spatial dependencies within the micrograph. (b) Additionally, we represent electron micrographs as a patch sequence, capturing pairwise spatial dependencies beyond the original sparse graph structure between different patches within a micrograph. Representing electron micrographs as both patch sequences and vision graphs serves distinct purposes in their respective contexts. We append a classification token ($<\hspace{-1mm}\textit{cls}\hspace{-1mm}>$) to a patch sequence to obtain an embedding of the entire patch sequence that captures global information. We augment each vision graph by introducing a virtual node that is bidirectionally connected to all the other nodes in the graph through virtual edges. These virtual edges represent the pairwise relations between each real node and the virtual node. The virtual node embedding captures the long-range dependencies between nodes by considering the global information of the vision graph. We hypothesize that electron micrographs exhibit hierarchical dependencies among patches, which can be captured using multiple patch sequences or vision graph structures at different spatial resolutions of the patches. We present Hierarchical network fusion (HNF), a cascading network architecture that constructs a multi-scale representation of an electron micrograph by creating a series of patch sequences and vision graphs at multiple scales of patch sizes with increasing resolutions. 

\vspace{-4mm}
\begin{figure}[!ht]
\centering
\resizebox{0.975\linewidth}{!}{ 
\hspace*{0mm}\includegraphics[keepaspectratio,height=4.5cm,trim=0.0cm 0.0cm 0cm 0.0cm,clip]{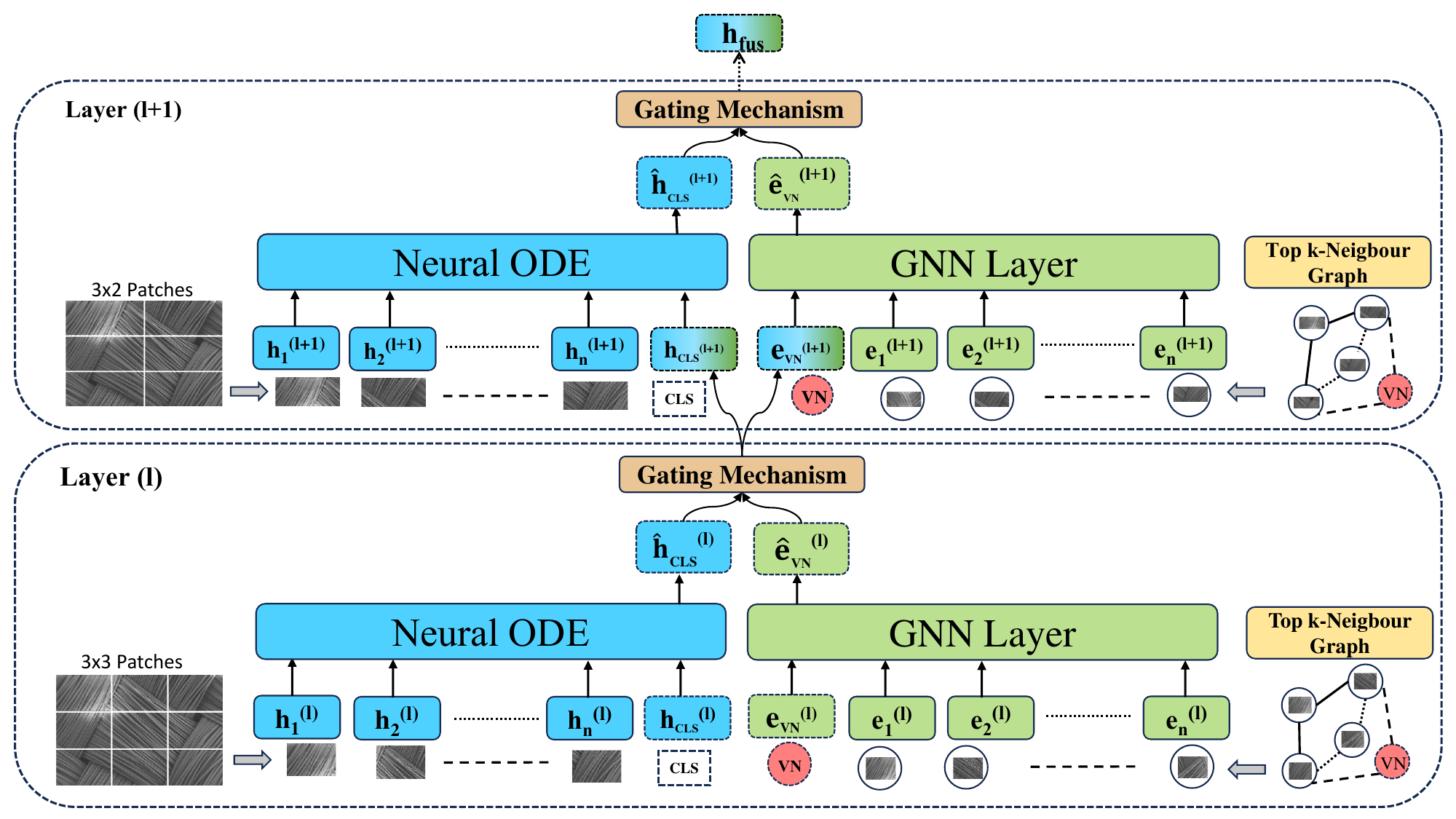} % left, bottom, right, top
}
\vspace{-3mm}
\caption{Overview of the HNF module. The HNF module utilizes a multi-layered network with increasing patch sizes to represent the electron micrograph-based patch sequence and vision graph at various scales, facilitating computation of hierarchical embeddings that encapsulate the global context. The cascaded structure incorporates multiple stacked layers; each layer involves bidirectional Neural ODEs and Graph Chebyshev convolution to compute patch sequence and vision graph embeddings, respectively. A gating mechanism integrates these cross-domain embeddings, generating unified hierarchical embeddings that offer a comprehensive view of the electron micrographs. Overall, the HNF module, facilitates seamless information fusion at multiple scales, producing a cohesive representation of the micrographs. $<\hspace{-1mm}\textit{cls}\hspace{-1mm}>$ is the cls token and VN is the virtual node. $h^{l}_{i}$ and $e^{l}_{i}$ denotes the patch and node representation at layer $l$ of patch or node $i$, respectively.} 
\label{fig:figure2}
\vspace{-2mm}
\end{figure}

\vspace{-1mm}
The HNF architecture synergistically combines patch sequences and vision graphs representations at different scales, enhancing electron micrograph analysis by seamlessly integrating global insights through a multi-layered network structure. The layers are constructed by progressively increasing the patch size. Each layer of the network represents the original micrograph-based patch sequence and vision graph at different scales, with increasing resolutions. By considering information at multiple scales, the network offers a more comprehensive representation of the micrograph, capturing both fine-grained details and the global context. Figure \ref{fig:figure2} illustrates the Hierarchical Network Fusion (HNF) method. Each layer uses a bidirectional Neural ODE (refer to the appendix) to iteratively refine patch embeddings, facilitating the smooth, causal evolution of the patch embedding and capturing global inter-patch relationships and dependencies. It also incorporates a Graph Chebyshev Convolution Network (refer to the appendix) that maps the high-dimensional discrete vision graph information to low-dimensional node-level embeddings while optimally preserving the high-level visual features and structural information embedded in the graphs. Additionally, at each layer, the mixture-of-experts (MOE) technique employs a gating mechanism to combine predictions from the bidirectional Neural ODEs and the Graph Chebyshev Convolution methods. These predictions are integrated through a weighted sum of their $<\hspace{-1mm}\textit{cls}\hspace{-1mm}>$ token and virtual node embeddings. The training objectives include optimizing the weight distribution of the gating function for accurate classification of nanomaterial categories in electron micrographs and training the methods using the weights determined by the gating function. Overall, our framework aims to improve classification accuracy by leveraging the strengths of multiple learning methods and optimizing the weights of the gating mechanism, which serves as the bottleneck through which the two modalities interact to obtain the fused representation. In the subsequent layers, the fused information is combined with the individual modalities at higher patch resolutions. Our framework incorporates bidirectional Neural ODEs and Graph Chebyshev networks to facilitate the exchange of mutual information between patch sequences and visual graphs across multiple scales of patch size through the gating mechanism. This approach allows the patch embeddings to be grounded with structural and semantic information from the vision graph while enabling causal relations within the patch sequence to transform the graph embeddings. Overall, the framework fosters interactive knowledge integration between modalities within its architecture.

\vspace{-4mm}
\subsection{Beyond Conventional Analysis: Leveraging LLMs for Nanomaterial Characterization} 
\label{Languagemodels}
\vspace{-3mm}
The advent of large pre-trained language models (LLMs), such as OpenAI's ChatGPT \cite{brown2020language}, Google's PaLM \cite{chowdhery2022palm}, and Meta's LLaMA \cite{touvron2023llama}, has significantly revolutionized performance in various natural language processing tasks, achieving state-of-the-art results across a wide range of applications. In contrast, small-scale language models (LMs), such as BERT \cite{devlin2018bert} and DeBERTa \cite{he2020deberta}, lack the strong logical reasoning capabilities of LLMs and are limited in their ability to generate coherent and contextually relevant responses compared to larger models. However, small-scale LMs are computationally affordable for fine-tuning using labeled data for specialized task adaptation. In addition, they allow access to logits or token embeddings for downstream applications of smaller LMs across various tasks, aiding in explainability. Owing to their substantial model complexity and scale, general-purpose LLMs require significant computational resources for repurposing through fine-tuning for task-specific customization. Additionally, they do not provide access to latent token embeddings and logits, this black-box nature can limit the interpretability of LLMs. To overcome the challenges, the Language Modeling as a Service (LMaaS \cite{sun2022black}) platform provides access to LLMs via text-based API interaction through cloud-based services. However, the integration of LLMs with vision graphs remains an underexplored area, opening up the possibility for innovative techniques that combine language models and graph representation learning algorithms to improve nanomaterial identification applications. To address this, our approach capitalizes on zero-shot chain-of-thought (Zero-Shot CoT) prompting of LLMs to generate technical descriptions of nanomaterials. We pre-train smaller LMs on the generated textual descriptions using the masked language modeling (MLM) technique (i.e., pre-training for domain-customization) to learn expressive token embeddings for a better understanding of language structure and semantics. We then fine-tune smaller LMs for downstream supervised multi-class classification task (i.e., fine-tuning for task adaptation) to compute context-aware token embeddings. We employ weighted sum-pooling attention mechanisms to obtain contextualized text-level embeddings from token embeddings, which are used to perform inference in the nanomaterial identification task. Our work evaluates two LLMs: GPT-3.5-turbo, and Google BARD\footnote{https://bard.google.com}. GPT-3.5-turbo, a newer and larger extension of GPT-3.5 model from OpenAI, excels in various language tasks and shows cost-effectiveness, while Google BARD is significantly larger than GPT-3.5 models. We also utilize a pre-trained small-scale LM, DeBERTa\footnote{For more information, refer to the DeBERTa model documentation available at \url{https://huggingface.co/docs/transformers/index}.}\cite{he2020deberta}, which is an improved version of the BERT architecture. The technical details of these language models are given in Table \ref{model}. In the GPT-3.5-turbo and BARD, text generation diversity is mainly influenced by two parameters: Top-p (nucleus sampling) and temperature. Top-p sets a probability threshold for token inclusion, filtering out excessively rare or common tokens to balance the output. The temperature parameter dictates the randomness of generated text; high values foster creativity, while low values ensure focused and deterministic outputs. In our experiments, we set Top-p to 1 and temperature to 0 for accurate and controlled text generation.

\vspace{-6mm}
\begin{table}[ht!]
	\centering
	\small
	\caption{Technical specifications of the LLMs and LMs. The \emph{Cost} category indicates the price for using 1k tokens, while the \emph{Date of Last Update} category denotes the the most recent date the knowledge base of the LLMs was updated.}
	\vspace{0mm}
	\hspace{-5mm}\begin{tabular}{c|c|c|c|c}
            \toprule
            \textbf{Model} &\textbf{Organization} &\textbf{Cost}  &\textbf{Date of Last Update} &\textbf{Vocabulary Size}\\
            \hline
            ChatGPT &Open-AI &0.002\$ &Jun. 2021 & 175B \\
            BARD & Google &Free & Undisclosed & 1,560B \\
            \hline
            DeBERTa & Hugging Face & Free & N/A & 50M \\ \bottomrule
	\end{tabular}
	\vspace{-0.2cm}
	\label{model}
\end{table}

\vspace{-3mm}
\paragraph{Zero-Shot CoT LLMs Prompting:}
We access LLMs via the LMaaS platform, using text-based API interactions. We employ open-ended natural language prompts with task-specific instructions to query the LLMs, thereby generating detailed textual descriptions pertaining to the structure, properties, and applications of given nanomaterials. Utilizing a tailored zero-shot prompt template, we guide the LLMs through a series of chain-of-thought prompts\cite{wei2022chain}, extracting comprehensive domain knowledge embedded within the language model parameters to generate rich, detailed technical descriptions of nanomaterials. The customized CoT prompt format is as follows:

\vspace{-2mm}
\begin{tcolorbox}[colback=white!5!white,colframe=black!75!black]% [width=10cm]
\vspace{-2mm}
\textbf{Prompt 1:} Introduction: Provide an overview of the  nanomaterial category and its significance in various fields.
\textbf{Prompt 2:} Definition and Structure: Define the nanomaterial category and describe its typical structure at the nanoscale.
\textbf{Prompt 3:} Synthesis Methods: Explore different methods used to synthesize or fabricate nanomaterials in this category. Discuss their advantages and limitations.
\textbf{Prompt 4:} Properties: Highlight the unique physical, chemical, and electronic properties exhibited by nanomaterials in this category. Discuss how these properties differ from their bulk counterparts.
\textbf{Prompt 5:} Applications: Explore the wide range of applications where nanomaterials in this category are utilized. Discuss their potential impact in fields such as electronics, energy, medicine, environmental remediation, etc.
\textbf{Prompt 6:} Surface Modification: Describe the strategies used to modify the surface properties of nanomaterials in this category, such as functionalization, coating, or doping. Explain how these modifications enhance their performance or enable specific applications.
\textbf{Prompt 7:} Toxicity and Safety: Address the potential health and environmental concerns associated with nanomaterials in this category. Discuss studies on their toxicity, risk assessment, and safety measures to mitigate any potential hazards.
\textbf{Prompt 8:} Future Directions: Discuss current research trends and future prospects for nanomaterials in this category. Highlight emerging technologies, challenges, and areas of active exploration.
\vspace{-2mm}
\end{tcolorbox}

\vspace{-1mm}
Querying the LLMs generates technical descriptions of nanomaterial categories. It provides valuable insights into the characteristics, properties, and applications of different types of nanomaterials.

\vspace{-1mm}
\begin{tcolorbox}[colback=white!5!white,colframe=black!75!black]%[width=10cm]
\vspace{-1.5mm}
\centering
(\textbf{LLMs Response}) [Textual Outputs]
\vspace{-1.5mm}
\end{tcolorbox}

\vspace{-1mm}
In the following section, we will present our approach to integrating detailed textual descriptions into a small-scale LM for pre-training through the masked language modeling (MLM) technique, and fine-tuning for domain customization on the downstream supervised nanomaterial identification task.

\vspace{-3mm}
\paragraph{Domain Customization: Fine-Tuning LMs}
\vspace{-1mm}
Our approach employs a smaller language model (LM) to interpret and encode the textual outputs generated by a larger language model (LLM). We leverage the smaller LM as an intermediate network to bridge the LLMs and downstream classification layers. The encoder-only LMs\cite{pan2023unifying} are fine-tuned using a self-supervised learning approach known as masked language modeling (MLM). In this approach, the large corpus of LLM textual outputs is processed by randomly masking out tokens in each sentence. The model is then trained to predict the masked words, given the context of the surrounding non-masked words. This process helps the model learn the statistical relationships between words and phrases, thereby facilitating the generation of coherent language representations. Briefly, we pre-train smaller general-purpose language models (referred to as $\textrm{LM}_{\textrm{expl}}$) using the MLM technique for domain customization, enhancing language-based contextual understanding and semantic relationship extraction for aiding downstream applications. We then fine-tune the smaller LM for downstream task-specific adaptation to encapsulate the explanations generated by LLMs. Post pre-training on MLM technique, we input the text sequences generated by LLMs (denoted as $\mathcal{S}_\textrm{expl}$) into the $\textrm{LM}_{\textrm{expl}}$ model, which then generates expressive, context-aware embeddings for each token in the sentence, capturing the semantic relationships between the tokens as follows:

\vspace{-6mm}
\resizebox{0.965\linewidth}{!}{
\hspace{0mm}\begin{minipage}{\linewidth}
\begin{equation}
h_{\textrm{expl}} = \textrm{LM}_\textrm{expl}(s_{\textrm{expl}})
\end{equation}
\end{minipage}
} 

\vspace{-2mm}
where the context-aware embeddings are denoted as $h_{\textrm{expl}} \in \mathbb{R}^{\hspace{0.5mm}m \times d}$, where $m$ represents the number of tokens in $\mathcal{S}_{\textrm{expl}}$ and $d$ is token embedding dimension. We then perform sum-pooling attention mechanism to compute a weighted sum of these token embeddings to encode the textual explanations to obtain an text-level fixed-length embedding as follows:

\vspace{-2mm}
\resizebox{0.965\linewidth}{!}{
\hspace{0mm}\begin{minipage}{\linewidth}
\begin{equation}
\alpha_i = \mbox{softmax}(q_i); \hspace{2mm} q_i = \mathbf{u}^Th^{(i)}_{\textrm{expl}}
\end{equation}
\end{minipage}
} 

\vspace{-2mm}
\resizebox{0.915\linewidth}{!}{
\hspace{0mm}\begin{minipage}{\linewidth}
\begin{equation}
h^{\text{text}} = \sum_{i=0}^m{\alpha_i h^{(i)}_{\textrm{expl}}}
\end{equation}
\end{minipage}
} 

\vspace{-2mm}
where $\mathbf{u}$ is a differentiable vector. The text-level embedding $h^{\text{text}}\in \mathbb{R}^{d}$ captures the essence or core of the domain knowledge as a whole, extracted from the foundational LLMs for each nanomaterial. We calculate the relevance score between the text-level embedding($h^{\textit{text}}$) and the electron micrograph representations($ h_{\textrm{fus}}$) obtained from the hierarchical network fusion(HNF, refer to section \ref{HNF}), as detailed below,

\vspace{-5mm}
\resizebox{1\linewidth}{!}{
\hspace{0mm}\begin{minipage}{\linewidth}
\begin{equation}
    \beta^{*} = \arg \max_{c} [\mbox{softmax}(q_k h_{\textrm{fus}})]; \hspace{2mm} q_k = \mathbf{v}^T[h^{\textit{text}}_{1} || \cdot || h^{\textit{text}}_{c}] 
\end{equation}
\end{minipage}
} 

\vspace{-2mm}
where the subscript, $c$ denotes the the total number of nanomaterial categories and $\mathbf{v}$ is a trainable parameter. The above operator computes the list of scores or probabilities for each nanomaterial, and the $\arg \max$ operator selects the nanomaterial for which the probability score is maximized. We then select the appropriate/relevant nanomaterial text-level embedding conditioned on hierarchical embedding ($h_{\textrm{fus}}$) as follows:

\vspace{-6mm}
\resizebox{1\linewidth}{!}{
\hspace{0mm}\begin{minipage}{\linewidth}
\begin{equation}
    h^{\textit{text}}_{\textrm{fus}} = h^{\textit{text}}_{\beta^{*}} 
\end{equation}
\end{minipage}
} 

\vspace{-1mm}
$\beta^{*}$ denotes the nanomaterial label with the highest probability. \textcolor{black}{This is essentially a matching mechanism that tries to find the best pairwise alignment among the various nanomaterial text-level embeddings ($h^{\textit{text}}_{1}, \ldots, h^{\textit{text}}_{c}$) and the hierarchical embedding ($h_{\textrm{fus}}$) obtained from the hierarchical network fusion (HNF)}. We utilize backpropagation error in the downstream supervised multi-classification task to fine-tune the smaller LMs to maximize the pairwise alignment between the complementary hierarchical embedding ($h_{\textrm{fus}}$) and its corresponding text-level embedding $h^{\textit{text}}_{\textrm{fus}}$. To put it briefly, $h^{\textit{text}}_{\textrm{fus}}$ incorporates the expert knowledge obtained from foundational LLMs for the appropriate nanomaterial underlying the electron micrographs.

\vspace{-4mm}
\subsection{Overall Method}
\vspace{-3mm}
Figure \ref{fig:figure3} provides an overview of the ``MultiFusion-LLM'' framework. Our proposed framework comprises three distinct methods: a) \textbf{Hierarchical Network Fusion (HNF)} tokenizes micrographs into patches to obtain patch sequences and construct vision graphs. It introduces a $<\hspace{-1mm}\textit{cls}\hspace{-1mm}>$ token into the patch sequence and a virtual node for the vision graph to capture global characteristics. The network has a multi-layered structure; each layer of the network consists of bidirectional Neural ODEs and graph Chebyshev networks, and regulates the information flow through a gating mechanism to learn hierarchical embeddings with increasing patch sizes across each layer. It computes cross-modal embeddings, denoted as $\mathbf{h}_{fus}$, by integrating embeddings between modalities at different patch resolutions, thereby facilitating the exchange of information and integration of knowledge. For more detailed information, please refer to section \ref{HNF}. b)  \textbf{LLMs for Incorporating Domain Knowledge:} We generate technical descriptions of nanomaterials, capturing a wide range of information including structure, properties, and applications using Zero-Shot CoT prompting of LLMs. To illustrate, Table \ref{tab:lmprompts} provides a glimpse of the LLM-retrieved text obtained from GPT-3.5 turbo, specifically generated to address natural language queries regarding MEMS devices. Initially, we pre-train a smaller LM on the generated descriptions through masked language modeling (MLM). Later, we fine-tune this small-scale LM on a downstream supervised task to encapsulate the generated explanations. We then utilize the weighted sum-pooling attention mechanism to compute domain-specific knowledge-incorporated text-level embeddings, denoted as $\mathbf{h}^\text{text}_{fus}$. For additional details, please refer to subsection \ref{Languagemodels}. (c) We employ the \textbf{multi-head attention mechanism (MHA)}\cite{vaswani2017attention} to fuse text-level embeddings $\mathbf{h}^\text{text}_{fus}$ with hierarchical embeddings $\mathbf{h}_{fus}$, enabling the capture of contextually relevant information and achieving semantic alignment across different cross-domain embeddings. Simultaneously, by focusing on and aligning high-level textual descriptions (text-level embeddings) with detailed visual representations (hierarchical embeddings), we ensure a comprehensive understanding and analysis of electron micrographs from both descriptive and visual perspectives. This approach helps mitigate the inherent limitations arising from high intra-class dissimilarity, high inter-class similarity, and spatial heterogeneity in visual patterns across the electron micrographs, ultimately enhancing the performance of nanomaterial identification tasks. We compute the Query, Key, Value projections for 

\newpage
the text-level embedding $\mathbf{h}^\text{text}_{fus}$ for each head h as follows:

\vspace{-5mm}
\resizebox{0.92\linewidth}{!}{
\hspace{0mm}\begin{minipage}{\linewidth}
\begin{align}
Q^h_{\text{text}} &= \mathbf{h}^\text{text}_{fus}W^h_{Q_\text{text}}; K^h_{\text{text}} = \mathbf{h}^\text{text}_{fus} W^h_{K_\text{text}}; V^h_{\text{text}} = \mathbf{h}^\text{text}_{fus} W^h_{V_\text{text}}
\end{align}
\end{minipage}
} 

\vspace{-1mm}
Similarly, the Query, Key, Value projections for hierarchical embedding $\mathbf{h}_{fus}$ for each head $h$ as follows:

\vspace{-7mm}
\resizebox{0.92\linewidth}{!}{
\hspace{0mm}\begin{minipage}{\linewidth}
\begin{align}
Q^h_{\text{fus}} &= \mathbf{h}_{fus} W^h_{Q_{\text{fus}}}; K^h_{\text{fus}} = \mathbf{h}_{fus} W^h_{K_{\text{fus}}}; V^h_{\text{fus}} = \mathbf{h}_{fus} W^h_{V_{\text{fus}}}
\end{align}
\end{minipage}
} 

\vspace{-1mm}
We concatenate keys and values of text-level and hierarchical embeddings to create a unified representation.

\vspace{-7mm}
\resizebox{0.92\linewidth}{!}{
\hspace{3mm}\begin{minipage}{\linewidth}
\begin{align}
K^h_{\text{concat}} &= [K^h_{\text{text}}, K^h_{\text{fus}}]; V^h_{\text{concat}} = [V^h_{\text{text}}, V^h_{\text{fus}}]
\end{align}
\end{minipage}
} 

\vspace{-1mm}
We apply Softmax attention to integrate complementary information from the cross-domain embeddings, focusing on relevant information and aligning them semantically.

\vspace{-4mm}
\resizebox{0.82\linewidth}{!}{
\hspace{0mm}\begin{minipage}{\linewidth}
\begin{align}
A^h_{\text{cross}} &= \text{Softmax}\left(\frac{(Q^h_{\text{text}} + Q^h_{\text{fus}}) {K^h_{\text{concat}}}^T}{\sqrt{d_h}}\right)
\end{align}
\end{minipage}
} 

\vspace{-2mm}
Each head outputs a new vector representation that highlights the most relevant features in the mono-domain embeddings, tailored to specific aspects of the data.

\vspace{-4mm}
\resizebox{0.92\linewidth}{!}{
\hspace{0mm}\begin{minipage}{\linewidth}
\begin{align}
O^h_{\text{cross}} &= A^h_{\text{cross}} V^h_{\text{concat}}
\end{align}
\end{minipage}
} 

\vspace{-1mm}
Finally, we concatenate and linearly transform all head-specific outputs to create the final unified cross-modal embedding.

\vspace{-8mm}
\resizebox{0.92\linewidth}{!}{
\hspace{0mm}\begin{minipage}{\linewidth}
\begin{align}
O_{\text{concat}} &= [O^1_{\text{cross}}, O^2_{\text{cross}}, \ldots, O^H_{\text{cross}}] \\
y_{\text{cross}} &= O_{\text{concat}} W_{O_{\text{cross}}} \\
\text{p}_{i} &= \text{softmax}\big( \text{W}y_{\text{cross}} \big)   
\end{align}
\end{minipage}
} 

\vspace{-1mm}
where $W^h_{Q_\text{text}}$, $W^h_{K_\text{text}}$, $W^h_{V_\text{text}}$, $W^h_{V_\text{fus}}$,  $W^h_{Q_{\text{fus}}}$,  $W^h_{K_{\text{fus}}}$, $W_{O_{\text{cross}}}$ and $\text{W}$ are the trainable weight matrices. $d_h$ represents the dimensionality of the key/query/value for each head, and $\text{H}$ is the number of heads.  $\text{p}_{i}$ represents the probability distribution across nanomaterial categories, we apply the argmax operation to $\text{p}_{i}$ to determine the framework's predictions for the nanomaterial category. In summary, we conduct Zero-shot CoT prompting of LLMs to generate technical descriptions of nanomaterials and pre-train small-scale LMs using masked language modeling (MLM). Next, we jointly optimize the smaller pre-trained LM and the hierarchical network fusion (HNF) method on supervised learning tasks. The objective is to minimize the cross-entropy loss and enhance multi-class classification accuracy. In summary, the MHA offers a multi-faceted approach to capture and align varied information sources, making it a powerful tool for multi-modal data integration and analysis. It allows for a robust, synergistic, and comprehensive representation of data, especially in contexts like nanomaterial analysis where both modalities offer complementary insights.

\vspace{-5mm} 
\section{Experiments And Results}

\vspace{-4mm}
\subsection{Datasets}
\begin{figure}[htbp]
\centering
     \subfloat{\hspace{-0mm}\includegraphics[width=0.16\textwidth]{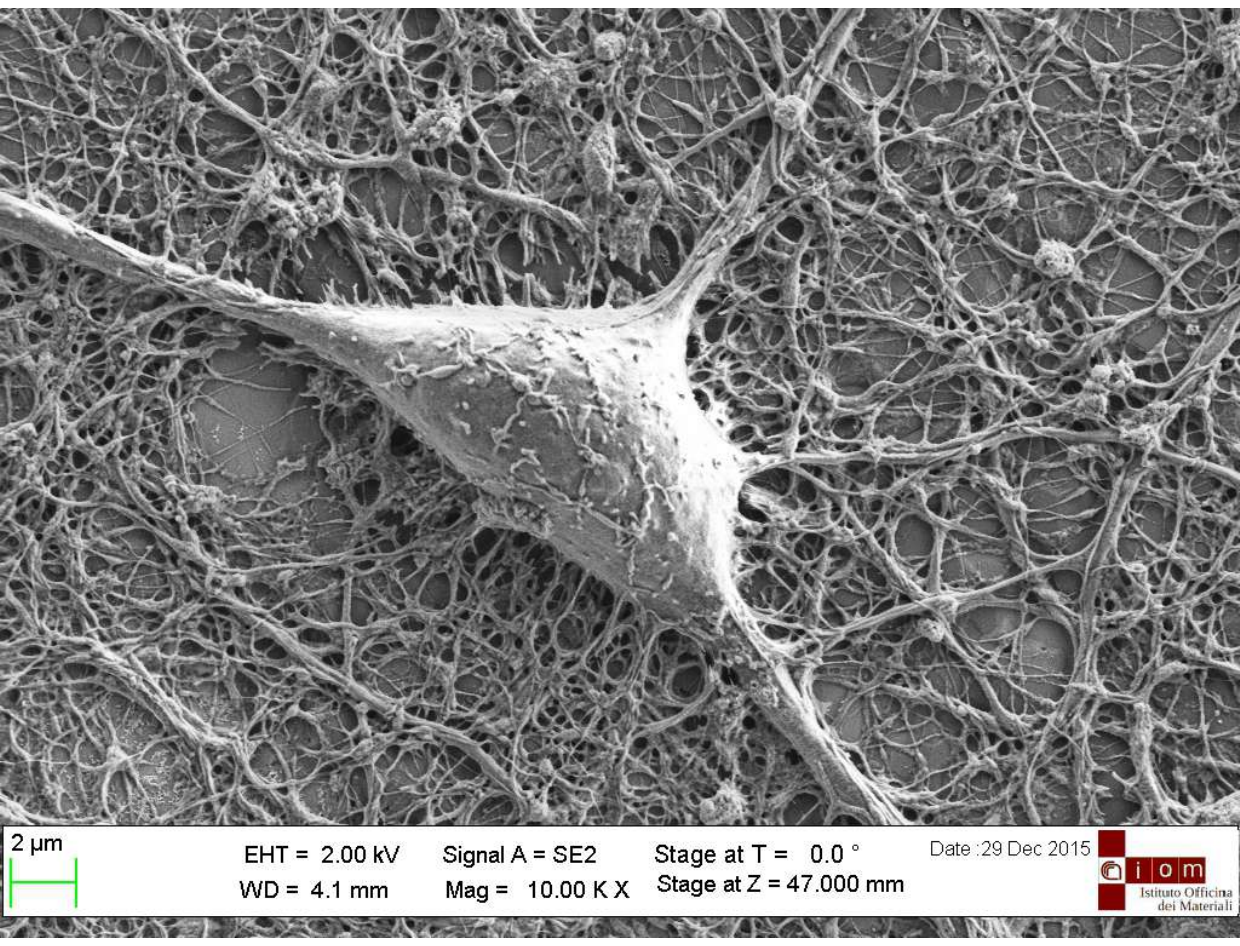}
     \includegraphics[width=0.16\textwidth]{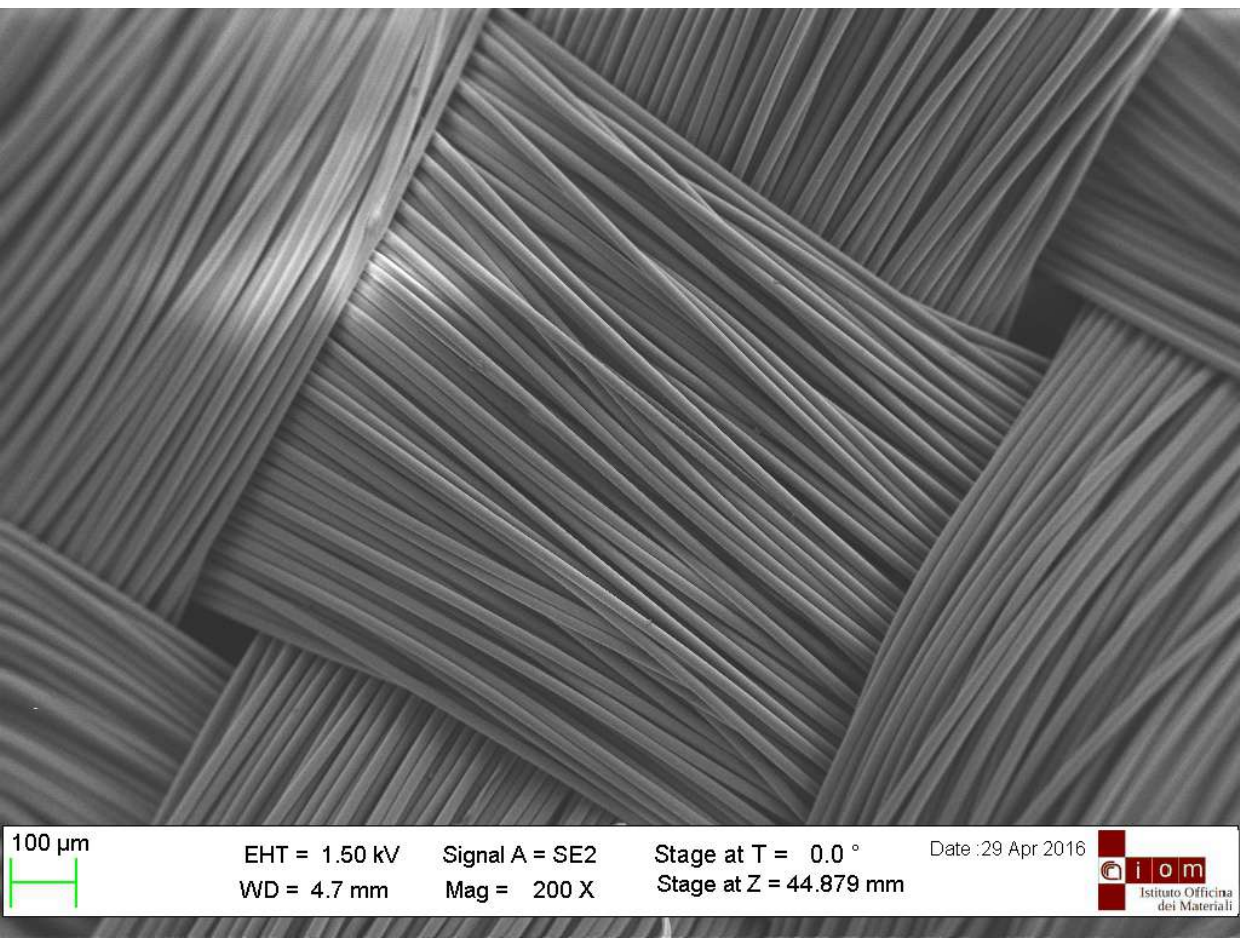}
     \includegraphics[width=0.16\textwidth]{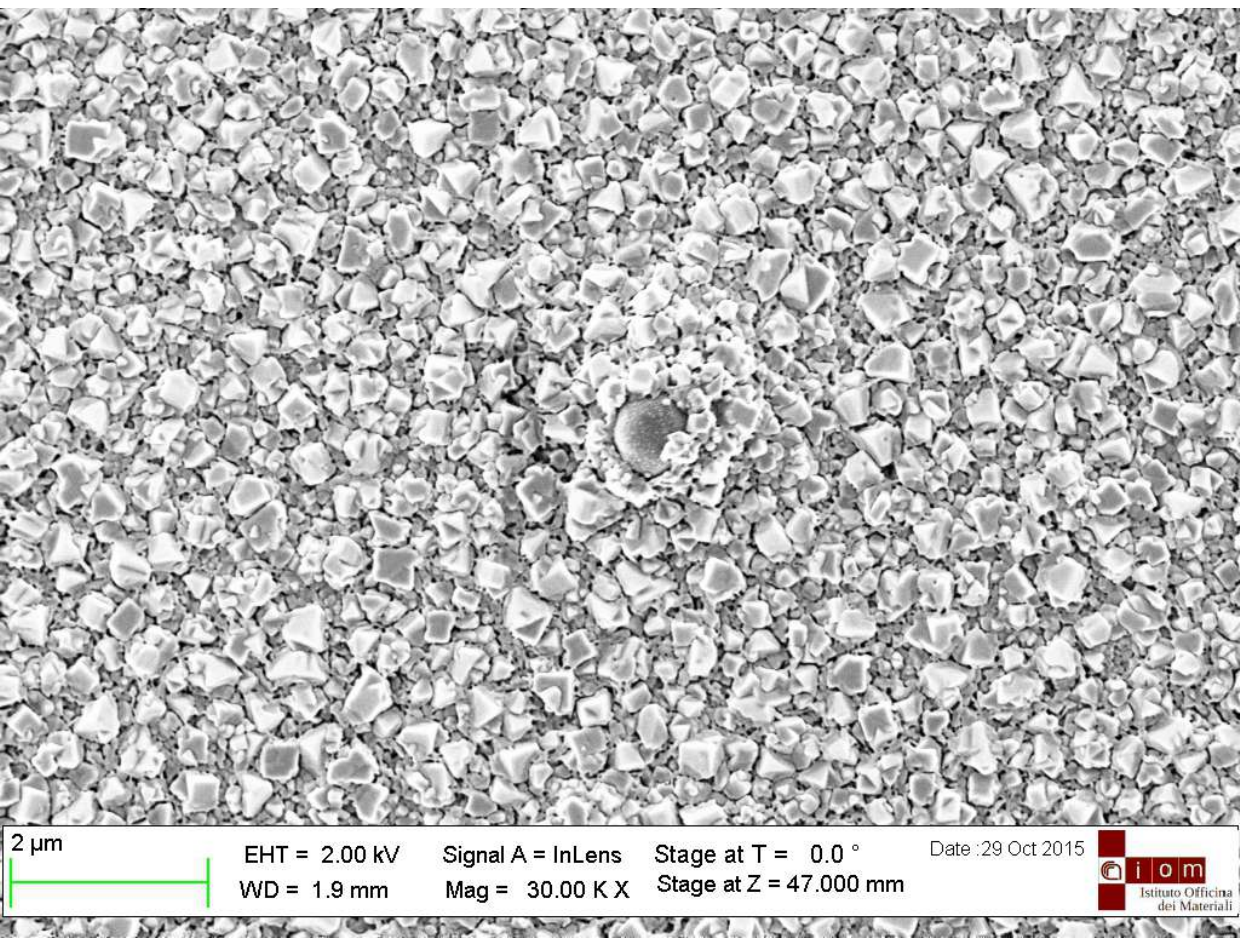}
     \includegraphics[width=0.16\textwidth]{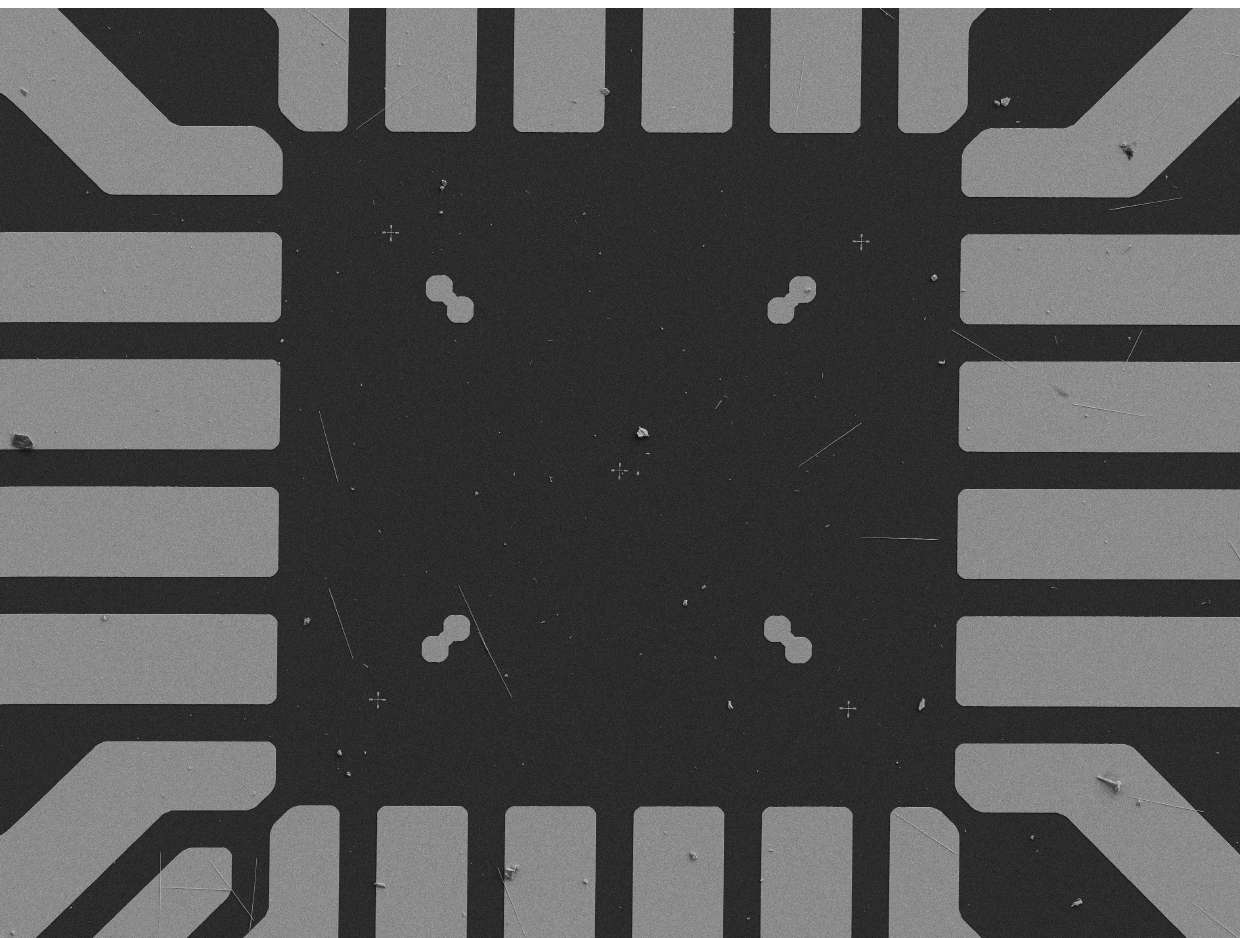}
     \includegraphics[width=0.16\textwidth]{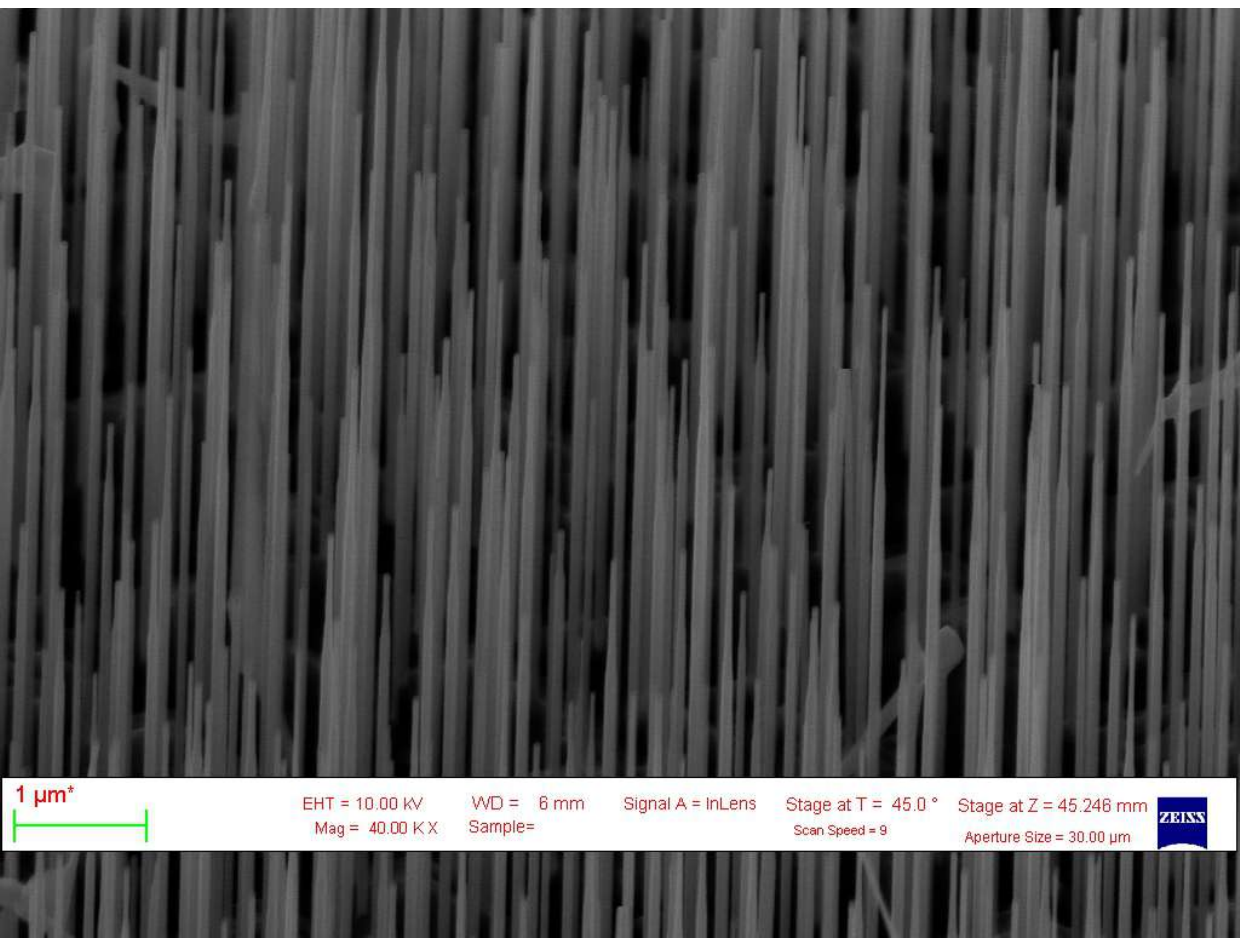}
     }
     \vspace{-18mm}
     \qquad
     \subfloat{\hspace{-0mm}\includegraphics[width=0.16\textwidth]{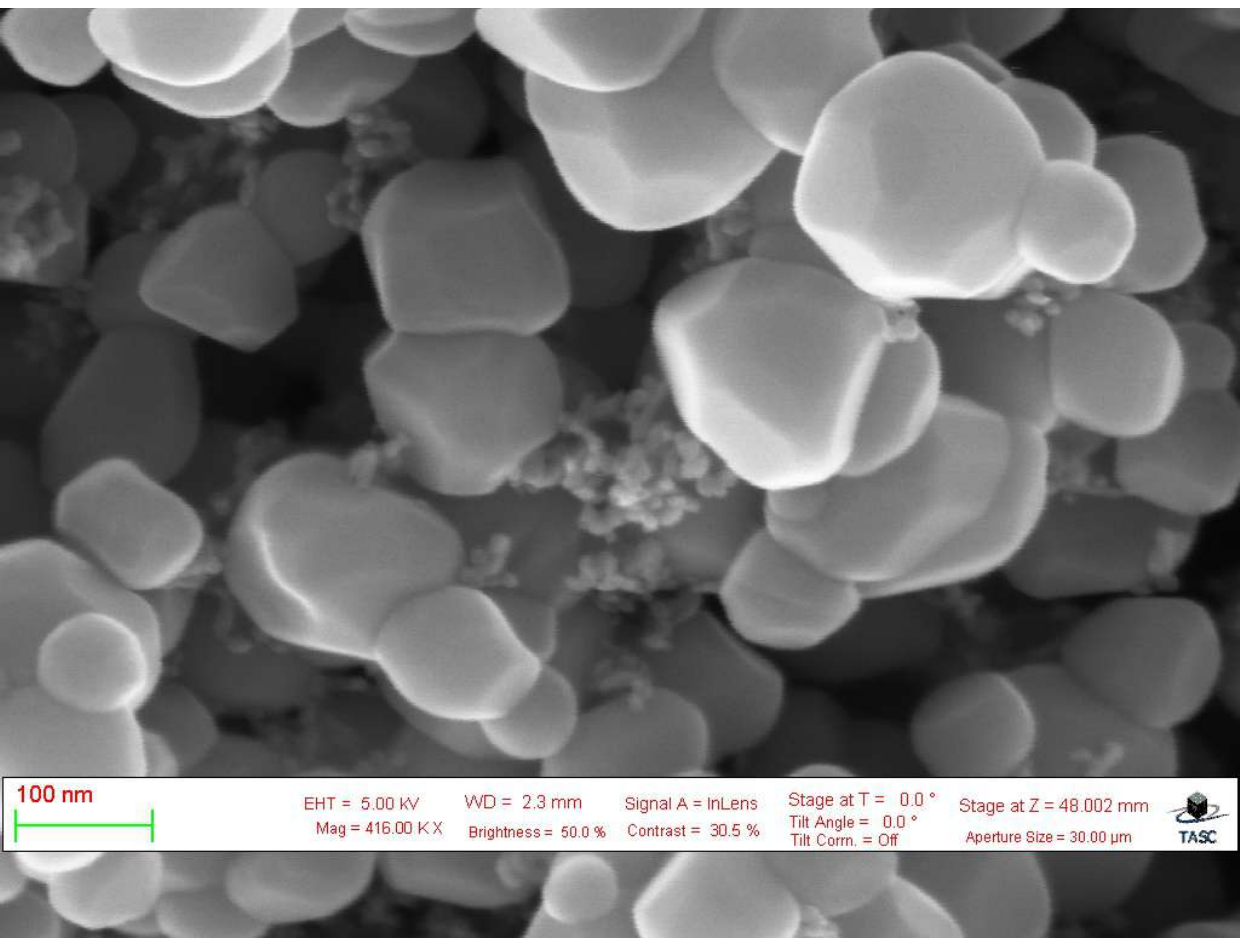}
     \includegraphics[width=0.16\textwidth]{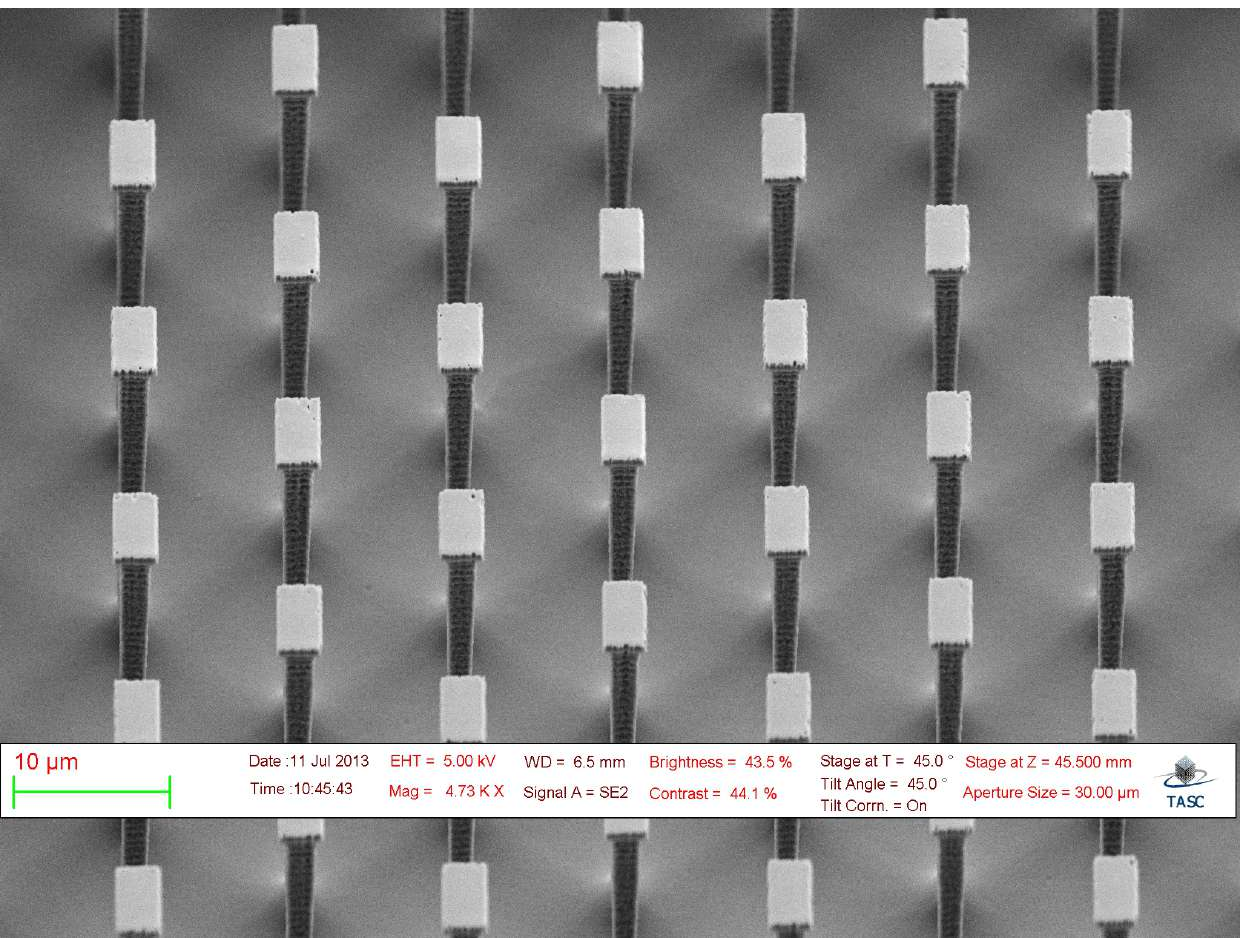}
     \includegraphics[width=0.16\textwidth]{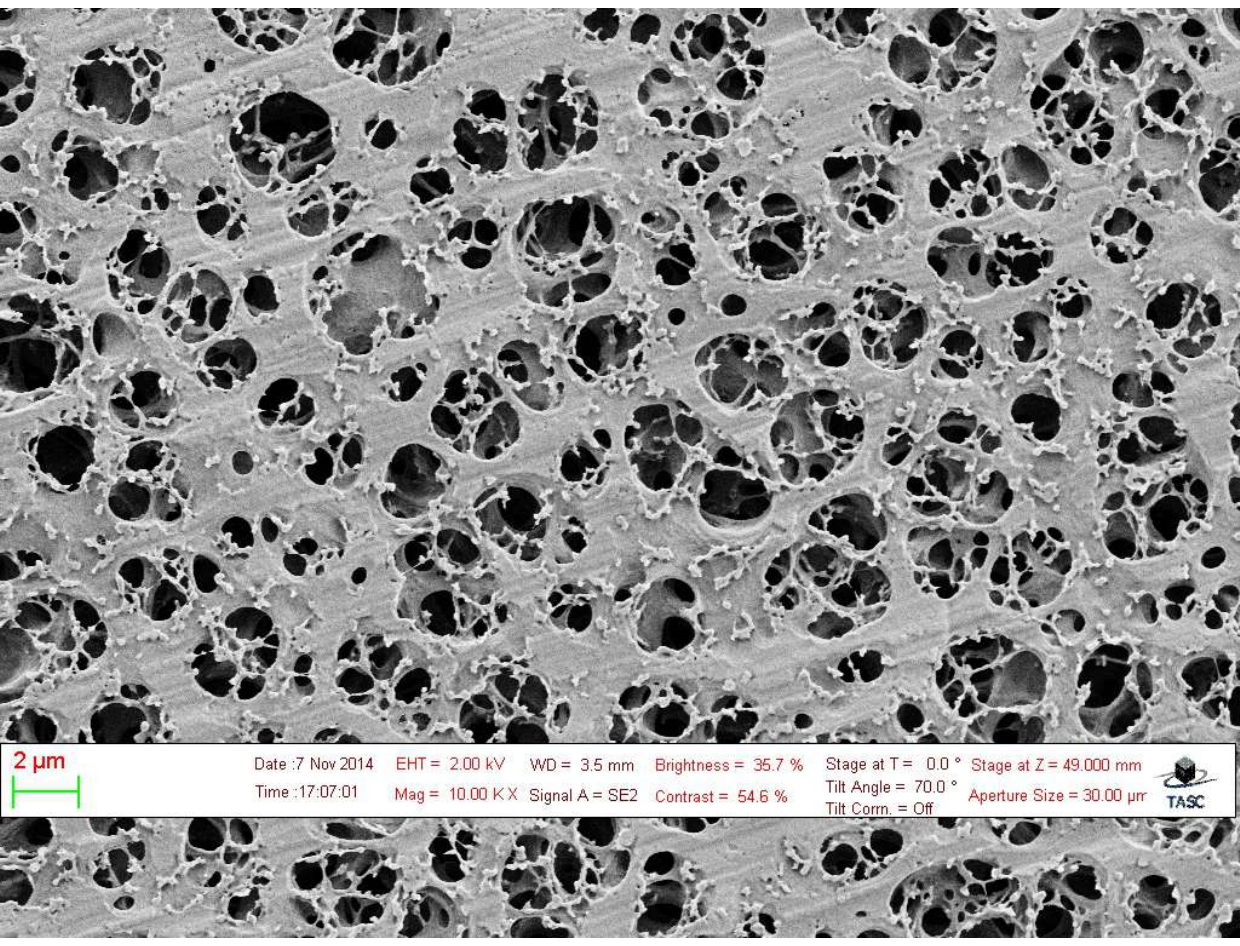}
     \includegraphics[width=0.16\textwidth]{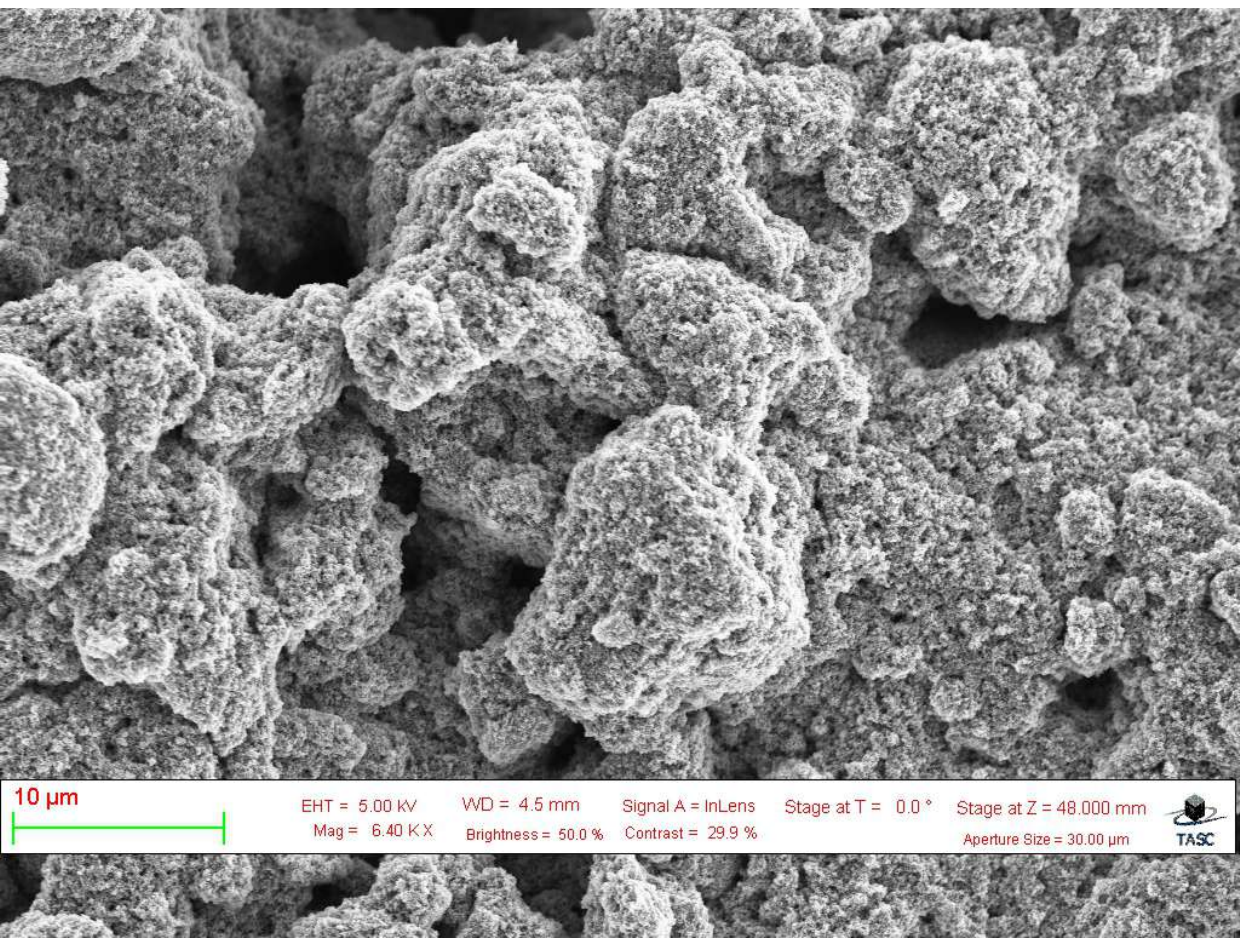}
     \includegraphics[width=0.16\textwidth]{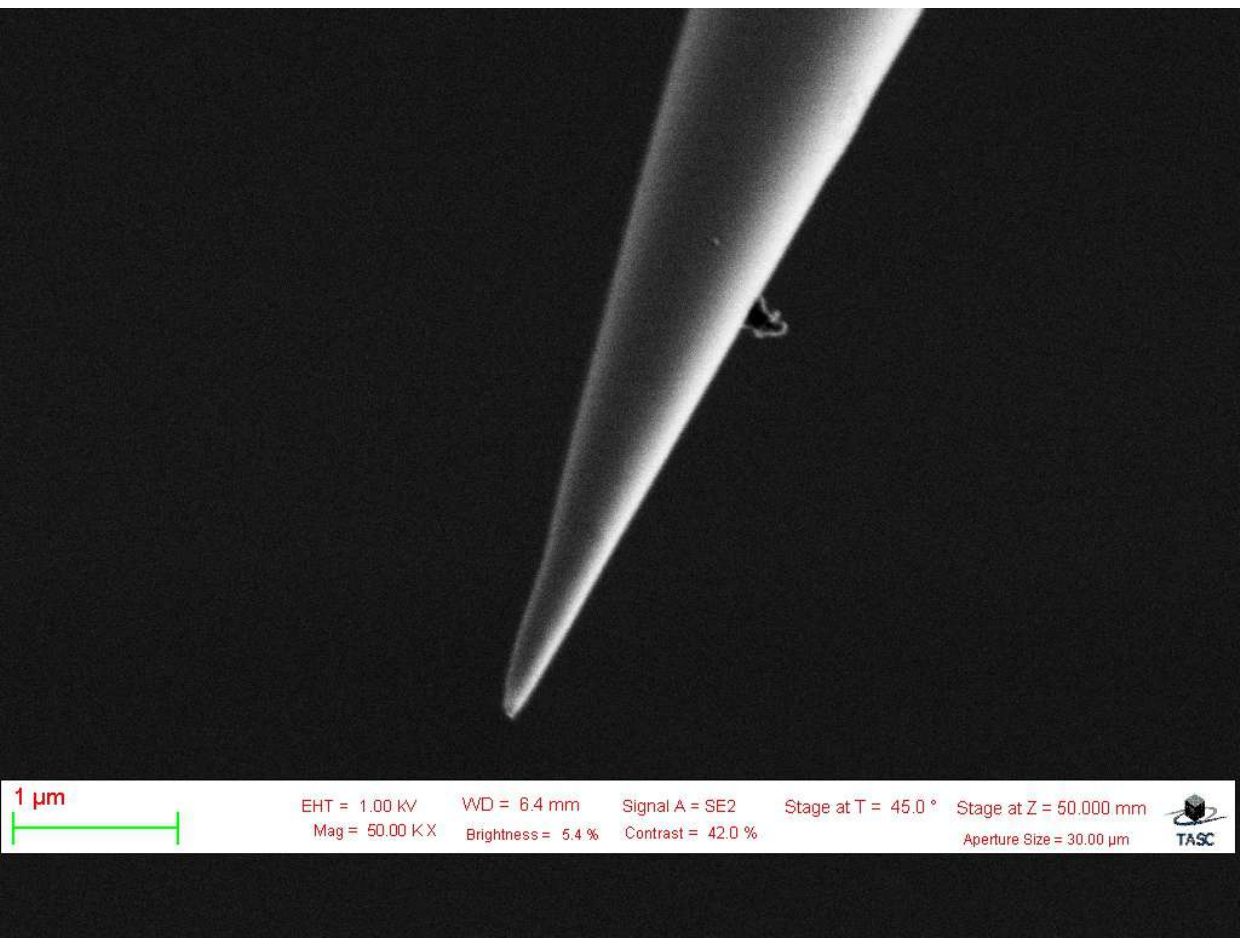}
     }
     \vspace{-9mm}
     \caption{The figure depicts the different types of nanomaterials found in the SEM dataset (\cite{aversa2018first}) (left to right in the first row: \textit{biological, fibers, films, MEMS, nanowires}; left to right in the second row: \textit{particles, patterned surface, porous sponges, powder, tips}).}
      \vspace{0mm}
     \label{fig:illustrationpics}
\end{figure}

\vspace{-3mm}
Our study primarily utilized the SEM dataset\cite{aversa2018first} to automate nanomaterial identification. The expert-annotated dataset spans across 10 distinct categories, representing a broad range of nanomaterials such as \textit{particles, nanowires, patterned surfaces, among others}. In total, it contains approximately 21,283 electron micrographs. Figure \ref{fig:illustrationpics} provides a visual representation of the different nanomaterial categories included in the SEM dataset. Despite the initial findings by \cite{modarres2017neural} on a subset of the original dataset, our research was based on the complete dataset since the subset was not publicly accessible. Although the original dataset curators, \cite{aversa2018first}, did not provide predefined splits for training, validation, and testing, we utilized the k-fold cross-validation method to evaluate our framework's performance. This strategy facilitated a fair comparison with popular baseline models in a competitive benchmark setting. Furthermore, we extended our evaluation by leveraging several open-source material benchmark datasets relevant to our study. These datasets were used to showcase the efficacy of our proposed framework and its applicability in a broader context beyond the SEM dataset.

\vspace{-10mm}

\vspace{-3mm}
\subsection{Results}
\vspace{-3mm}
We evaluated the effectiveness of our proposed framework through a comprehensive performance analysis, comparing it to commonly used computer vision baseline models. Our comparisons included supervised learning models such as ConvNets and ViTs (as referenced in \cite{philvformer, neelayvformer}), along with self-supervised learning techniques like Vision Contrastive Learning (VCL, as discussed in \cite{susmelj2020lightly}). Table \ref{tab:table2} reports the experimental results from our study. To ensure a fair and rigorous comparison, we conducted experiments with consistent settings across all algorithms, measuring performance using the Top-$N$ accuracy metric and evaluating specifically for $N \in \{1, 2, 3, 5\}$. Our proposed framework outperforms the baseline models, showing a substantial relative improvement of $25.8\%$ in the Top-1 score and a marginal improvement of $5.34\%$ in the Top-5 score compared to the next-best baseline model, T2TViT (\cite{T2TViT}).

\vspace{-5mm}
\begin{table}[htbp]
\footnotesize
\centering
\setlength{\tabcolsep}{5pt}
\caption{The table shows the comparison of our proposed method with baseline algorithms, including vision-based supervised ConvNets, ViTs, and self-supervised learning (VSL) algorithms.}
\label{tab:table2}
\vspace{1mm}
\begin{tabular}{cc|c|c|c|c|c|c}
\hline
\multicolumn{2}{c|}{\textbf{Algorithms}}                      &\textbf{Parameters}                & \textbf{Top-1} & \textbf{Top-2} & \textbf{Top-3} & \textbf{Top-5}  \\ \hline
\multicolumn{1}{c|}{\multirow{6}{*}{\rotatebox[origin=c]{90}{\textbf{ConvNets}}}} & AlexNet(\cite{krizhevsky2017imagenet})      & 5.70E+07 &   0.493                &   0.582                &   0.673                &      0.793             &                     \\
\multicolumn{1}{c|}{}                                          & DenseNet(\cite{huang2017densely})     & 2.39E+05  & 0.539             & 0.750             & 0.875             & 0.906             &             \\
\multicolumn{1}{c|}{}                                          & ResNet(\cite{he2016deep})       & 2.72E+05  & 0.512             & 0.766             & 0.891             & 0.906             &             \\
\multicolumn{1}{c|}{}                                          & VGG(\cite{simonyan2014very})          & 3.44E+07  &  0.517                 & 0.644                  & 0.717                  &                  0.779 &                     \\
\multicolumn{1}{c|}{}                                          & GoogleNet(\cite{szegedy2015going})    & 2.61E+05  & 0.560             & 0.844             & 0.906             & 0.938              \\
\multicolumn{1}{c|}{}                                          & SqueezeNet(\cite{iandola2016squeezenet})   & 7.41E+05  & 0.436             & 0.469             & 0.609             & 0.656              \\ \hline
\multicolumn{1}{c|}{\multirow{6}{*}{\rotatebox[origin=c]{90}{\textbf{VSL}}}} & Barlowtwins\cite{zbontar2021barlow}  & 8.99E+06    & 0.138             & 0.250             & 0.328             & 0.453                         \\
\multicolumn{1}{c|}{}                                          & SimCLR\cite{chen2020simple}       & 8.73E+06    & 0.157             & 0.234             & 0.359             & 0.469                       \\
\multicolumn{1}{c|}{}                                          & byol\cite{grill2020bootstrap}         & 8.86E+06   & 0.130             & 0.234             & 0.281             & 0.422                         \\
\multicolumn{1}{c|}{}                                          & moco\cite{he2020momentum}         & 8.73E+06   & 0.158             & 0.188             & 0.250             & 0.438                         \\
\multicolumn{1}{c|}{}                                          & nnclr\cite{dwibedi2021little}        & 9.12E+06   & 0.144             & 0.266             & 0.313             & 0.531                         \\
\multicolumn{1}{c|}{}                                          & simsiam\cite{chen2021exploring}      & 9.01E+6   & 0.170             & 0.266             & 0.391             & 0.500                          \\ \hline
\multicolumn{1}{c|}{\multirow{24}{*}{\rotatebox[origin=c]{90}{\textbf{Vision Transformers(ViTs)}}}}        & CCT\cite{hassani2021escaping}          & 4.10E+05    & 0.600             & 0.781             & 0.875             & 0.969             &             \\
\multicolumn{1}{c|}{}                                          & CVT\cite{CVT}   & 2.56E+05    & 0.537             & 0.750             & 0.828             & 0.953               \\
\multicolumn{1}{c|}{}                                          & ConViT\cite{ConViT}       & 6.00E+05    & 0.582             & 0.734             & 0.828             & 0.938                         \\
\multicolumn{1}{c|}{}                                          & ConvVT\cite{CVT}       & 9.23E+04   & 0.291             & 0.563             & 0.734             & 0.875              \\
\multicolumn{1}{c|}{}                                          & CrossViT\cite{Crossvit}     & 8.35E+05    & 0.466             & 0.719             & 0.828             & 0.938                      \\
\multicolumn{1}{c|}{}                                          & PVTC\cite{PVT}         & 1.30E+06    & 0.567             & 0.766             & 0.813             & 0.922             &            \\
\multicolumn{1}{c|}{}                                          & SwinT\cite{SwinT}        & 2.78E+07    & 0.675             & 0.766             & 0.891             & 0.938                         \\
\multicolumn{1}{c|}{}                                          & VanillaViT\cite{dosovitskiy2020image}   & 1.79E+06    & 0.623             & 0.828             & 0.859             & 0.938              \\
\multicolumn{1}{c|}{}                                          & Visformer\cite{visformer}    & 1.21E+05   & 0.371             & 0.578             & 0.641             & 0.797                         \\ 
\multicolumn{1}{c|}{}                                          & ATS\cite{fayyaz2021ats}          & 3.26E+06     & 0.511             & 0.703             & 0.828             & 0.938                       \\
\multicolumn{1}{c|}{}                                          & CaiT\cite{CaiT}         & 3.84E+07    & 0.616             & 0.750             & 0.906             & 0.938             &            \\
\multicolumn{1}{c|}{}                                          & DeepViT\cite{Deepvit}      & 3.26E+06    & 0.512             & 0.734             & 0.875             & 0.938                         \\
\multicolumn{1}{c|}{}                                          & Dino\cite{Dino}         & 2.02E+07    & 0.047             & 0.219             & 0.391             & 0.432             &             \\
\multicolumn{1}{c|}{}                                          & Distallation\cite{Distillation} & 2.06E+06  & 0.516             & 0.719             & 0.844             & 0.938                       \\
\multicolumn{1}{c|}{}                                          & LeViT\cite{Levit}        & 1.68E+07   & 0.597             & 0.813             & 0.875             & 0.953                         \\
\multicolumn{1}{c|}{}                                          & MA\cite{MA}           & 3.87E+06    & 0.192             & 0.288             & 0.350             & 0.459             &            \\
\multicolumn{1}{c|}{}                                          & NesT\cite{Nest}         & 1.61E+07      & 0.636             & 0.828             & 0.891             & 0.953             &             \\
\multicolumn{1}{c|}{}                                          & PatchMerger\cite{PatchMerger}  & 3.26E+06    & 0.549             & 0.719             & 0.859             & 0.922                         \\
\multicolumn{1}{c|}{}                                          & PiT\cite{PiT}          & 4.48E+06     & 0.520             & 0.703             & 0.828             & 0.953             &            \\
\multicolumn{1}{c|}{}                                          & RegionViT\cite{Regionvit}    & 1.22E+07    & 0.575             & 0.797             & 0.859             & 0.922                         \\
\multicolumn{1}{c|}{}                                          & SMIM\cite{SMIM}         & 2.38E+06     & 0.163             & 0.297             & 0.453             & 0.609             &             \\
\multicolumn{1}{c|}{}                                          & T2TViT\cite{T2TViT}      & 1.03E+07    & 0.702             & 0.859             & 0.906             & 0.938                         \\
\multicolumn{1}{c|}{}                                          & ViT-SD\cite{ViT-SD}          & 4.47E+06     & 0.613             & 0.766             & 0.906             & 0.953                         \\ 
\hline
\multicolumn{1}{c|}{}                                          & \textbf{MultiFusion-LLM W/GPT-3.5}      &  2.39E+07     &    \textbf{0.947}               &     \textbf{0.965}              &      \textbf{0.986}             &    \textbf{0.991}                     \\ \hline
\multicolumn{1}{c|}{}                                          & \textbf{MultiFusion-LLM W/Google Bard}      &  2.39E+07     &    \underline{0.852}               &     \underline{0.899}              &      \underline{0.927}             &    \underline{0.953}                     \\ \hline
\end{tabular}
\end{table}

\vspace{-7mm}
\section{Conclusion}
\vspace{-4mm}
To conclude, we have conducted the first in-depth study aimed at achieving state-of-the-art performance in nanomaterial characterization. This study introduces the innovative \texttt{MultiFusion-LLM} framework, a robust solution to the challenges associated with nanomaterial identification in electron micrographs. By synergistically integrating multi-modal representations and leveraging the analytical prowess of large language models, it promises more nuanced and accurate classification. Our comprehensive framework has outperformed traditional methods, showcasing cutting-edge performance on cost-efficient GPU hardware. Furthermore, it has demonstrated effectiveness and computational efficiency, particularly with large datasets, thereby accelerating high-throughput screening and advancing research holding implications for the advancement of the semiconductor industries.

\pagebreak

\newpage

\section{Technical Appendix}
\vspace{-4mm}
Table \ref{tab:table3} presents experimental findings comparing the proposed framework's performance to various supervised learning-based baseline models, including several GNN architectures (\cite{rozemberczki2021pytorch, Fey/Lenssen/2019}), and we use Graph Contrastive Learning (GCL, \cite{Zhu:2021tu}) algorithms for additional comparison. Our proposed framework achieves SOTA performance on the benchmark dataset \cite{aversa2018first} compared to the baselines.
 
\vspace{-5mm}
\begin{table}[htbp]
\footnotesize
\centering
\setlength{\tabcolsep}{4pt}
\caption{The table presents the results of a comparative study between our proposed method and supervised-learning based GNNs, as well as self-supervised graph contrastive learning (GCL) algorithms, on the SEM dataset \cite{aversa2018first}.}
\label{tab:table3}
\vspace{0.5mm}
\begin{tabular}{cc|c|c|c|c|c|c}
\hline
\multicolumn{2}{c|}{\textbf{Algorithms}}                           &\textbf{Parameters}           & \textbf{Top-1} & \textbf{Top-2} & \textbf{Top-3} & \textbf{Top-5}  \\ \hline
\multicolumn{1}{c|}{\multirow{4}{*}{\rotatebox[origin=c]{90}{\textbf{GSL}}}} & GBT\cite{bielak2021graph}    & 7.09E+05     & 0.513             & 0.595             & 0.686             & 0.778                        \\
\multicolumn{1}{c|}{}                                          & GRACE\cite{zhu2020deep}         & 7.44E+05    & 0.581             & 0.646             & 0.711             & 0.773              \\
\multicolumn{1}{c|}{}                                          & BGRL\cite{thakoor2021bootstrapped}          & 6.92E+05    & 0.573             & 0.629             & 0.671             & 0.728                       \\
\multicolumn{1}{c|}{}                                          & InfoGraph\cite{sun2019infograph}        & 6.82E+05    & 0.560             & 0.631             & 0.694             & 0.756                 \\
\hline
\multicolumn{1}{c|}{\multirow{15}{*}{\rotatebox[origin=c]{90}{\textbf{Graph Neural Networks}}}}        & APPNP\cite{klicpera2018predict}           & 7.35E+05    & 0.604             & 0.713             & 0.792             & 0.823                                                \\
\multicolumn{1}{c|}{}                                          & AGNN\cite{thekumparampil2018attention}  & 5.22E+05    & 0.517             & 0.733             & 0.841             & 0.943                        \\
\multicolumn{1}{c|}{}                                          & ARMA\cite{bianchi2021graph}       & 4.57E+05    & 0.553             & 0.747             & 0.848             & 0.925                         \\
\multicolumn{1}{c|}{}                                          & DNA\cite{fey2019just}        & 8.48E+05    & 0.593             & 0.677             & 0.786             & 0.891                      \\
\multicolumn{1}{c|}{}                                          & GAT\cite{velivckovic2017graph}      & 6.31E+05    & 0.507             & 0.724             & 0.807             & 0.914                         \\
\multicolumn{1}{c|}{}                                          & GGConv\cite{li2015gated}          & 8.05E+05    & 0.583             & 0.778             & 0.841             & 0.944                         \\
\multicolumn{1}{c|}{}                                          & GraphConv\cite{morris2019weisfeiler}         & 5.85E+05     & 0.623             & 0.787             & 0.875             & 0.953                         \\
\multicolumn{1}{c|}{}                                          & GCN2Conv\cite{chen}   & 6.18E+05    & 0.697             & 0.813             & 0.867             & 0.945             &            \\
\multicolumn{1}{c|}{}                                          & ChebConv\cite{defferrard2016convolutional}     & 5.00E+05    & 0.547             & 0.762             & 0.834 & 0.896                                                \\ 
\multicolumn{1}{c|}{}                                          & GraphConv\cite{morris2019weisfeiler}           & 6.79E+05    & 0.533             & 0.727             & 0.847             & 0.961                         \\
\multicolumn{1}{c|}{}                                          & GraphUNet\cite{gao2019graph}        & 9.57E+05    & 0.622             & 0.738             & 0.866             & 0.912                         \\
\multicolumn{1}{c|}{}                                          & MPNN\cite{gilmer2017neural}       & 5.22E+05    & 0.643             & 0.792             & 0.873             & 0.959                         \\
\multicolumn{1}{c|}{}                                          & RGGConv\cite{bresson2017residual}          & 6.58E+05    & 0.633             & 0.727             & 0.886             & 0.928                        \\
\multicolumn{1}{c|}{}                                          & SuperGAT\cite{kim2022find}  & 5.54E+05    & 0.561             & 0.676             & 0.863             & 0.935                         \\
\multicolumn{1}{c|}{}                                          & TAGConv\cite{du2017topology}         & 5.74E+05    & 0.614             & 0.739             & 0.803             & 0.946                         \\
\hline
\multicolumn{1}{c|}{}                                          & \textbf{MultiFusion-LLM W/GPT-3.5}      &  2.39E+07     &    \textbf{0.947}               &     \textbf{0.965}              &      \textbf{0.986}             &    \textbf{0.991}                     \\ \hline
\multicolumn{1}{c|}{}                                          & \textbf{MultiFusion-LLM W/Google Bard}      &  2.39E+07     &    \underline{0.852}               &     \underline{0.899}              &      \underline{0.927}             &    \underline{0.953}                     \\ \hline
\end{tabular}
\vspace{-2mm}
\end{table}

\vspace{-3mm}
\subsection{Experimental Setup}
\vspace{-2mm}
The SEM dataset\cite{aversa2018first} consists of electron micrographs with dimensions of $1024\times 768\times 3$ pixels. To facilitate our analysis, we downscale these micrographs to $224\times 224\times 3$ pixels. As part of the data preprocessing, we normalize the electron micrographs by adjusting the mean and covariance to achieve a value of 0.5 across all channels. This normalization results in the micrographs falling within the range of [-1, 1]. We tokenize the downscaled and normalized micrographs into discrete, non-overlapping patches. Subsequently, we represent the electron micrographs as patch sequences and construct vision graphs using the Top-K nearest neighbor search algorithm. Specifically, we set the value of K to 10, 6, and 4 for each layer in the hierarchical network fusion (HNF) method, resulting in a total of three layers. This process generates multi-scale vision graphs and patch sequences with patch resolutions increasing of 16, 28, and 32 pixels. The patch dimension ($d_\text{pos}$) and position embedding dimension ($d$) are both set to 64. The framework is evaluated using a 10-fold cross-validation strategy and trained for 50 epochs with an initial learning rate of $1e^{-3}$ and a batch size of 48. We have a few more hyperparameters set for the cross-modal attention layer with the number of attention heads(H) to 4, and the dimensionality of Key/Query/Value ($d_{h}$) is 16. To enhance the performance of the \texttt{MultiFusion-LLM} framework, we employ two key strategies: (a) early stopping on the validation set, which halts training when the framework's performance on the validation data plateaus to prevent overfitting; and (b) a learning rate scheduler that systematically reduces the learning rate by half if the validation loss stagnates for five consecutive epochs. Reducing the learning rate can help the framework converge to a better solution and avoid overfitting. In addition, we utilize the Adam optimization algorithm \cite{kingma2014adam} to update the trainable parameters of the framework. Our proposed framework enhances the accuracy of multi-class classification tasks by seamlessly integrating both large language models (LLMs) and small-scale language models (LMs). The framework fully leverages the capabilities of LLMs in generating technical descriptions of nanomaterials, an approach that can significantly exploit domain-specific linguistic insights critical for nanomaterial identification tasks. The framework interacts with off-the-shelf LLMs through a Language Model as a Service (LaMaaS) platform through the text-based API interactions. In this study, we utilized GPT-3.5-turbo and Google Bard as representative LLMs. The hyperparameters for our framework were not individually fine-tuned for each LLM. Instead, they were consistently applied across all LLMs. This method underscores our framework's generality, ease of use, and compatibility with existing off-the-shelf LLMs. For decoder-only LLMs, the maximum output token sequence length is 4096 for GPT-3.5-turbo and 4000 for Google Bard. To optimize computational resource use, the system is trained on eight V100 GPUs, each boasting 8 GB of GPU memory, utilizing the PyTorch framework. This configuration ensures the training process is completed within a reasonable timeframe. Given the potentially high computational cost of using prompting with LLMs, we conducted each experiment twice and reported the averaged results.

\vspace{-4mm}
\subsection{Ablation Study}
\begin{figure}[htbp]
\centering

\begin{tikzpicture}[auto, node distance=4cm,>=latex']
    \node [input, name=input] {};
    \node [sum, right of=input] (sum) {};
    \node [block,  below left of=sum, node distance = 1.0cm, xshift=-0.6cm, yshift=0cm] (controller) {$\textbf{HNF}$};
    \node [block, below right of=sum, node distance = 1.0cm, xshift=0.6cm, yshift=0cm] (system_) {$\textbf{LLM's}$};
    \node [block, below of=sum, node distance = 2.0cm] (system__) {$\textbf{Output Layer}$};

    \draw [->] (sum) -- node[label={[xshift=0.3cm, yshift=0.25cm]$\textbf{Electron Micrograph}$}] {} (controller);
    \draw [->] (sum) -- node[label={[xshift=-1.0cm, yshift=0.15cm]$ $}] {} (system_);

    \node [output, below right of=controller, xshift=-0.65cm, yshift=0.15cm, node distance = 1.62cm] (output) {};
    \draw [->] (controller) -- node[label={[xshift=-0.5cm, yshift=-0.5cm]}] {} (output); 

    \node [output, below left of=system_, xshift=0.65cm, yshift=0.15cm, node distance = 1.62cm] (output) {};
    \draw [->] (system_) -- node[label={[xshift=0.3cm, yshift=-0.35cm]}] {} (output); 
    
    \node [output, below of=system__, node distance = 0.8cm] (output) {};
    \draw [->] (system__) -- node[label={[xshift=0.6cm, yshift=-0.5cm]$\textbf{Label}$}] {} (output); 

\end{tikzpicture}
\vspace{-2mm}
\caption{Overall, the architecture of our framework involves using zero-shot CoT prompting with LLMs to generate technical textual descriptions and pre-train smaller language models (LMs) using masked language modeling (MLM). We then jointly optimize the smaller LM along with the HNF method in supervised learning tasks, aiming to minimize cross-entropy loss and improve multi-class classification accuracy.}
\label{fig:overall}
\vspace{-5mm}
\end{figure}
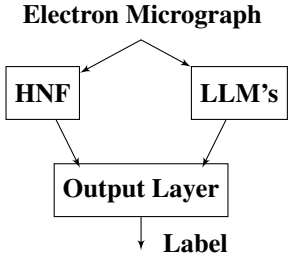

\begin{figure}[!ht]
\centering
\resizebox{0.625\linewidth}{!}{ 
\hspace*{0mm}\includegraphics[keepaspectratio,height=4.5cm,trim=0.0cm 0.0cm 0cm 0.0cm,clip]{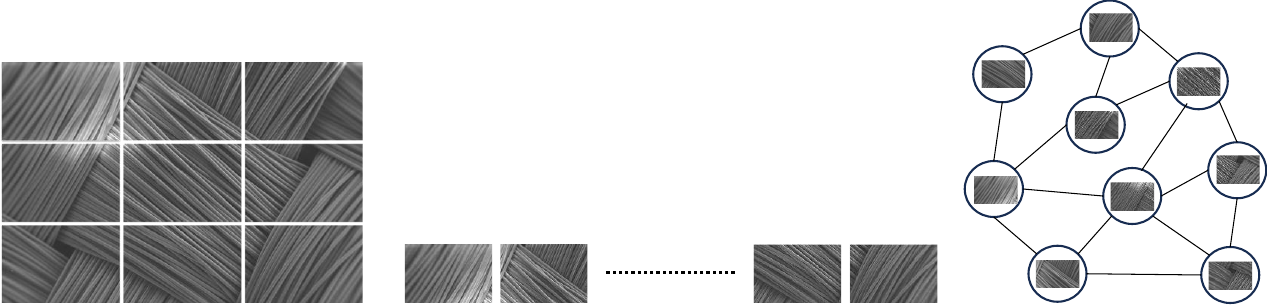} % left, bottom, right, top
}
\vspace{0mm}
\caption{In this illustrative example, we divided an electron micrograph (MEMS device, \cite{aversa2018first}) into a grid of $\text{3} \times \text{3}$ patches. The image presents various representations of the micrograph, including a regular grid, a sequence, and a graph representation from left to right, respectively. Different approaches for processing these representations include ConvNets that operate on pixel grids, ViTs that operate on patch sequences, and GNNs that operate on vision graphs. These graphs represent patches as nodes and are constructed using a nearest neighbor search algorithm, connecting patches based on visual similarity rather than spatial proximity. Each method offers a unique perspective for analyzing electron micrographs, providing distinct advantages and insights into patterns.} 
\label{fig:figure0}
\vspace{-3mm}
\end{figure}

\vspace{-3mm}
Figure \ref{fig:overall} illustrates the overview of the framework. Our proposed framework comprises three distinct methods: (a) The Hierarchical Network Fusion (HNF) is a multi-layered, cascading network architecture designed to enhance the classification accuracy of electron micrographs. It integrates two complementary representations at multiple layers: (a) patch sequences, which assist in capturing spatial dependencies among patches beyond pairwise dependencies, and (b) vision graphs, which capture the local pairwise patch relationships. These techniques provide a detailed multi-scale representation of the micrographs, encapsulating both fine-grained and coarse-grained details. HNF uses an inverted pyramid structure, incorporating increasing patch sizes at each layer, and utilizes bidirectional Neural ODEs and Graph chebyshev convolution(GCC) networks for iterative patch embeddings refinement and the computation of the optimal node-level embeddings, respectively. A mixture-of-experts technique further optimizes the integration of these cross-domain modalities, fostering efficient knowledge exchange and improving classification accuracy by effectively modeling structural, semantic, and causal information from both techniques. (b) Using Zero-shot CoT prompting with LLMs, we generate detailed technical descriptions of nanomaterials. We pre-train smaller LMs using masked language modeling (MLM) on these descriptions to facilitate domain-specific customization. These pre-trained LMs are then fine-tuned for task-specific adaptation to generate contextualized token embeddings. We apply a sum-pooling attention mechanism to obtain text-level embeddings from these token embeddings, thereby capturing the vast domain-specific knowledge embedded in the generated textual descriptions.  (c) We use the cross-modal multi-head attention mechanism to integrate and align information from different modalities --- specifically, from hierarchical network fusion (HNF) and language models --- into a coherent and unified representation that captures complex, hierarchical, and potentially cross-modal patterns, emphasizing relevant features to enhance the accuracy of the multi-class classification task.

\vspace{-4mm}

\vspace{-2mm}
\begin{table}[htbp]
\footnotesize
\centering
\setlength{\tabcolsep}{3.5pt}
\vspace{-2mm}
\resizebox{0.7\textwidth}{!}{%
\subfloat{%
\begin{tabular}{cc|c|c|c|c|c|c|cc}
\hline
\multicolumn{2}{c|}{\textbf{Algorithms}}                                     &  \textbf{Avg-Precision} & \textbf{Avg-Recall} & \textbf{Avg-F1 Score}  \\ \toprule
\multicolumn{1}{c}{\multirow{4}{*}{\rotatebox[origin=c]{90}{\textbf{}}}} & MultiFusion-LLM W-GPT4     & \textbf{0.941}	& \textbf{0.945}	& \textbf{0.939} \\ \midrule
\multicolumn{1}{c}{}                                          &  w/o HNF    &  0.776	& 0.753	& 0.745 \\
\multicolumn{1}{c}{}                                          &    w/o LLMs   & 0.714	& 0.726	& 0.721 \\
\multicolumn{1}{c}{}                                          &    w/o MHA    & 0.827 	& 0.831	& 0.823 \\ \hline
\end{tabular}}}
\vspace{1mm}
\caption{In the ablation study, we systematically disable individual methods to assess their respective contributions and importance. The goal of this study is to understand the impact or significance of specific methods on the overall performance of the framework. The experimental findings reveal the significance of the disabled methods, as indicated by the consistent decrease in performance metrics of the ablated variants compared to the baseline. These results substantiate our hypothesis regarding the joint optimization of HNF (see subsection \ref{HNF}) and LLMs (see subsection \ref{Languagemodels}) methods, demonstrating improved framework performance.}
\label{tab:table4}
\vspace{-4mm}
\end{table}

\vspace{-1mm}
To perform ablation studies, we systematically disabled certain methods to create various ablated variants, which were subsequently evaluated using the SEM dataset \cite{aversa2018first}, with our original framework serving as the baseline for comparison. This approach enables us to verify the effectiveness of our methods, substantiate their design decisions, and justify their inclusion in the framework.  A substantial decrease in performance of the ablated variants, compared to the baseline, underscores the significance of the omitted method. The ablated variants that exclude the hierarchical network fusion (HNF), large language models (LLMs), and the multi-head attention layer are denoted as proposed framework ``w/o HNF", ``w/o LLMs", and ``w/o MHA" respectively. The abbreviation "w/o" stands for "without". For the case of ``w/o MHA", we concatenate the cross-domain embeddings and transform them through a linear layer to predict the label. The findings from the ablation study are presented in Table \ref{tab:table4}. On the SEM dataset\cite{aversa2018first}, the ``w/o HNF" variant shows a substantial decline in performance relative to the baseline, evidenced by a significant drop of $17.53\%$ in \textbf{Avg-Precision}. Similarly, the ``w/o LLMs" variant performs much worse than the baseline, with a drop of $24.12\%$  in \textbf{Avg-Precision}. In addition, the ``w/o MHA" variant exhibited a notable deterioration in performance compared to the baseline, manifested by a substantial decrease of $11.9\%$  in \textbf{Avg-Precision}. This is attributed to the overly simplified linear operator in the output layer. The results of our ablation study clearly illustrate the crucial role of each omitted method, with the ablated variants demonstrating a consistent decline in performance metrics compared to the baseline.

\vspace{-4mm}
\subsection{An In-Depth Empirical Insights into Nanomaterial Classification}
\vspace{-3mm}
We have conducted additional experiments to gauge the efficacy of our framework, which sheds light on its ability to categorize electron micrographs across various nanomaterial categories. The experimental results, presented in Table \ref{tab:table5}, demonstrate that our proposed framework can generalize to a wide range of nanomaterials, including those with complex patterns. We evaluated the performance of our framework using the SEM dataset\cite{aversa2018first}, employing standard metrics such as precision (P in $\%$), recall (R in $\%$), and F1-score (F1 in $\%$). We adopt a multi-metric approach to ensure a fair and thorough comparison with baseline models. To facilitate this, we utilize a confusion matrix encompassing various metrics for multi-class classification. This confusion matrix aids in scrutinizing our framework's performance by offering insights into how it categorizes electron micrographs across different nanomaterial categories. The metrics included in the confusion matrix are as follows: True Positives (TP) represent micrographs that are correctly classified as belonging to a specific category. False Negatives (FN) represent micrographs that actually belong to a category but are incorrectly classified or missed.
True Negatives (TN) represent micrographs that are correctly identified as not belonging to a particular category. False Positives (FP) represent micrographs that are mistakenly classified as belonging to a category despite not actually belonging to that category. These metrics evaluate the accuracy and effectiveness of our framework in micrograph categorization. Precision (TP / (FP + TP)) measures the proportion of correctly classified micrographs for a specific category, while recall (TP / (FN + TP)) measures the proportion of all micrographs of a category that were accurately identified. The F1-score is computed as the balanced mean of precision and recall. It is important to note that the SEM dataset is highly class-imbalanced. Our framework demonstrates a relatively higher score in the classification of nanomaterial categories with a large number of labeled instances compared to those with fewer. This favorable performance of our proposed framework can be attributed to its reduced dependency on nanomaterial-specific relational inductive bias, setting it apart from traditional methods.

\vspace{-4mm}
\begin{table}[htbp]
\footnotesize
\centering
\resizebox{0.65\textwidth}{!}{%
\begin{tabular}{@{}c|ccc|c@{}}
\toprule
\multirow{2}{*}{\textbf{Category}}   & \multicolumn{3}{c|}{\textbf{Multi-class metrics}}                                    \\ \cmidrule(lr){2-4}
                            & \multicolumn{1}{c|}{\textbf{Precision}} & \multicolumn{1}{c|}{\textbf{Recall}} & \textbf{F1 Score} &                        \\ \midrule
Biological                  & \multicolumn{1}{c|}{0.931$\pm$0.009}     & \multicolumn{1}{c|}{0.943$\pm$0.007}      & 0.935$\pm$0.013 \\
Tips                        & \multicolumn{1}{c|}{0.909$\pm$0.005}     & \multicolumn{1}{c|}{0.919$\pm$0.008}      &  0.916$\pm$0.011                           \\
Fibres                      & \multicolumn{1}{c|}{0.979$\pm$0.007}     & \multicolumn{1}{c|}{0.965$\pm$0.012}      &  0.963$\pm$0.014                        \\
Porous Sponge               & \multicolumn{1}{c|}{0.929$\pm$0.014}     & \multicolumn{1}{c|}{0.941$\pm$0.013}      &  0.925$\pm$0.010                            \\
Films Coated Surface        & \multicolumn{1}{c|}{0.938$\pm$0.005}     & \multicolumn{1}{c|}{0.934$\pm$0.009}      &  0.941$\pm$0.008                  \\
Patterned surface           & \multicolumn{1}{c|}{0.946$\pm$0.016}     & \multicolumn{1}{c|}{0.942$\pm$0.006}      &  0.941$\pm$0.014                         \\
Nanowires                   & \multicolumn{1}{c|}{0.938$\pm$0.012}     & \multicolumn{1}{c|}{0.945$\pm$0.007}      &  0.948$\pm$0.011                        \\
Particles                   & \multicolumn{1}{c|}{0.935$\pm$0.006}     & \multicolumn{1}{c|}{0.937$\pm$0.011}      &  0.929$\pm$0.023                       \\
MEMS devices                & \multicolumn{1}{c|}{0.939$\pm$0.011}     & \multicolumn{1}{c|}{0.932$\pm$0.008}      &  0.923$\pm$0.009                        \\
Powder                      & \multicolumn{1}{c|}{0.941$\pm$0.014}     & \multicolumn{1}{c|}{0.928$\pm$0.009}      &  0.917$\pm$0.011
                          \\ \bottomrule
\end{tabular}}
\vspace{1mm}
\caption{The table illustrates the effectiveness of our proposed framework in identifying individual nanomaterial categories within the SEM dataset.}
\label{tab:table5}
\vspace{-4mm}
\end{table}

\vspace{-5mm}
\subsection{Baseline Algorithms}
\vspace{-3mm}
We have categorized our baseline methods into four distinct groups: Graph Neural Networks (GNNs) (\cite{rozemberczki2021pytorch, Fey/Lenssen/2019}), Graph Contrastive Learning (GCL) \cite{Zhu:2021tu}), Convolutional Neural Networks (ConvNets)\cite{philvformer, neelayvformer}, Vision Transformers (ViTs) (\cite{philvformer, neelayvformer}) and Vision Contrastive Learning (VCL) (\cite{susmelj2020lightly}) algorithms . We construct vision graphs to represent electron micrographs using the Top-K nearest neighbor search technique. In this representation, patches are treated as nodes, and pairwise associations between semantically similar nearest-neighbor nodes are represented as edges. For the baselines, we avoid constructing multi-scale vision graphs with increasing patch resolutions. Instead, we set the patch size to 32 pixels to reduce the complexity of the baseline models and set $K$ to 5 for finding the nearest neighbors. The baseline Graph Neural Networks (GNNs)\cite{rozemberczki2021pytorch, Fey/Lenssen/2019}) are used for the multi-class classification task on vision graphs through supervised learning. The graph contrastive learning (GCL) algorithms (\cite{Zhu:2021tu}) utilize several graph data augmentation strategies to create multiple correlated views of a vision graph. GCL aims to maximize the similarity between positively correlated views of a graph while minimizing dissimilarity with others, thereby learning invariant self-supervised node-level embeddings. The GCL algorithms employ the Graph Attention Network (GAT) (\cite{velivckovic2017graph}) as the node-level graph encoder. Graph-level embeddings are generated by performing sum-pooling on the node-level embeddings. During inference, the Random Forest (RF) algorithm utilizes these robust self-supervised graph-level embeddings to predict nanomaterial categories, having been trained using supervised learning. To evaluate the effectiveness of the unsupervised embeddings, we measure the classification accuracy of the RF model on the holdout data. In addition, we employ baseline ConvNets (\cite{philvformer, neelayvformer}) operating on the regular grid of pixels in electron micrographs for classification tasks using supervised learning. We also utilize baseline Vision Transformers (ViTs) (\cite{philvformer, neelayvformer}) trained through supervised learning to analyze patch sequences within each electron micrograph for classification tasks. Furthermore, we utilize visual-contrastive learning (VCL) techniques (\cite{susmelj2020lightly}), which are self-supervised algorithms designed for contrastive learning in computer vision tasks. We employ the ResNet backbone architecture for feature extraction.

\vspace{-3mm}

\vspace{-4mm}
\subsection{Hyperparameter Studies}
\vspace{-3mm}
We performed an in-depth hyperparameter tuning to determine the optimal hyperparameters for our framework. The hyperparameters of the algorithm are: (1) the dimensionality of the embedding ($d$), and (2) batch size ($b$). The hyperparameters were chosen from the following ranges: embedding dimension ($d$) $\in [32, 64, 128, 256]$ and batch size ($b$) $\in [32, 48, 64, 96]$. We conducted hyperparameter optimization using the random-search technique to achieve the optimal performance of our proposed framework on the validation dataset, measured in terms of Top-1 classification accuracy. For each experiment, we altered the hyperparameter under investigation to ascertain its impact on the framework's performance. The study determined that the optimal hyperparameters are $d = 64$ and $b = 48$.

\vspace{-4mm}
\begin{table}[htbp]
\centering
\setlength{\tabcolsep}{2pt}
\resizebox{0.90\textwidth}{!}{%
\subfloat{%
\begin{tabular}{@{}c|c|c|c|cc@{}}
\hline
$(d, b) $ & $(32, 48)$ & $(64, 48)$  & $(128, 48)$ & $(256, 48)$ \\
\hline
 &  0.941  & 0.947  &  0.935 &   0.927 \\
\hline
\end{tabular}}
\qquad% --- set horizontal distance between tables here
\subfloat{%
\begin{tabular}{@{}c|c|c|c|cc@{}}
\hline
$(d, b)$ &  $(64,32)$ & $(64, 48)$  & $(64, 64)$ & $(64,96)$ \\
\hline
  &  0.943 & 0.947  &  0.939 & 0.936  \\
\hline
\end{tabular}}
\qquad% --- set horizontal distance between tables here
}
 \vspace{1mm}
\caption{The table reports the experimental findings of the hyperparameter study.}
\label{table:Hs}
\end{table}

\vspace{-8mm}
\subsection{Benchmarking with open-source material datasets}
\begin{figure}[htbp]
    \centering
    \includegraphics[scale=.4]{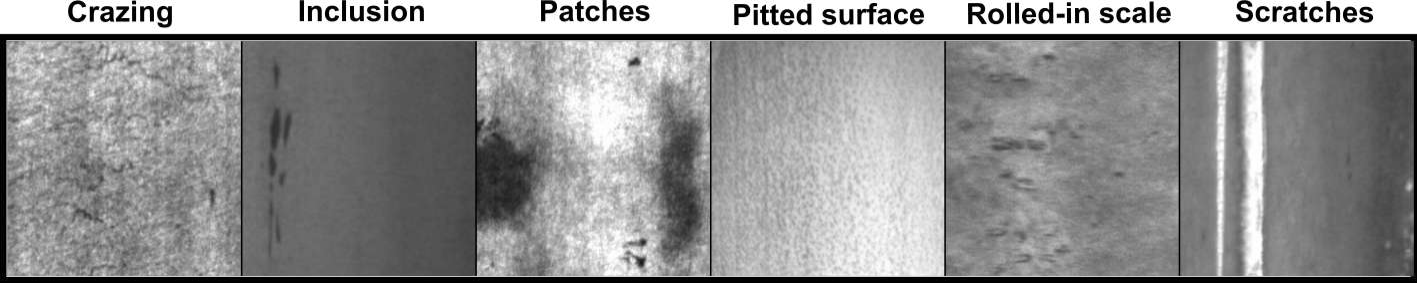}
    \vspace{-1mm}
    \caption{The NEU-SDD dataset contains six distinct defect categories found in hot-rolled steel strips, which are described in reference \ref{note1}(\cite{deshpande2020one}).}
    \label{fig:NUE}
     \vspace{-1mm}
\end{figure}

\begin{figure}[htbp]
    \centering
    \includegraphics[scale=0.45]{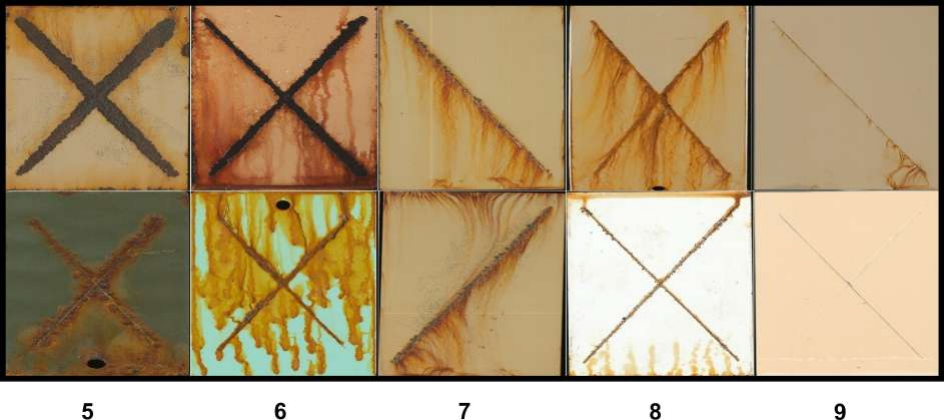}
    \vspace{-2mm}
    \caption{The CMI dataset is a collection of electron micrographs that represent five corrosion rating categories. These categories are described in reference \ref{note2}.}
    \label{fig:corrosion}
    \vspace{-4mm}
\end{figure}

\begin{figure}[htbp]
    \centering
    \includegraphics[scale=0.475]{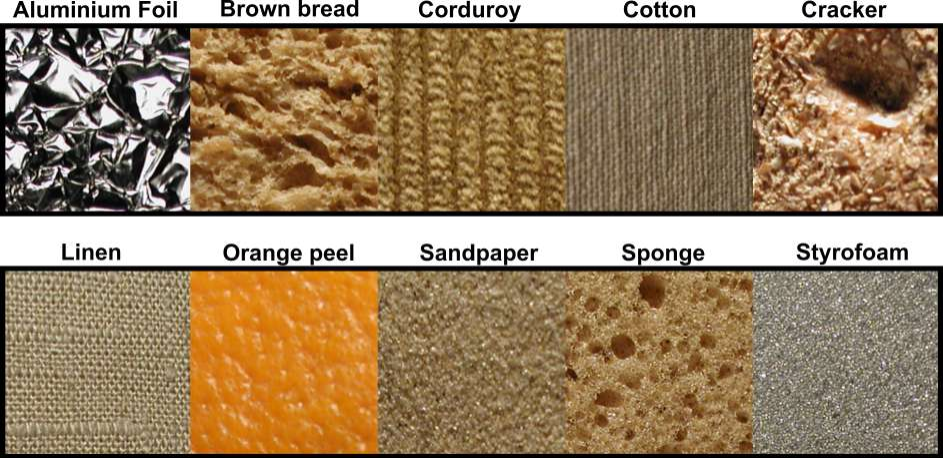}
    \vspace{-1mm}
    \caption{The KTH-TIPS dataset contains samples of electron micrographs of ten distinct materials. These materials are described in reference \ref{note3}.}
    \label{fig:KTH}
     \vspace{-2mm}
\end{figure}

\vspace{-2mm}
\begin{itemize}
    \item \textbf{NEU-SDD\footnote{Datasource: \url{http://faculty.neu.edu.cn/yunhyan/NEU_surface_defect_database.html}\label{note1}}
    }(\cite{deshpande2020one}) is a comprehensive database comprising 1800 grayscale electron micrographs of surface defects on hot-rolled steel strips. The dataset is divided into six distinct defect classes, each containing 300 micrographs with a resolution of 200$\times$200 pixels. The defect categories include \textit{pitted surfaces, scratches, rolled-in scale, crazing, patches, and inclusion defects}. Figure \ref{fig:NUE} displays representative images from each category. We conducted a comparative analysis using various standard algorithms to evaluate the effectiveness of our proposed approach, specifically in the domain of multi-class classification tasks for surface defect identification.
    \vspace{-1mm}      
    \item \textbf{CMI\footnote{\url{https://arl.wpi.edu/corrosion_dataset}\label{note2}}
    } consists of 600 high-resolution electron micrographs depicting corroding panels. Each micrograph has been annotated by corrosion experts following ASTM-D1654 standards, assigning discrete ratings ranging from 5 to 9. The dataset includes 120 distinct micrographs for each corrosion rating with a spatial resolution of 512×512 pixels. Figure \ref{fig:corrosion} illustrates a selection of representative images for each rating. Our proposed method for multiclass classification task is evaluated by comparing its performance against several standard algorithms. 
     \vspace{-1mm}          
    \item \textbf{KTH-TIPS\footnote{\url{https://www.csc.kth.se/cvap/databases/kth-tips/index.html}\label{note3}}
    }  represents a comprehensive texture dataset, comprising 810 electron micrographs, each depicting one of ten distinct material types. These micrographs, with a resolution of 200 x 200 pixels, encompass a wide range of materials captured under different illuminations, poses, and scales. The diverse material categories encompass textures such as \textit{sponge, orange peel, styrofoam, cotton, cracker, linen, brown bread, sandpaper, crumpled aluminum foil, and corduroy}. Figure \ref{fig:KTH} showcases a selection of sample images from each category. In order to assess and demonstrate the efficacy of our proposed method, we conduct a comparative analysis of its performance against various standard algorithms within the domain of multi-class identification tasks.   
\end{itemize}

\vspace{-3mm}

\vspace{-3mm}

\vspace{-1mm}

\vspace{-2mm}
Table \ref{table:table6} presents a comprehensive comparison of the performance achieved by our proposed approach in contrast to various baseline methods, evaluated across all datasets. The experimental results demonstrate that our method achieves state-of-the-art performance on all datasets, underscoring the efficacy and robustness of our framework.

\vspace{-4mm}
\begin{table}[htbp]
\footnotesize
\centering
\resizebox{0.55\textwidth}{!}{%
\subfloat{%
\setlength{\tabcolsep}{3pt}
\begin{tabular}{cc|cccc}
\hline
\multicolumn{2}{c|}{\textbf{Algorithms}}                                       & \textbf{NEU-SDD} & \textbf{CMI} & \textbf{KTH-TIPS}  \\ \hline
\multicolumn{1}{c|}{\multirow{4}{*}{\rotatebox[origin=c]{90}{\textbf{Baselines}}}} & ResNet                   & 0.906	& 0.928	& 0.941 &             \\
\multicolumn{1}{c|}{}                                          & GoogleNet                & 0.936	& 0.928	& 0.929
              \\
\multicolumn{1}{c|}{}                                          & SqueezeNet                & 0.955	& 0.943	& 0.963
              \\ 
\multicolumn{1}{c|}{}                                          & VanillaViT               & 0.962	& 0.968	& 0.972
 \\ 
\hline
\multicolumn{1}{c|}{}                                          & \textbf{MultiFusion-LLM}                  &     \textbf{0.993}              &      \textbf{0.989}             &    \textbf{0.991}               &                     \\ \hline
\end{tabular}}}
\vspace{2mm}
\caption{The table presents the comparative evaluation of our proposed framework's performance against several benchmark algorithms on a variety of datasets.}
\label{table:table6}
\vspace{-9mm}
\end{table}

\vspace{0mm}
\subsection{Graph Chebyshev convolution}
\vspace{-2mm}
The graph convolution is a powerful tool in the realm of learning from graph-structured data. The spectral graph convolution\cite{tanaka2021graph} is a popular approach, but it can be computationally expensive for large graphs. To tackle this issue, Chebyshev graph convolution\cite{defferrard2016convolutional} offers a more scalable approach that can be used to achieve similar performance in capturing the local connectivity and spectral properties of the graph. More precisely, Graph Chebyshev Convolution is a method that approximates the spectral graph convolution by using Chebyshev polynomials. Graph Chebyshev Convolution allows us to apply convolutional filters on graph-structured data based on the Chebyshev polynomial approximation of the graph Laplacian. The Chebyshev polynomials are calculated based on the normalized Laplacian matrix of the graph. The normalized Laplacian matrix, denoted as $\hat{L}$, is defined as:

\vspace{-4mm}
\resizebox{0.95\linewidth}{!}{
\begin{minipage}{\linewidth}
\begin{equation}
    \hat{L} = \hat{D}^{-1/2} \hat{A} \hat{D}^{-1/2}
\end{equation}
\end{minipage}
}

\vspace{-1mm}
where $\hat{A}$ is the normalized adjacency matrix and $\hat{D}$ is the diagonal degree matrix of the graph. The Chebyshev approximation of the graph Laplacian up to any desired degree is obtained by using a truncated expansion of Chebyshev polynomials, denoted as $T_k(\hat{L})$, where k represents the degree of the polynomial. These polynomials are computed recursively using the following recurrence relation as follows:

\vspace{-3mm}
\resizebox{0.935\linewidth}{!}{
\begin{minipage}{\linewidth}
\[
T_k(\hat{L}) = 
\begin{cases}
I, & \text{if } k = 0 \\
\hat{L}, & \text{if } k = 1 \\
2\hat{L}T_{k-1}(\hat{L}) - T_{k-2}(\hat{L}), & \text{otherwise}
\end{cases}
\]
\end{minipage}
}

\vspace{-1mm}
where $I$ is the identity matrix. Given an input graph feature matrix $I \in \mathbb{R}^{\hspace{0.25mm}n \times d}$, where $n$ denotes the number of patches and $d$ is the patch embedding dimension, and the Chebyshev polynomials denoted by $T_k(\hat{L})$. The Chebyshev graph convolution operation can be defined as follows:

\vspace{-1mm}
\resizebox{0.925\linewidth}{!}{
\begin{minipage}{\linewidth}
\begin{equation}
E = \sigma\left(\sum_{k=0}^{K-1} T_k(\hat{L}) I \Theta_k\right)
\end{equation}
\end{minipage}
}

\vspace{-1mm}
where $\sigma(\cdot)$ is a non-linear ReLU activation function applied element-wise and $\Theta_k \in \mathbb{R}^{d \times d}$ is the parameter matrix (weights) for the $k$-th order Chebyshev polynomial. It is important to note that the parameter matrices $\Theta_k$ are typically learnable and optimized during the training process to adaptively capture the global graph characteristics. $K$ denotes the maximum order of the Chebyshev polynomials and influences the expressive power of the approximation. $E \in \mathbb{R}^{\hspace{0.25mm}n \times d}$ is the transformed node feature matrix, which captures the local structure and relationships within the graph, where $e_{i} \in \mathbb{R}^{\hspace{0.5mm}d}$ denotes the node embedding.

\vspace{-4mm}
\subsection{Neural Ordinary Differential Equations (NODE)}
\vspace{-2mm}
Neural Ordinary Differential Equations (Neural ODE) \cite{chen2018neural} represent a deep neural network model designed for continuous-time systems, in contrast to traditional discrete-time neural networks. In the Neural ODE framework, we denote the hidden state of a dynamic system at a given time $t$ as $\textbf{z}(t)$. The objective is to determine the evolution of $\textbf{z}(t)$ by calculating its derivative with respect to time to capture the temporal dynamics of the system. This derivative is represented by a parameterized neural network function, denoted as $f(\textbf{z}(t), t, \theta)$, as follows:

\vspace{-1mm}
\resizebox{0.95\linewidth}{!}{
\begin{minipage}{\linewidth}
\begin{equation}
    \frac{d\textbf{z}(t)}{dt} = f(\textbf{z}(t), t, \theta) 
\end{equation}
\end{minipage}
}

\vspace{-1mm}
Here, $\theta$ represents the parameters of the neural network $f(\cdot)$. To compute the output of the Neural ODE framework, an ODE solver takes the initial hidden state $\textbf{z}({t_{0}})$ at the starting time point $t_{0}$ and integrates the hidden state derivative over time to produce the hidden state $\textbf{z}({t_{1}})$ at the specified output time point $t_{1}$, as described below:

\vspace{-2mm}
\resizebox{0.95\linewidth}{!}{
\begin{minipage}{\linewidth}
\begin{equation}
\textbf{z}(t_1) = \textbf{z}(t_0) + \int_{t_0}^{t_1}f(\textbf{z}(t), t, \theta) dt  
\label{eqn:equation1}
\end{equation}
\end{minipage}
}

In summary, by formulating neural networks as continuous-depth models through Neural ODEs, this framework can generate the hidden state of a dynamic system at any given time point and effectively handle continuous-time data. This characteristic makes it particularly useful for modeling continuous-time dynamic systems. Furthermore, to reduce memory requirements during backpropagation, Chen et al. \cite{chen2018neural} introduced the adjoint sensitivity method for Neural ODEs. An adjoint, denoted as $\textbf{a}(t) = \frac{\partial\mathcal{L}}{\partial \textbf{z}(t)}$, is defined, where $\mathcal{L}$ represents the loss function. The gradient of $\mathcal{L}$ with respect to the network parameters $\theta$ can be directly computed using the adjoint and an ODE solver:

\vspace{-2mm}
\resizebox{0.95\linewidth}{!}{
\begin{minipage}{\linewidth}
\begin{equation}
\frac{d\mathcal{L}}{d\theta} = - \int_{t_1}^{t_0} \textbf{a}(t)^T \frac{\partial f(\textbf{z}(t), t, \theta)}{\partial \theta} dt 
\end{equation}
\end{minipage}
}

In essence, the adjoint sensitivity method solves an augmented ODE backward in time, enabling the computation of gradients without the need for backpropagation through the ODE solver operations. This means that the model doesn't have to store intermediate results (partial derivatives) from forward propagation, resulting in a constant memory cost as a function of the depth. In this work, we incorporate Neural ODEs into computer vision tasks for electron micrograph classification by segmenting an electron micrograph into a sequence of patches. The sequence length is determined by the total number of patches generated through the tokenization of the electron micrograph, with each patch serving as an individual token in the sequence. Treating the input sequence of patches as a continuous-time system enables Neural ODEs to capture the evolution of the patch embeddings smoothly and continuously. Moreover, this approach facilitates the causal modeling of spatial relationships and transformations between consecutive patches by encoding them into patch embeddings.
In this work, we model the neural network $f(\cdot)$ using a transformer encoder\cite{vaswani2017attention}. It consists of a stack of encoder layers, each containing self-attention mechanisms and feed-forward neural networks. The encoder layers capture the relationships and dependencies between the patches in the image. We learn the bidirectional representation of sequences to capture information from both the past and future context of a given patch in a sequence. Our bidirectional representation learning approach incorporates two separate Neural ODEs: one that processes the sequence from left to right (forward pass) and another that processes the sequence from right to left (backward pass). Each pass maintains its own hidden state, and the outputs of both passes are combined through a gating mechanism. Let's denote the forward Neural ODE estimate of the patch embedding at time point $t_1$ as $\textbf{z}_{f}(t_1)$ and the backward Neural ODE as $\textbf{z}_{b}(t_1)$ using Equation \ref{eqn:equation1}. A gating mechanism is implemented to regulate the information flow from $\textbf{z}_{f}(t_1)$ and $\textbf{z}_{b}(t_1)$, which produces a weighted combination of representations $h_{(t_1)}$. The gating mechanism is described as follows:

\vspace{-4mm}
\resizebox{0.975\linewidth}{!}{
\begin{minipage}{\linewidth}
\begin{align}
 g &= \sigma \big( f^{\prime}(\textbf{z}_{f}(t_1)) + f^{\prime\prime}(\textbf{z}_{b}(t_1)) \big)  \\
h_{(t_1)}  &= \sigma \big( g(\textbf{z}_{f}(t_1)) + (1-g)(\textbf{z}_{b}(t_1)) \big) 
\end{align} 
\end{minipage}
} 

\vspace{-1mm}
where $f^{\prime}$ and $f^{\prime\prime}$ are linear projections. In our work, the use of adaptive ODE solvers can lead to significant time consumption. To ensure manageable training time, we use fixed-grid ODE solvers in combination with the Interpolated Reverse Dynamic Method (IRDM) proposed by Daulbaev et al.\cite{daulbaev2020interpolated}. The IRDM employs Barycentric Lagrange interpolation\cite{berrut2004barycentric} on a Chebyshev grid\cite{tyrtyshnikov1997brief} to approximate the solution of patch embeddings during the reverse-mode differentiation (referred to as backpropagation) through the ODE solver. By incorporating IRDM, we can reduce the computational time during backpropagation while maintaining satisfactory learning accuracy. Specifically, we adopt a fixed-grid ODE solver, namely the fourth-order Runge-Kutta method\cite{butcher2007runge}, and implement the interpolated reverse dynamic method with 3 Chebyshev nodes. This approach enables us to ensure tractable training time without compromising precision.

\vspace{-4mm}
\subsection{Related Work}
\label{rw}
\vspace{-3mm}
In this section, we will first review the backbone architectures used in computer vision. Next, we will survey the evolution of graph neural networks, with a particular focus on Graph Convolutional Networks (GCN)\cite{kipf2016semi} and their utilization in vision tasks. The landscape of computer vision has been significantly shaped by convolutional networks(i.e., ConvNets or CNNs), which have brought about a seismic shift in the field and have established themselves as the predominant architecture (LeCun et al.\cite{lecun1998gradient}, Krizhevsky et al.\cite{krizhevsky2017imagenet}, He et al.\cite{he2016deep}). LeNet\cite{lecun1998gradient} significantly influenced the development and popularity of ConvNets, paving the way for more advanced and deeper networks in subsequent years across a broad spectrum of vision tasks, including image classification\cite{krizhevsky2017imagenet}, object detection\cite{faster2015towards}, and semantic segmentation\cite{long2015fully}. Over the past decade, groundbreaking advancements such as ResNet\cite{he2016deep}, MobileNet\cite{howard2017mobilenets}, and NAS\cite{zoph2016neural, yang2020cars} have further shaped the landscape of CNN architectures. The advent of the vision transformer(ViT)\cite{dosovitskiy2020image, han2022survey, carion2020end, chen2021pre} has been a trailblazer, leading to the development of a myriad of improved ViT variants\cite{dosovitskiy2020image}. These improvements encompass pyramid architectures\cite{SwinT, wang2021pyramid}, local attention mechanisms\cite{han2021transformer, SwinT}, and position encoding techniques\cite{wu2021rethinking}. Inspired by the vision transformer, researchers have also explored the potential of Multilayer Perceptrons (MLP) in computer vision tasks\cite{touvron2022resmlp, tolstikhin2021mlp}. By incorporating tailored modules\cite{chen2021cyclemlp, lian2021mlp, guo2022hire, tang2022image}, MLP-based techniques have demonstrated exceptional performance in general vision tasks, including object detection and segmentation. Graph Neural Networks (GNNs) originated from the early work of Scarselli et al.\cite{scarselli2008graph} and Gori et al.\cite{gori2005new}, introducing the concept of spatial graph convolutional networks with non-recursive layers\cite{micheli2009neural} to learn from graph-structured data. Since then, spatial GCNs have seen numerous adaptations and improvements, as proposed in previous works such as Niepert et al.\cite{niepert2016learning}, Atwood et al.\cite{atwood2016diffusion}, and Gilmer et al.\cite{gilmer2017neural}. Spectral GCNs, grounded in spectral graph theory, were first introduced in a study by Bruna et al.\cite{bruna2013spectral}. Subsequent methods to enhance these networks have been proposed in studies by Kipf et al.\cite{kipf2016semi}, Henaff et al.\cite{henaff2015deep}, and Defferrard et al.\cite{defferrard2016convolutional}. In the realm of computer vision\cite{xu2017scene, landrieu2018large, jing2022learning, wang2019learning}, Graph Convolutional Networks (GCNs) have been applied to diverse tasks including point cloud classification, scene graph generation, and action recognition. Point clouds refer to sets of 3D points derived from LiDAR scans, where GCNs have been leveraged for classification and segmentation\cite{landrieu2018large, wang2019dynamic, yang2020distilling}. The process of scene graph generation involves parsing an input image into a graph representation that delineates objects and their interrelationships, often integrating object detection with GCN techniques\cite{yang2018graph}. Furthermore, GCNs have been instrumental in facilitating human action recognition tasks by analyzing graphs representing linked human joints\cite{yan2018spatial, jain2016structural}. Current frameworks in the semiconductor manufacturing sector fall short in various aspects, especially when compared to recently proposed advancements. Many existing solutions fail to capitalize on the detailed analysis achievable through the synergy of patch sequences and vision graphs at different scales in electron micrographs. Moreover, these frameworks typically analyze data at a singular scale, missing the opportunities that a multi-scale approach could offer in enhancing classification accuracy. Furthermore, the industry has yet to fully embrace the utilization of large language models (LLMs) in generating technical descriptions of nanomaterials, a strategy that can significantly deepen domain-specific insights critical for nanomaterial identification tasks. This glaring gap in the integration of image-based and linguistic insights renders current architectures less comprehensive and nuanced, potentially impeding breakthroughs in the semiconductor industry. The proposed ``MultiFusion-LLM" framework represents a pioneering approach to automatic nanomaterial identification, addressing the prevailing challenges in semiconductor manufacturing processes. At its core, it employs a hierarchical network fusion (HNF) architecture that amalgamates two diverse representations of electron micrographs: patch sequences and vision graphs. These representations undergo iterative refinement through a layered, inverted pyramid structure, integrating multi-scale information to enhance classification accuracy. Additionally, the framework introduces a zero-shot chain-of-thought (Zero-Shot CoT) prompting technique that utilizes large language models (LLMs) to generate detailed technical descriptions of nanomaterials. These descriptions, rich in domain-specific knowledge, serve as auxiliary information that bolsters identification tasks. By encapsulating both image-based and linguistic insights, the framework aims to facilitate a more comprehensive and nuanced analysis of electron micrographs, thereby holding significant promise for advancements in the semiconductor industry.

\clearpage

\begin{tabularx}{1.25\textwidth}{bss}
\caption{LLM-retrieved text of GPT-3.5 for natural language queries about MEMS devices.}\label{tab:lmprompts}\\
\hline
\toprule
\textbf{Query prompt}  & \textbf{LLM-retrieved text} \\
\midrule
\endhead
\textcolor{black}{Introduction: Provide an overview of the MEMS and its significance in various fields?} & Introduction: Micro-Electro-Mechanical Systems, commonly known as MEMS, is a cutting-edge technology that integrates miniaturized mechanical and electromechanical elements with electronics on a single silicon chip or substrate. These systems typically consist of components with dimensions ranging from micrometers to millimeters. MEMS devices have revolutionized various industries and fields due to their unique capabilities, compact size, low power consumption, and cost-effectiveness. They enable the creation of tiny, sophisticated systems with enhanced functionalities, making them indispensable in numerous applications.

\vspace{1mm}
Significance in Various Fields:

\vspace{1mm}
Consumer Electronics: MEMS devices have played a crucial role in the advancement of consumer electronics. Accelerometers, gyroscopes, and magnetometers in smartphones enable motion sensing, orientation detection, and navigation. MEMS microphones are widely used for voice and audio input in smartphones, laptops, and other gadgets.

\vspace{1mm}
Automotive: MEMS sensors are fundamental to the functioning of modern vehicles. They are used in airbag deployment systems, tire pressure monitoring systems, electronic stability control, and more. MEMS accelerometers and gyroscopes provide data for advanced driver assistance systems (ADAS) and vehicle navigation.

\vspace{1mm}
Aerospace and Defense: MEMS technology has revolutionized aerospace and defense applications. Inertial sensors, such as accelerometers and gyroscope arrays, are vital for aircraft and spacecraft stabilization, guidance, and control. MEMS-based pressure sensors are used in altimeters and barometers, and microbolometers for infrared imaging.

\vspace{1mm}
Industrial Applications: MEMS devices find extensive use in industrial automation and control systems. They are used for precise measurement, monitoring, and feedback control in manufacturing processes, structural health monitoring, and predictive maintenance.

\vspace{1mm}
Robotics: MEMS sensors are essential components in robotics and autonomous systems. They provide feedback for motion control, balance, and obstacle detection, enabling robots to navigate and interact with their environment safely.

\vspace{1mm}
Telecommunications: MEMS technology has been instrumental in the miniaturization of RF switches, tunable capacitors, and filters, leading to improved performance and efficiency in wireless communication devices.

\vspace{1mm}
Energy Harvesting: MEMS energy harvesters are used to capture and convert ambient energy sources such as vibrations, solar, and thermal energy into usable electrical power for low-power electronic devices.

\vspace{1mm}
The significance of MEMS in these and many other fields stems from their ability to offer scalable, high-performance, and cost-effective solutions for various applications. The continuous advancement in MEMS fabrication techniques and materials has opened up new possibilities for even more innovative and diverse applications in the future. As research and development in this field continue, we can expect MEMS to remain at the forefront of technological progress across numerous industries. \\

\midrule

\textcolor{black}{Definition and Structure: Define the MEMS and describe its typical structure at the nanoscale.?} & Definition of MEMS: Micro-Electro-Mechanical Systems (MEMS) refers to a class of miniaturized devices and systems that integrate micro-scale mechanical components, electrical circuits, and sometimes optical elements on a single substrate. These components are typically fabricated using semiconductor manufacturing techniques, similar to those used for integrated circuits (ICs). MEMS devices are characterized by their small size, with components ranging from micrometers to millimeters, and they often operate in the micro/nano scale. 

\vspace{1mm}
Typical Structure of MEMS at the Nanoscale: MEMS devices exhibit a wide variety of structures and designs, each tailored to their specific application. At the nanoscale, MEMS structures often involve mechanical components with dimensions in the micrometer range and electrical connections at the nanoscale level. The following are the key components commonly found in MEMS devices at the nanoscale:

\vspace{1mm}
Substrate: The foundation of a MEMS device is typically a silicon substrate. Silicon wafers are commonly used due to their excellent mechanical properties, compatibility with semiconductor processes, and availability in various sizes.

\vspace{1mm}
Mechanical Components: MEMS devices consist of micro or nano-scale mechanical elements that can move or respond to external stimuli. These components may include beams, cantilevers, diaphragms, membranes, gears, and more. The design and material of these structures depend on the specific application.

\vspace{1mm}
Actuators: Actuators are responsible for generating mechanical motion in MEMS devices. They can be electrostatic, piezoelectric, thermal, magnetic, or other types, depending on the required movement and sensitivity.

\vspace{1mm}
Sensors: MEMS sensors are vital for detecting and measuring physical quantities such as acceleration, pressure, temperature, humidity, etc. Common MEMS sensors include accelerometers, gyroscopes, pressure sensors, and temperature sensors.

\vspace{1mm}
Transducers: Transducers convert one form of energy into another. In MEMS, they are often used to transform mechanical motion or physical parameters into electrical signals or vice versa.

\vspace{1mm}
Interconnects: Nanoscale electrical interconnects provide the necessary connections between the MEMS components and the external circuits or systems. These connections are usually made using metal traces or vias.

\vspace{1mm}
Control Electronics: For some MEMS devices, integrated control electronics are present on the same chip to process, amplify, and transmit signals to and from the mechanical components and sensors.

\vspace{1mm}
The fabrication of these nano/micro-scale components involves various advanced processes, including photolithography, etching, deposition, and bonding. The choice of materials, such as silicon, polymers, metals, and composites, is critical to achieving the desired mechanical and electrical properties. The combination of these components and their precise integration on a single substrate allows MEMS devices to perform a wide range of functions across various industries, making them a fundamental technology in the world of miniaturized electronics and sensors. \\

\midrule

\textcolor{black}{Synthesis Methods:
Explore different methods used to synthesize or fabricate MEMS devices in this category. Discuss their advantages and limitations.?} & Various methods are used to synthesize or fabricate MEMS devices, each with its own advantages and limitations. Below are some commonly used synthesis methods:

\vspace{1mm}
Bulk Micromachining: Advantages: (a) Well-established and widely used method, (b) Suitable for fabricating simple, 2D MEMS structures and (c) Relatively low cost and straightforward process. Limitations: (a) Limited design flexibility for complex 3D structures, (b) Fabrication of delicate or high-aspect-ratio structures can be challenging, and (c) Not suitable for integrating multiple layers of different materials.

\vspace{1mm}
Surface Micromachining: Advantages: (a) Enables the fabrication of complex 3D structures, (b) Suitable for high-volume production due to batch processing capabilities, and (c) Can integrate multiple layers of different materials. Limitations: (a) Typically limited to planar structures, (b) Requires sacrificial layer techniques, which can be more complicated and time-consuming than bulk micromachining, and (c) Material choices are often limited to specific combinations suitable for sacrificial release.

\vspace{1mm}
LIGA (Lithography, Electroplating, and Molding): Advantages: (a) Capable of fabricating high-aspect-ratio structures with excellent precision, (b) Suitable for mass production of microstructures, and (c) Compatible with various materials. Limitations: (a) Complex and costly process due to multiple steps involved, (b) Limited to metallic materials for electroplating, and (c) Difficulties in creating free-standing structures.

\vspace{1mm}
Soft Lithography: Advantages: (a) Simple and cost-effective technique, (b) Suitable for rapid prototyping and small-scale production, and (c) Enables the use of a wide range of materials, including polymers and biomaterials. Limitations: (a) Limited to simple structures and low aspect ratios, (b) Resolution and precision may not be as high as other techniques, and (c) Not suitable for high-temperature or harsh environments.

\vspace{1mm}
Silicon-on-Insulator (SOI) Technology: Advantages: (a) Provides high mechanical stability and isolation, (b) Suitable for integrating MEMS and CMOS components on the same chip, and (c) Excellent electrical isolation between components. Limitations: (a) Higher cost compared to standard silicon wafers, (b) Limited availability of thick SOI wafers, restricting the aspect ratio of structures, and (c) Challenges in wafer bonding and handling.

\vspace{1mm}
Nanolithography and Nanofabrication: Advantages: (a) Enables fabrication of nanoscale MEMS devices with high precision, (b) Allows for the development of novel and highly sensitive sensors and actuators, and (c) Can be used to create complex nanostructures and patterns. Limitations: (a) Expensive and requires specialized equipment, (b) Limited to research and development due to high costs and complexity, and (c) Mass production of nanoscale MEMS devices remains challenging.

\vspace{1mm}
3D Printing/Additive Manufacturing: Advantages: (a) Enables rapid prototyping and customization of MEMS devices, (b) Supports the creation of complex 3D geometries and intricate structures, and (c) Reduces the need for complex lithography and etching steps. Limitations: (a) Limited resolution compared to traditional lithography-based methods, (b) Material choices may be restricted for some 3D printing techniques, and (c) Post-processing steps might be required for improved surface finishes and accuracy.

\vspace{1mm}
Overall, the choice of synthesis method depends on the requirement of material properties, volume of production, and budget constraints.  \\

\midrule

\textcolor{black}{Properties: Highlight the unique physical, chemical, and electronic properties exhibited by MEMS in this nanomaerial category. Discuss how these properties differ from their bulk counterparts.?} & MEMS (Micro-Electro-Mechanical Systems) exhibit several unique physical, chemical, and electronic properties at the nanoscale, which differentiate them from their bulk counterparts. Below are some of the key properties:

\vspace{1mm}
Size and Scale: The most obvious property of MEMS is their miniature size, typically ranging from micrometers to millimeters. This reduced scale allows for compact, lightweight, and portable devices, which is not achievable in their bulk counterparts.

\vspace{1mm}
Surface-to-Volume Ratio: At the nanoscale, MEMS devices have a higher surface-to-volume ratio compared to bulk devices. This increased surface area enhances interactions with the surrounding environment, leading to improved sensitivity in sensors and more efficient reactions in chemical and biological applications.

\vspace{1mm}
Mechanical Properties: MEMS structures often exhibit different mechanical properties compared to their bulk counterparts. For example, nanoscale mechanical components can have enhanced flexibility, higher resilience, and reduced inertia due to their reduced mass and size. These properties are advantageous in applications such as accelerometers and gyroscopes.

\vspace{1mm}
Thermal Properties: The reduced dimensions of MEMS components result in lower thermal mass and faster response to changes in temperature. This property is vital for accurate temperature sensing and thermal management in compact devices.

\vspace{1mm}
Electrical Properties: MEMS devices at the nanoscale can exhibit unique electrical properties, such as high sensitivity to electrical signals. Piezoelectric and electrostatic properties are commonly utilized for sensing and actuation in MEMS devices.

\vspace{1mm}
Optical Properties: Some MEMS devices, especially those incorporating optical elements, can exhibit novel optical properties due to the interaction of light with nanoscale structures. This enables the development of miniaturized optical sensors, displays, and communication systems.

\vspace{1mm}
Material Properties: At the nanoscale, materials can exhibit different properties than at the bulk level. For example, nanomaterials may have altered mechanical, electrical, and optical characteristics due to quantum effects and surface interactions. MEMS devices can leverage these unique material properties for specific functionalities.

\vspace{1mm}
Sensitivity: MEMS sensors at the nanoscale are highly sensitive to various stimuli, such as mechanical, electrical, thermal, and chemical changes. This sensitivity enables precise measurements and detections, making MEMS devices suitable for a wide range of applications, including healthcare, environmental monitoring, and industrial automation.

\vspace{1mm}
Energy Efficiency: The reduced size and lower power requirements of MEMS devices at the nanoscale make them more energy-efficient compared to bulk systems. This energy efficiency is crucial for portable and battery-operated devices.

\vspace{1mm}
Integration: MEMS devices can integrate multiple functionalities, such as sensing, actuation, and signal processing, on a single chip. This integration capability allows for the development of complex, multifunctional systems in a compact form factor.

\vspace{1mm}
In summary, the unique physical, chemical, and electronic properties of MEMS at the nanoscale make them ideal for a wide range of applications, including sensors, actuators, and biomedical devices. \\

\midrule

\textcolor{black}{Applications: Explore the wide range of applications where MEMs are utilized. Discuss their potential impact in fields such as electronics, energy, medicine, environmental remediation, etc.} & MEMS (Micro-Electro-Mechanical Systems) have found extensive applications across various industries due to their unique properties and miniaturized form factor. Here are some of the key applications of MEMS in different fields:

\vspace{1mm}
Electronics and Consumer Devices: (a) MEMS accelerometers and gyroscopes in smartphones and wearable devices enable motion sensing and orientation detection for gaming, navigation, and fitness tracking, (b) MEMS microphones are used in smartphones, tablets, and other consumer electronics for voice recording and noise cancellation, and (c) MEMS-based displays, such as digital micromirror devices(DMDs), are utilized in projectors and digital cinema.

\vspace{1mm}
Healthcare and Medicine: (a) MEMS-based lab-on-a-chip devices allow for rapid and precise analysis of biological samples, enabling diagnostics, DNA sequencing, and disease detection, (b) Microfluidic MEMS devices are used for drug delivery systems, implantable medical devices, and micro-pumps for controlled drug release, and (c) MEMS sensors monitor vital signs in wearable health devices and provide real-time patient data for telemedicine applications.

\vspace{1mm}
Automotive and Transportation: (a) MEMS accelerometers and gyroscopes are essential components in automotive safety systems, such as airbags, electronic stability control, and tire pressure monitoring systems, (b) MEMS pressure sensors are used in engine management and emissions control systems to optimize performance and fuel efficiency, and (c) MEMS-based inertial navigation systems provide precise navigation and positioning for autonomous vehicles and drones.

\vspace{1mm}
Environmental Monitoring: (a) MEMS sensors are used for monitoring air quality, temperature, humidity, and gas concentrations in environmental monitoring systems, and (b) MEMS devices enable remote sensing and data collection for climate research and weather forecasting.

\vspace{1mm}
Aerospace and Defense: (a) MEMS gyroscopes and accelerometers are critical components in aerospace applications for attitude control, navigation, and guidance systems, (b) MEMS pressure sensors are used in altitude and airspeed measurements in aircraft, and (c) MEMS-based infrared imaging devices are utilized in night vision systems for military and security applications.

\vspace{1mm}
Energy Harvesting:(a) MEMS energy harvesters convert ambient energy, such as vibrations or thermal gradients, into electrical power, providing a sustainable energy source for low-power electronics and IoT devices.

\vspace{1mm}
Industrial Automation: (a) MEMS-based sensors are used for condition monitoring, predictive maintenance, and feedback control in manufacturing and industrial processes, improving efficiency and reducing downtime.

\vspace{1mm}
Robotics: MEMS sensors and actuators enable precise motion control and sensing in robots, making them more autonomous and capable of interacting with their environment.

\vspace{1mm}
The potential impact of MEMS in these fields is immense. They contribute to increased efficiency, enhanced functionality, reduced energy consumption, and improved safety in various applications. As MEMS technology continues to advance, we can expect further integration, miniaturization, and performance improvements, leading to even more innovative applications across industries and benefiting society as a whole.  \\

\midrule

\textcolor{black}{Applications: Explore the wide range of applications where MEMs are utilized. Discuss their potential impact in fields such as electronics, energy, medicine, environmental remediation, etc.} & Surface modification plays a crucial role in tailoring the properties of MEMS devices in the nanomaterials category. It involves altering the surface characteristics of the MEMS components to enhance their performance or enable specific applications. Here are some common strategies used for surface modification:

\vspace{1mm}
Functionalization: Functionalization involves attaching or grafting specific molecules or functional groups onto the surface of MEMS devices. This process enhances the surface's chemical reactivity and allows for specific interactions with target substances. Functionalization can be achieved through chemical reactions or self-assembled monolayers (SAMs). Some applications of functionalization in MEMS include:

\vspace{1mm}
Biomolecule Immobilization: Functionalizing the surface with biomolecules, such as antibodies or DNA probes, enables biosensing applications for disease detection and medical diagnostics.

\vspace{1mm}
Gas Sensing: The surface functionalization of MEMS gas sensors with specific materials enhances their selectivity and sensitivity to target gases, making them suitable for environmental monitoring and industrial safety.

\vspace{1mm}
Coating: Surface coating involves depositing thin layers of materials onto the MEMS surface to alter its properties. Coatings can be functional (active) or protective (passive) in nature. Some coating methods include physical vapor deposition (PVD), chemical vapor deposition (CVD), and atomic layer deposition (ALD). Coatings can enhance MEMS performance in various ways, such as: Anti-Stiction Coatings: Coating the MEMS surface with lubricants or hydrophobic materials reduces stiction and friction, which is crucial for reliable operation in micro-mechanical devices.

\vspace{1mm}
Chemical Protection: Coatings can protect MEMS devices from chemical corrosion or degradation, increasing their durability and lifespan in harsh environments.
Optical Coatings: Coatings with specific optical properties can enhance light transmission or reflection, improving the efficiency of optical MEMS devices like micro-mirrors and displays.

\vspace{1mm}
Doping: Doping involves introducing impurity atoms into the material's surface to modify its electrical properties. This strategy is commonly used in semiconductors to create p-type or n-type regions and adjust the device's conductivity. Doping is critical for MEMS devices with electronic functionality, such as sensors and actuators. By carefully controlling the doping process, the sensitivity, response time, and power consumption of MEMS devices can be optimized for specific applications.

\vspace{1mm}
Nanopatterning: Nanopatterning involves creating specific patterns or nanostructures on the MEMS surface. Techniques like electron beam lithography, nanoimprint lithography, and nanocontact printing can be used to fabricate intricate patterns. Nanopatterning enables applications such as:

\vspace{1mm}
Enhanced Adhesion: Nanostructures can improve the adhesion between MEMS components and bonding surfaces, reducing the risk of delamination or failure.
Superhydrophobicity: Nanostructured surfaces can achieve superhydrophobic properties, repelling water and other liquids. This is useful in self-cleaning or water-repellent MEMS devices.

\vspace{1mm}
Overall, surface modification techniques allows to fine-tune the material's properties, improve performance.  \\

\midrule

\textcolor{black}{Toxicity and Safety: Address the potential health and environmental concerns associated with MEMS in this nanomaterials category. Discuss studies on their toxicity,
risk assessment, and safety measures to mitigate any potential hazards.} & As with any emerging technology, there are concerns about the potential health and environmental impacts associated with MEMS in the nanomaterials category. While MEMS devices themselves are generally considered safe, certain nanomaterials and fabrication processes used in their production can raise toxicity and safety issues. It's essential to address these concerns through rigorous studies, risk assessments, and implementation of safety measures. Here are some key points to consider:

\vspace{1mm}
Nanomaterial Toxicity: Nanomaterials, due to their small size and increased surface area, may exhibit different properties and behaviors compared to their bulk counterparts. Some nanomaterials can potentially pose health and environmental risks if they are released into the environment or come into contact with living organisms. Toxicity studies are essential to assess the potential harm caused by exposure to nanomaterials used in MEMS devices.

\vspace{1mm}
Risk Assessment: Researchers and regulatory bodies conduct risk assessments to evaluate the potential hazards associated with the use of nanomaterials in MEMS devices. These assessments consider exposure pathways, potential toxicity, and the likelihood of adverse effects. Risk assessment helps in identifying potential risks and implementing appropriate safety measures to minimize or eliminate hazards.

\vspace{1mm}
Safety Measures: To mitigate potential health and environmental risks, safety measures can be implemented throughout the lifecycle of MEMS devices. These measures include:

\vspace{1mm}
Engineering Controls: Implementing engineering controls during the fabrication process to minimize exposure to hazardous materials and ensure safe handling and disposal of nanomaterials.

\vspace{1mm}
Personal Protective Equipment (PPE): Providing employees with appropriate PPE to prevent inhalation or skin contact with nanomaterials during fabrication or handling of MEMS devices.
Workplace Safety Protocols: Establishing workplace safety protocols and guidelines for the safe handling, storage, and disposal of nanomaterials and MEMS devices.
Environmental Regulation: Complying with environmental regulations and guidelines for the safe disposal of waste materials generated during MEMS fabrication.

\vspace{1mm}
Environmental Impact Assessment: Environmental impact assessments evaluate the potential environmental impacts of introducing MEMS devices into various applications. These assessments help identify potential risks and provide insights into how to design MEMS devices with minimal environmental impact.

\vspace{1mm}
Continued Research: Continued research into the toxicity of nanomaterials and the potential hazards associated with MEMS devices is essential. As the technology advances, new materials and fabrication processes may emerge, warranting further investigation to ensure safety.

\vspace{1mm}
In conclusion, while MEMS devices in the nanomaterials category offer exciting possibilities for various applications, it is crucial to address potential health and environmental concerns associated with nanomaterials. Conducting thorough toxicity studies, risk assessments, and implementing safety measures are essential steps to ensure the responsible and sustainable development and use of MEMS technology. By prioritizing safety and environmental considerations, the benefits of MEMS can be harnessed while minimizing any potential risks.   \\

\midrule

\textcolor{black}{Future Directions: Discuss current research trends and future prospects for MEMS . Highlight emerging technologies, challenges, and areas of active exploration.} & Current research trends and future prospects for MEMS (Micro-Electro-Mechanical Systems) are driven by a combination of technological advancements, societal demands, and industry needs. Here are some key areas of active exploration and emerging technologies in the field of MEMS: 

\vspace{1mm}
Internet of Things (IoT) and Smart Devices: The proliferation of IoT and smart devices is fueling the demand for MEMS sensors and actuators that are smaller, more power-efficient, and capable of providing precise data. Research is focused on developing low-power, miniaturized MEMS devices for applications in smart homes, wearable devices, environmental monitoring, and industrial automation, (b) Energy Harvesting: Energy harvesting using MEMS devices is a promising area of research. Researchers are exploring the use of MEMS energy harvesters to capture ambient energy from vibrations, thermal gradients, and solar radiation to power low-energy electronic devices and sensors, (C) MEMS-NEMS Integration: The integration of MEMS with NEMS (Nano-Electro-Mechanical Systems) is an area of active exploration. NEMS components, due to their smaller size and increased sensitivity, can enhance the performance and capabilities of MEMS devices, especially in sensing and actuation applications, (d) Nanomaterials and Nanofabrication: Advances in nanomaterials and nanofabrication techniques are enabling the development of novel MEMS devices with enhanced functionalities and improved performance. Researchers are exploring nanomaterial-based MEMS devices for applications in gas sensing, chemical detection, and bio-imaging, and (e) 3D Printing and Additive Manufacturing: 3D printing and additive manufacturing are being investigated for rapid prototyping and fabrication of complex MEMS structures. This technology allows for greater design flexibility and customization, facilitating innovation in MEMS development. 

\vspace{1mm}
Challenges: Despite the promising future of MEMS technology, some challenges need to be addressed: (a) Integration Complexity: As MEMS devices become more sophisticated and multifunctional, integration challenges arise. The integration of different materials, electronics, and sensors on a single chip requires precise fabrication techniques and design optimization. (b) Reliability and Longevity: MEMS devices are often used in harsh environments, and ensuring their reliability and longevity under such conditions is crucial. Materials compatibility, stiction, and packaging issues need to be carefully addressed, (c) Standardization: The lack of standardized processes and testing methods for MEMS devices can hinder their widespread adoption and commercialization. Standardization efforts are essential to ensure consistent performance and compatibility across different MEMS devices, and (d) Scaling and Manufacturing Costs: Scaling down MEMS devices to nanoscale dimensions presents manufacturing challenges and can increase production costs. Cost-effective fabrication methods for mass production are crucial for widespread adoption.\\

\midrule

\textcolor{black}{} & Future Prospects: The future prospects for MEMS technology are promising, with potential impact in various sectors: (a) Healthcare and Medical Devices: MEMS-based medical devices are expected to revolutionize diagnostics, treatment, and personalized medicine, leading to better patient outcomes and healthcare efficiency, (b) Autonomous Systems: MEMS sensors and actuators will play a crucial role in enabling autonomous vehicles, drones, and robotics, advancing automation and safety across industries, (c) Environmental Monitoring and Sustainability: MEMS devices will continue to contribute to environmental monitoring and sustainability efforts by providing accurate data on pollution levels, climate change, and resource management, (d) Smart Cities: MEMS technology will be integral to the development of smart cities, enabling efficient infrastructure management, traffic control, and environmental monitoring, and (e) Wearable and Implantable Devices: MEMS-based wearable and implantable devices will enable continuous health monitoring and intervention, improving overall well-being and quality of life.

\vspace{1mm}
In conclusion, the future of MEMS technology holds great promise, with advancements in nanomaterials, 3D printing, IoT, and healthcare applications driving innovation. Addressing current challenges and promoting collaboration between researchers, industry, and regulatory bodies will be key to unlocking the full potential of MEMS and realizing its transformative impact across various sectors.   \\

\bottomrule
\end{tabularx}

\bibliographystyle{plain}
\bibliography{reference}

%%%%%%%%%%%%%%%%%%%%%%%%%%%%%%%%%%%%%%%%%%%%%%%%%%%%%%%%%%%%

\end{document}